\documentclass[final]{cvpr}

\usepackage{times}
\usepackage{epsfig}
\usepackage{graphicx}
\usepackage{amsmath,amsfonts,bm}

\def\eqref#1{equation~\ref{#1}}
\def\1{\bm{1}}

\newcommand{\valid}{\mathcal{D_{\mathrm{valid}}}}

\def\vz{{\bm{z}}}

\DeclareMathAlphabet{\mathsfit}{\encodingdefault}{\sfdefault}{m}{sl}
\SetMathAlphabet{\mathsfit}{bold}{\encodingdefault}{\sfdefault}{bx}{n}

\def\sO{{\mathbb{O}}}

\newcommand{\pder}[2]{\frac{\partial #1}{\partial #2}}

\newcommand{\Ls}{\mathcal{L}}

\newcommand{\softmax}{\mathrm{softmax}}

\DeclareMathOperator*{\argmax}{arg\,max}
\DeclareMathOperator*{\argmin}{arg\,min}

\usepackage{amssymb}
\usepackage{amsthm}
\usepackage{multirow}
\usepackage{subcaption}
\usepackage{booktabs}

\newtheorem{theorem}{Theorem}[section]

\usepackage[pagebackref=true,breaklinks=true,colorlinks,bookmarks=false]{hyperref}

\begin{document}

%%%%%%%%% TITLE
\title{Single-level Optimization For Differential Architecture Search}

\author{Pengfei Hou\\
{\tt\small houpengfei2020@126.com}
\and
Ying Jin\\
{\tt\small jiny18@mails.tsinghua.edu.cn}
}

\maketitle

%%%%%%%%% ABSTRACT
\begin{abstract}
In this paper, we point out that differential architecture search (DARTS) makes gradient of architecture parameters biased for  network weights and architecture parameters are updated in different datasets alternatively in the bi-level optimization framework. The bias causes the architecture parameters of non-learnable operations to surpass that of learnable operations. Moreover, using softmax as architecture parameters' activation function and inappropriate learning rate would exacerbate the bias. As a result, it's frequently observed that non-learnable operations are dominated in the search phase.
To reduce the bias, we propose to use single-level to replace bi-level optimization and non-competitive activation function like sigmoid to replace softmax.  As a result, we could search high-performance architectures steadily. Experiments on NAS Benchmark 201 validate our hypothesis and stably find out nearly the optimal architecture. On DARTS space, we search the state-of-the-art architecture with 77.0\% top1 accuracy (training setting follows PDARTS and without any additional module) on ImageNet-1K and steadily search architectures up-to 76.5\% top1 accuracy (but not select the best from the searched architectures) which is comparable with current reported best result.

\end{abstract}

%%%%%%%%% BODY TEXT
\section{Introduction}

Neural architecture search (NAS) has helped to find more optimal architecture than manual design. Generally NAS is formulated as a bi-level optimization problem\cite{baker2016designing} as:
\begin{equation}\label{equ:nas1}
\begin{split}
    \alpha^* &= \argmin_{\alpha \in \bm{A}} \mathcal{L}_{val}(\alpha, w_\alpha^*) \\
    s.t.\ w_\alpha^* &= \argmin_{w_{\alpha}} \mathcal{L}_{train}(\alpha, w)\\
\end{split}
\end{equation}
where $\alpha$ denote architecture and $\bm{A}$ denote architecture search space, $w_\alpha$ denote the network weights bound with the architecture $\alpha$, $\mathcal{L}_{train}$ and $\mathcal{L}_{val}$ denote optimization loss on training and validation dataset. Due to inner optimization on $\mathcal{L}_{train} $ that any architecture has to be trained fully, therefore it costs huge computation sources to search the optimal architecture. To avoid training each architecture from scratch, weight-sharing methods \cite{liu2018darts} are proposed to construct a super network where all architectures share the same weights. DARTS relaxes the search space to be continuous and approximates $w^*_\alpha$ by adapting $w$ using only a single training step, without solving the inner optimization \ref{equ:nas1} completely. The approximation scheme is as follows:
\begin{equation}\label{equ:darts}
	\nabla_\alpha{ \mathcal{L}_{val}(\alpha, w^*_{\alpha})} \\
	\approx \nabla_\alpha  \mathcal{L}_{val}(\alpha, w-\xi\nabla_w{ \mathcal{L}_{train}(\alpha, w)}) 
\end{equation}
 It saves computation costs a lot and finds out competitive architectures \cite{cai2018proxylessnas, xu2019pc}.\\
 
\begin{figure}[]
%\begin{center}
\centering
%\fbox{\rule[-.5cm]{0cm}{4cm} \rule[-.5cm]{4cm}{0cm}}
 \begin{subfigure}[b]{0.49\linewidth}
 \centering
\includegraphics[width=1\textwidth]{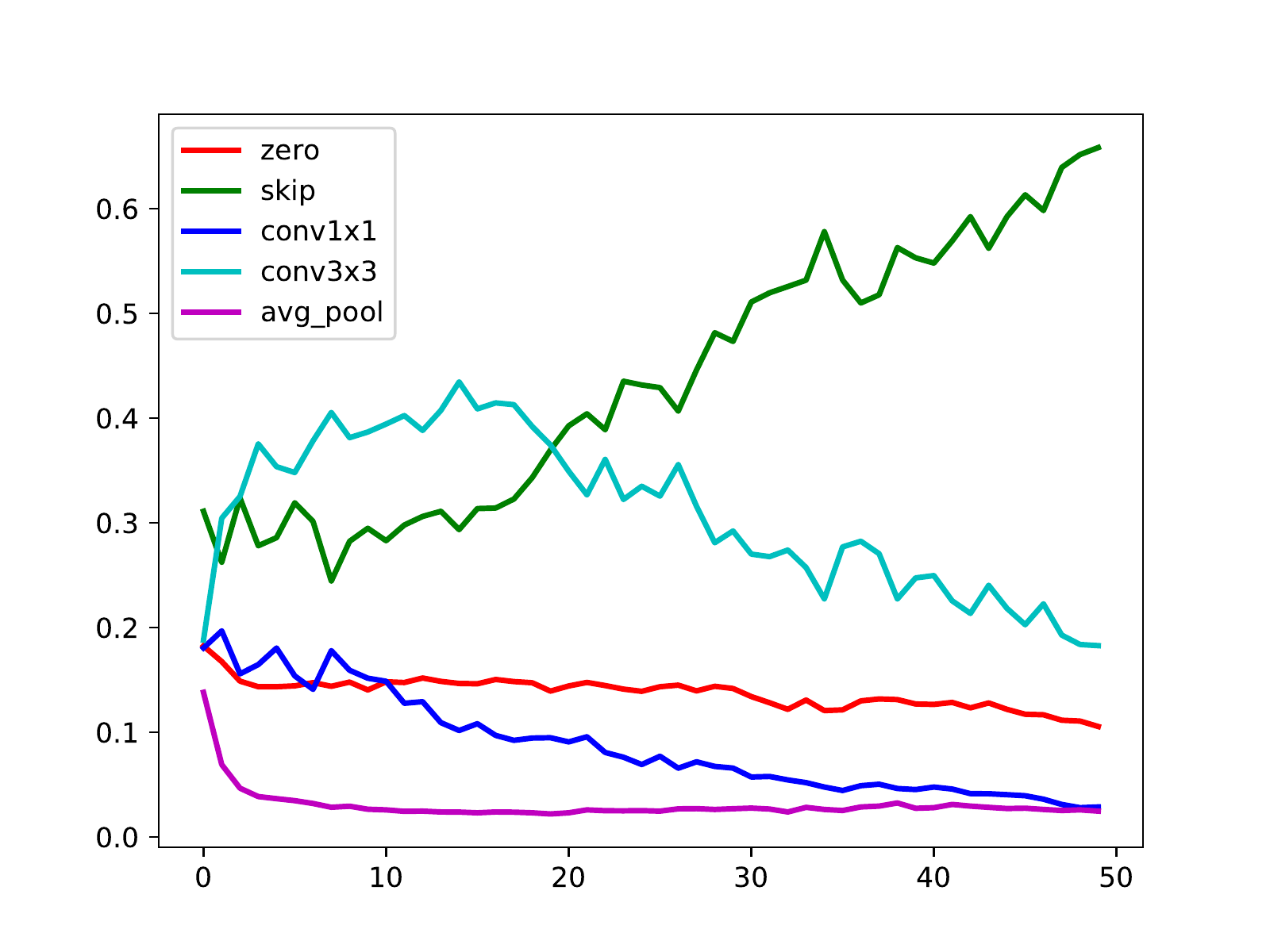}
 \caption{edge.$1-0$}
\end{subfigure}
\hfill
 \begin{subfigure}[b]{0.49\linewidth}
 \centering
\includegraphics[width=\textwidth]{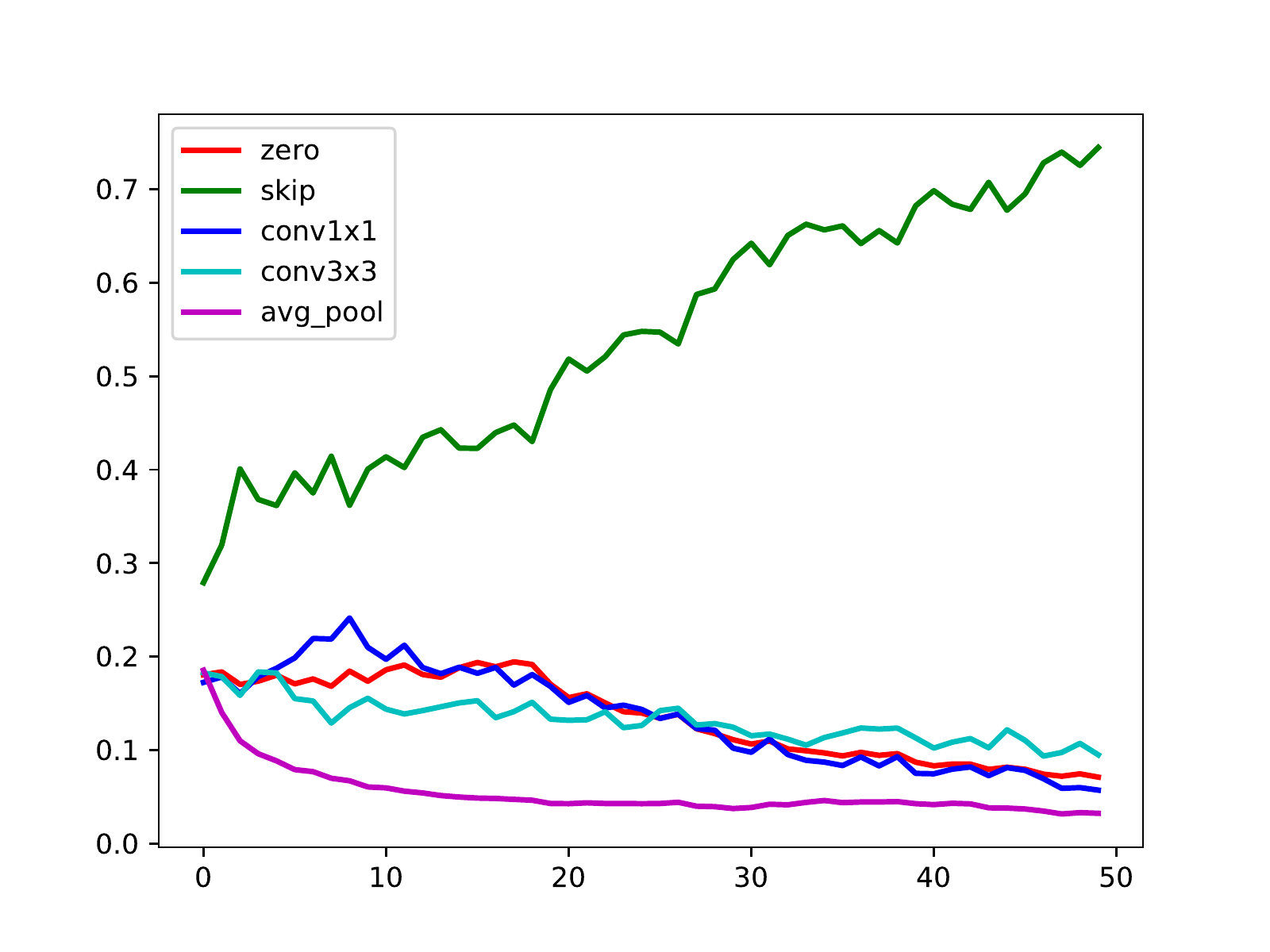}
 \caption{edge.$2-0$}
\end{subfigure}
\hfill
\quad
 \begin{subfigure}[b]{0.49\linewidth}
 \centering
\includegraphics[width=\textwidth]{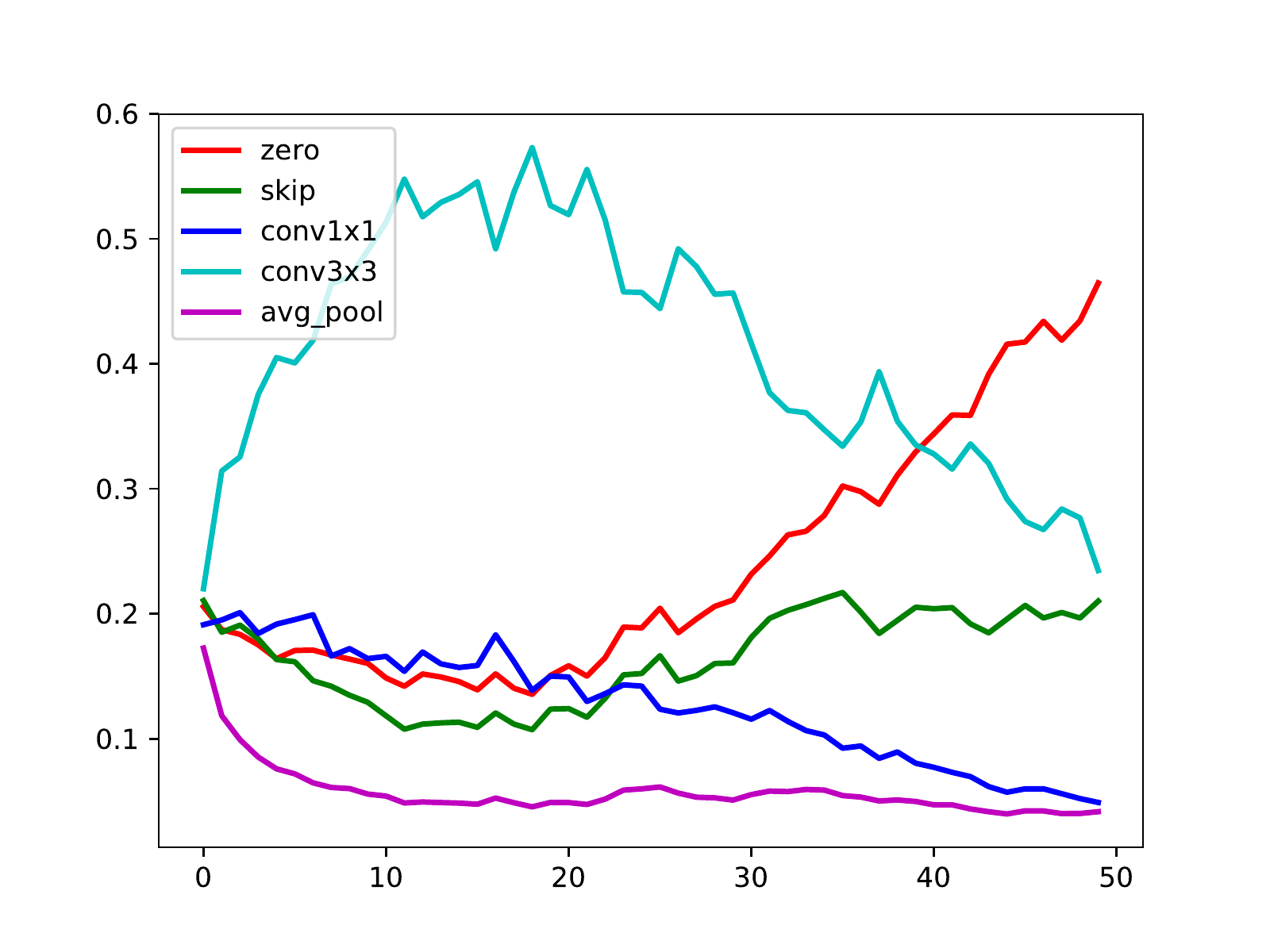}
 \caption{edge.$2-1$}
\end{subfigure}
\hfill
 \begin{subfigure}[b]{0.49\linewidth}
 \centering
\includegraphics[width=\textwidth]{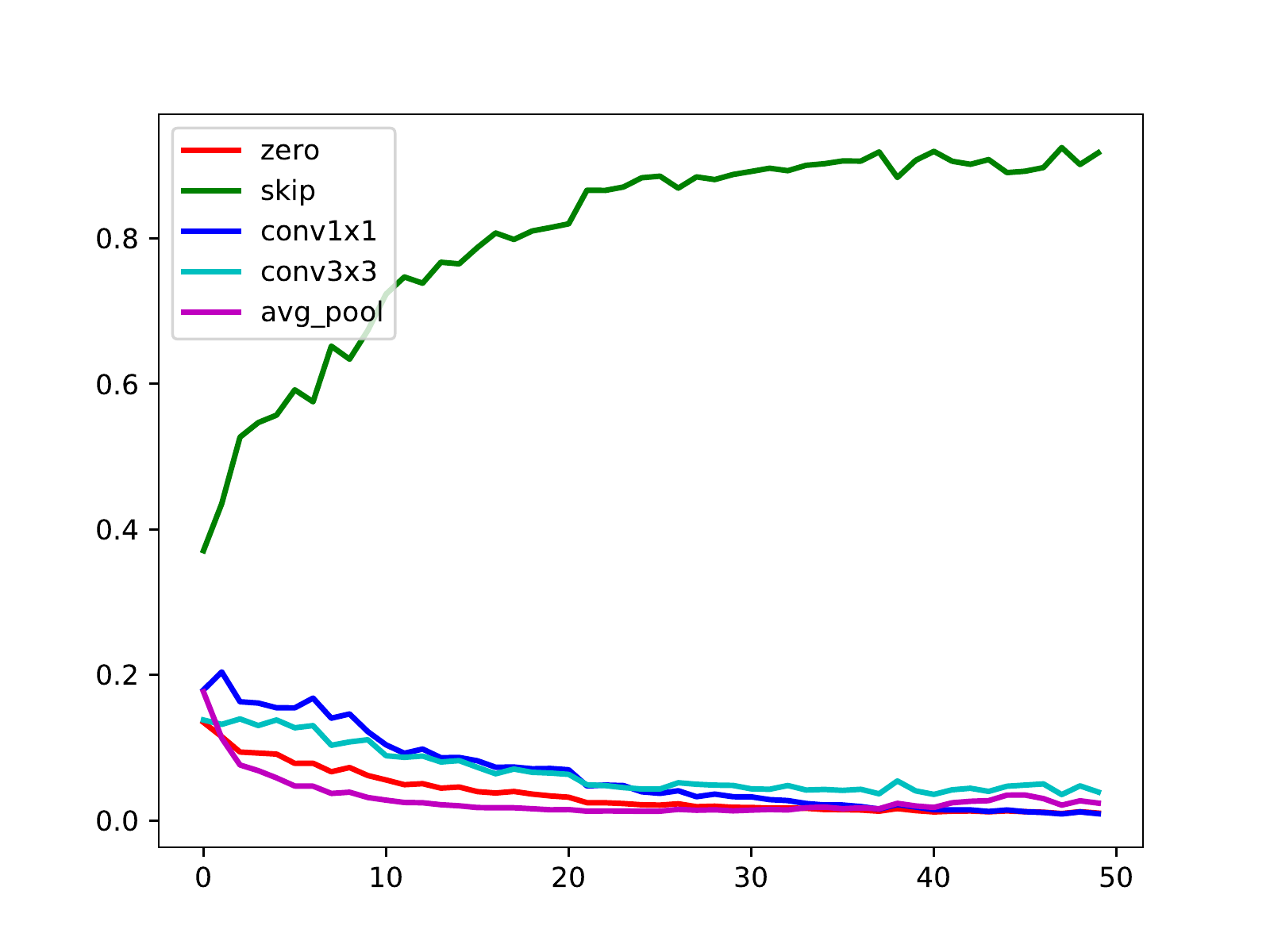}
 \caption{edge.$3-0$}
\end{subfigure}
\hfill
\quad
 \begin{subfigure}[b]{0.49\linewidth}
 \centering
\includegraphics[width=\textwidth]{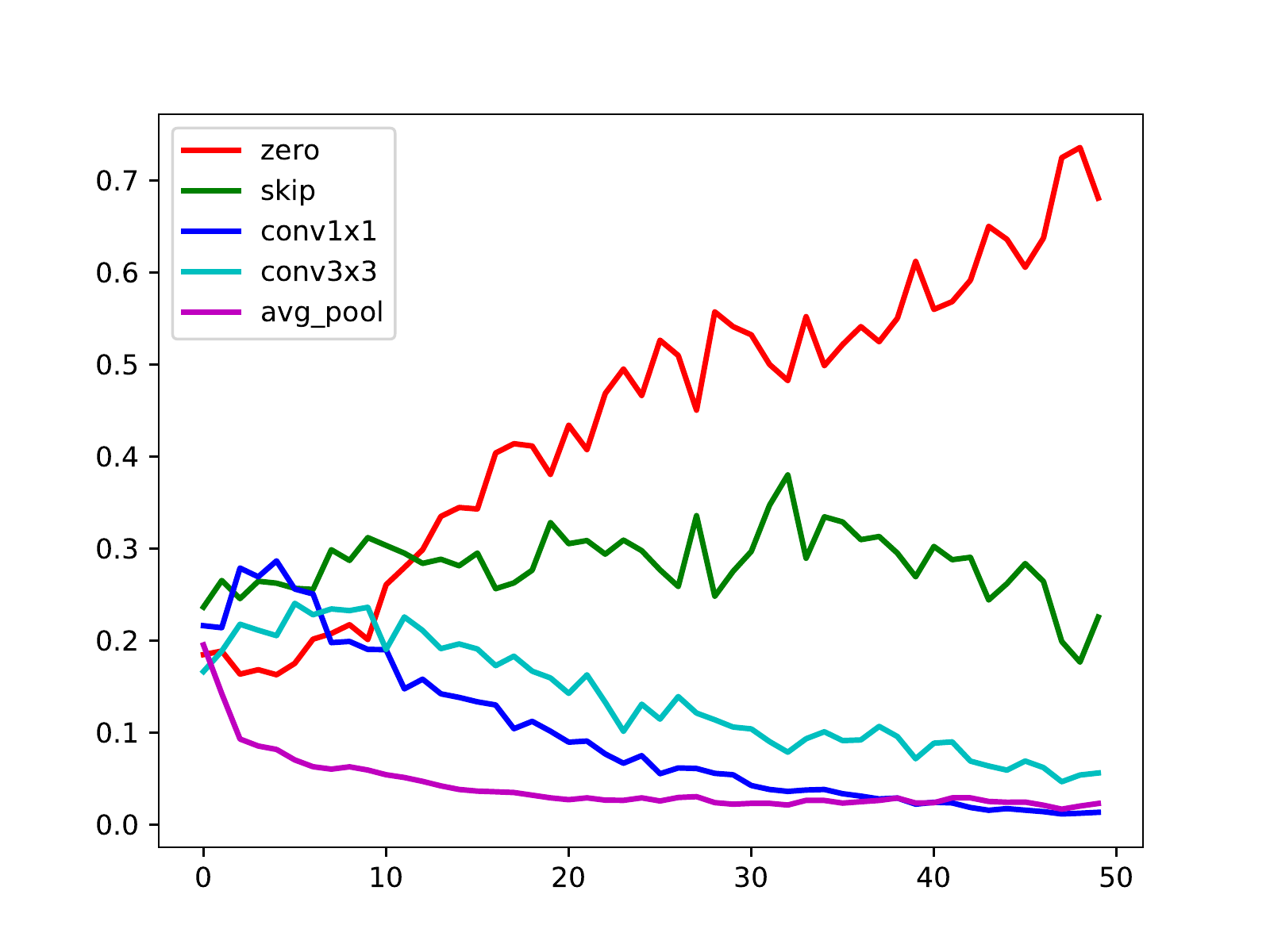}
 \caption{edge.$3-1$}
\end{subfigure}
\hfill
 \begin{subfigure}[b]{0.49\linewidth}
 \centering
\includegraphics[width=\textwidth]{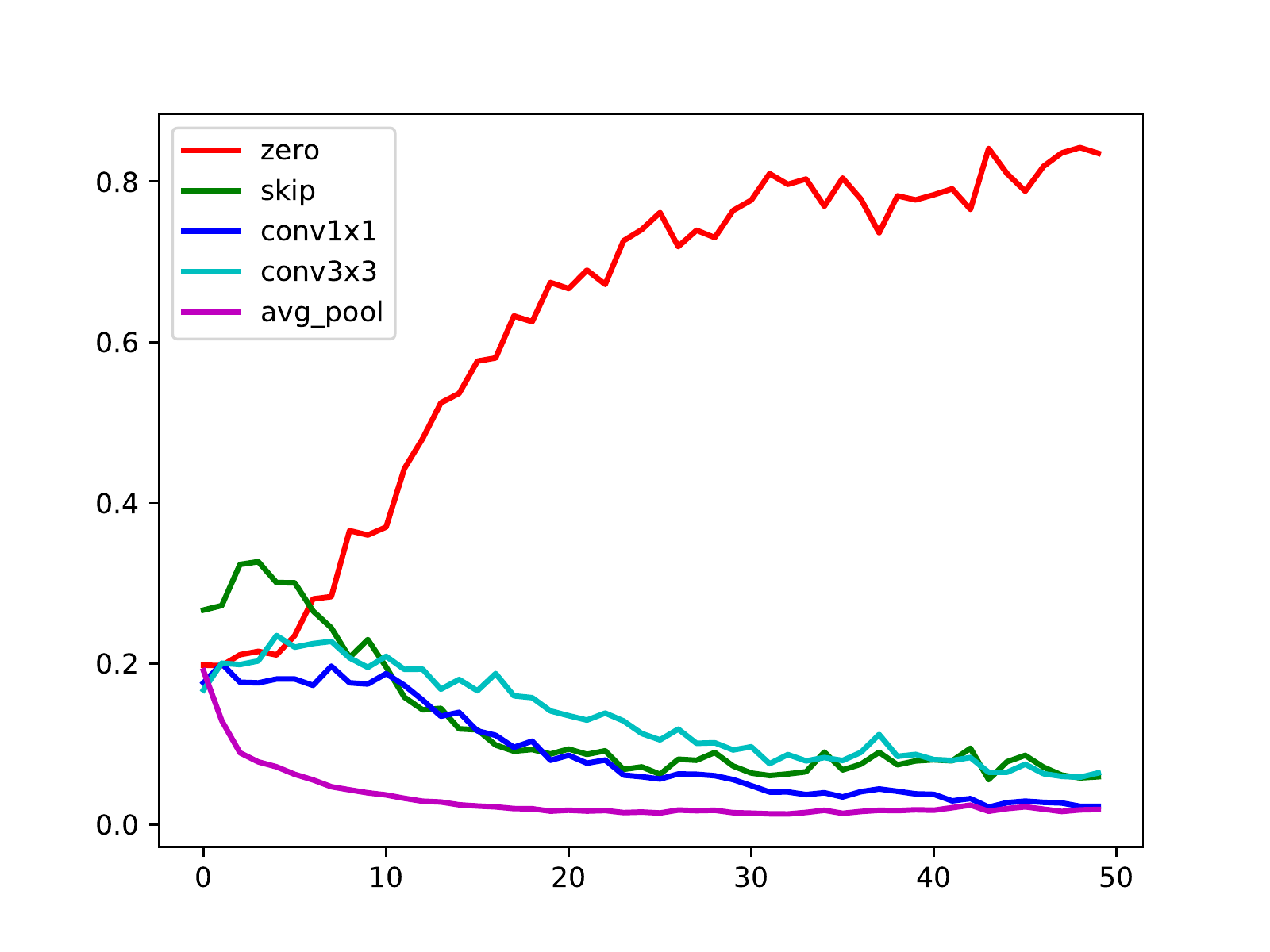}
 \caption{edge.$3-2$}
\end{subfigure}
\caption{Values of $\softmax (\alpha)$(probability) on NAS-Benchmark-201. Use bi-level optimization(DARTS) to train for 50 epochs on CIFAR10. Non-learnable operations' probability is far bigger than learnable operations'. It shows that non-learnable operations are likely dominated in the early stage and the situation would last until the end.} 
\label{fig:nonop_201}
\end{figure}

However, many papers \cite{xu2019pc, liu2018progressive, zela2019understanding, li2019stacnas, chu2019fair, dong2020bench} have reported that DARTS easily converges to non-learnable operations including skip-connet, pooling and zero, etc. and it could not search steadily high-performance architectures. 
Furthermore we find out that non-learnable operations are dominated in the very early stage of search phase. And once the situation happens,
%it could not be retrieved any more. 
it's likely to last until the end of the search phase \ref{fig:nonop_201}.

For the phenomenon, we propose the hypothesis that it's caused from two aspects and try to give theoretic explanation. 
For one thing, the approximation to bi-level optimization which has to train $w$ and $\alpha$ on different datasets computes biased gradients for $\alpha$. Even the gradients of $w$ and $\alpha$ are computed on the same dataset but not on the same batch, the bias also exists. On the early stage of training process,  learnable operations' $\alpha$ could not be learned normally. As a result, non-learnable operations' $\alpha$ would surpass over learnable operations'. 
For the other thing, using softmax as $\alpha$'s activation function and inappropriate (in specific is too large) learning rate  would make surpassing architecture parameters larger. It means under the bi-level optimization framework the non-learnable operations would dominate over learnable operations from the start to finish. \\

In view of the above two aspects, we conduct experimental analysis to verify them based on NAS-Benchmark-201\cite{dong2020bench}. Furthermore, we give improvement on DARTS. On one side, we propose to use single-level optimization instead of bi-level optimization and update $w$ and $\alpha$ on the same data batch. On another side, we propose to use uncompetitive activation function like sigmoid to replace softmax. And if there are too many non-operations in the searched result, we advice to decrease learning rate. As a result, we steadily search high-performance architectures.
Experiments on NAS-Benchmark-201\cite{dong2020bench} search nearly the optimal architecture in the space with variance of 0. On DARTS space, we search the SOTA architecture with top1 accuracy of 77\% on ImageNet-1K \cite{krizhevsky2012imagenet} for the first time. And we could search architectures steadily with top1 accuracy up to 76.5\% which is comparable with reported best result 76.6\% \cite{hong2020dropnas, yu2020cyclic}. The training phase is the same as PDARTS\cite{liu2018progressive} and without any additional module.\\

In general, our contributions are as follows:
\begin{itemize}
\item We find out that non-learnable operations dominate learnable operations from the start to finish. According to the phenomenon, we propose a hypothesis that bi-level optimization bias the gradients of architecture parameters. Using softmax as activation function and inappropriate learning rate would make the bias more serious. We give a theoretical explanation and conduct experiments to verify it.\\
\item As a substitude, we propose to use single-level optimization which calculates and backward the gradients of $w$  and $\alpha$ on the same data batch as well as at the same time: 
\begin{equation}\label{equ:single-level}
\begin{split}
    \alpha^t, w^t &+= \eta \nabla_{\alpha, w} \mathcal{L}_{train} (\alpha^{t-1}, w^{t-1})
\end{split}
\end{equation}
And we use uncompetitive activation function like sigmoid to replace softmax. Meanwhile, we do normalization for the initialization of sigmoid. More details are introduced bellow. \\
\item Our improvement on DARTS could search high-performance architectures with little variance. On NAS-Benchmark-201 we find out nearly the optimal architecture. On DARTS space, we find the SOTA architecture which get 77.0\% top1 accuracy on ImageNet. And we steadily find architectures (not select the best architecture of several tries) up to 76.5\% top1 accuracy which is comparable with current reported best result.
\end{itemize}

%-------------------------------------------------------------------------

\section{Related Works}
\textbf{Neural Architecture Search.} Neural architecture search(NAS) is an automatic method to design neural architecture instead of human design. Early NAS methods adopt reinforcement learning (RL) or evolutionary strategy \cite{zoph2016neural, baker2016designing, bello2017neural, real2017large, real2019regularized, zoph2018learning} to search among thousands of individually trained completely networks, which costs huge computation sources. Recent works focus on efficient weight-sharing methods, which falls into two categories: one-shot approaches \cite{brock2017smash, bender2018understanding, akimoto2019adaptive, cai2019once, guo2019single, stamoulis2019single, pham2018efficient} and gradient-based approaches \cite{shin2018differentiable, liu2018darts, chen2019progressive, cai2018proxylessnas, hundt2019sharpdarts, chu2019fair, xu2019pc, liang2019darts}, both achieve state-of-the-art results on a series of tasks \cite{chen2019detnas, ghiasi2019fpn, liu2019auto, xu2020autosegnet, fu2020autogan, nascimento2020finding} in various search space. They construct a super network/graph which share weights with all sub network/graphs. The former commonly does heuristic search and evaluation sampled architectures in the super network to get the optimal architecture. The latter relaxes the search space to be continuous and introduce differential and learnable architecture parameters. In the this paper, we mainly argue gradient-based approaches.\\

\textbf{Differential Architecture Search.} Gradient-based approaches are commonly formulated as an approximation to the bi-level optimization which updates network weights and updates architecture parameters in the training and validation dataset alternatively\cite{liu2018darts}. Although it has searched competitive architectures, however, many papers pointed out DARTS-based methods don't work stably. \cite{dong2020bench} show DARTS perform badly on NAS-Benchmark-201 and performance drops fast during search. There are many works trying to give explanation and solve it. \cite{bender2018understanding} show the relationship between DARTS searched architectures' performance and the domain eigenvalue of $\nabla_{\alpha}^2 \mathcal{L}_{valid}$. \cite{chu2019fair} show darts easily converge to skip-connect operation. \cite{liu2018progressive, xu2019pc, hong2020dropnas, li2019stacnas} show DARTS also easily over-converge to non-parametric operations. \cite{chu2019fair} introduces collaborative competition approach which offer each operation an independent architecture weight to avoid competition between operations. \cite{hong2020dropnas} observes the co-adaption problem and Matthew effect that operations with less parameters are trained maturely earlier. However, we indicate that Matthew effect is mainly caused by gradient bias in bi-level optimization but not co-adaption problem. In fact, in single-level optimization, the Matthew effect would be disappeared.\\ 

\textbf{Single-level Optimization.} Recently there are some works adopt single/one-level optimization to replace bi-level optimization and get much more stable results. 
Although in the original paper \cite{liu2018darts} it shows that single-level performs worse than bi-level optimization and they indicate single-level would cause overfitting, but recent works show the inverse results \cite{li2019stacnas, bi2020gold, hong2020dropnas}. \cite{li2019stacnas} advocates the overfitting phenomenon is caused by network's depth gap between search phase and evaluation phase and get much more stable results than bi-level-based DARTS. \cite{bi2020gold} indicates that super network parameters are trained much more effective than architecture parameters and do data augmentation in search phase. And based on single-level optimization, 
\cite{hong2020dropnas} combines operations dropout with single-level optimization to solve co-adaptation problem in the paper. However, on one side, these works focus on other technology to get satisfactory results but single-level optimization is as basis. Our experiments demonstrate that stand-alone single-level optimization is enough to search steadily high-performance architectures. On the other side, these works show single-level better than bi-level optimization mainly by empirical experiments but not give clear explanation on why bi-level doesn't work.  In this paper, we propose that the reason is from biased gradients caused by calculating gradients on different data batch in the bi-level optimization framework. %And there are few works  concerning on steadily searching the best architectures.

\section{Hypothesis And Methodology}
In this section, we propose the hypothesis of the gradient bias of architecture parameters in the bi-level optimization framework and try to give theoretic explanation on the domination phenomenon of non-learnable operations. 
In outline, bi-level optimization has to compute gradients of network weights $w$ and architecture parameters $\alpha$ separably on the training and validation dataset. It's the computation on different dataset that makes the gradients of learnable operations' $\alpha$  biased. Moreover, the bias leads non-learnable operations' $\alpha$ accumulating gradient faster than learnable operations'. With softmax as $\alpha$'s activation function, the gap between them would be expanded. Finally, it results in the domination phenomenon. 
\subsection{Preliminary of Differential Architecture Search}
Following \cite{liu2018darts} we search the DARTS space which stacks of several cells, where each cell is a directed acyclic graph.  Each cell consists of sequential nodes where each node represent latent feature map $x^i$. The edge from node $i$ to node $j$ represent a connection and is bounded with one from candidate operations $\sO$: convolution, pooling, zero, skip-connect, etc. Let $o^{i, j}$ denote operation from node $i$ to $j$. Each intermediate node is computed based on all of its predecessors:
\begin{equation}
x_{j} = \sum_{i<j}o^{i,j}(x_i))
\end{equation}
The problem of searching the best architecture is transformed to search the best operation  $o^{i,j}_{k^*}$ among all cells and edges.
DARTS make the search space continuous by relaxing the categorical choice of a particular operation to a softmax over all possible operations. Let $\alpha^{i, j}_{o_k}$ or $\alpha^{i, j}_k$ denote bounded parameter with operation $o_k$
\begin{equation}
	f_{i,j}(x_i) = \sum_{o \in O} \frac{exp(\alpha^{i,j}_o)}{\sum_{o' \in O}exp(\alpha^{i,j}_{o'})} o(x_i)
\end{equation}
Then the intermediate node is computed based on all of its predecessors:
\begin{equation}
x_j = \sum_{i<j}f_{i,j}(x_i))
\end{equation}
Without loss of generality, we consider a single edge.
Loss function could be seen as 
\begin{equation}\label{darts_format}
 \mathcal{L}  = f(\sum_i \frac{exp(\alpha_i)}{\sum_j exp(\alpha_j)}o_i(x) ; X, y)
\end{equation}

\subsection{Hypothesis}
Prior works point out that DARTS easily converge to non-parametric operations like skip-connect, zero, pooling, etc. In fact, we find that in the early stage of search phase, non-learnable operations' architecture parameters have suppressed learnable operations'. Once the situation happens, it's hard to be retrieved \ref{fig:nonop_201}.  For the phenomenon, we have the following explanation. \\
Let $p_i$ denote $\frac{exp(\alpha_i)}{\sum_j exp(\alpha_j)}$, assuming $o_i(x)$ could be expanded as $W_i x$, we have \\
\begin{equation}\label{form:grad_pi}
\pder{\Ls}{p_i} = \mathbb{E}[\pder{\Ls}{\bar{x}}^T (W_i x); X, y]
\end{equation}
\begin{equation}
\pder{\Ls}{W_i} = \mathbb{E}[p_i \pder{\Ls}{\bar{x}} x^T; X, y]
\end{equation}
Considering $\pder{\Ls}{p_i}$, if $W_i x$ is negative correlated with $\pder{\Ls}{\bar{x}}$, $p_i$ would be increased under gradient descent. Heuristically, if $W_i$ fit $x$ better than $W_j$, then $\pder{\Ls}{p_i}$ should be smaller than $\pder{\Ls}{p_j}$. 
For DARTS, $p_i$ is updated on the validation dataset. On the iteration $t$ and on validation dataset $\valid = \{X_{val}, y_{val}\}$, we have:
\begin{equation}
\pder{\Ls}{p_i^t} = \frac{1}{N}\sum_{j=1}^N \pder{\Ls(x^j_{val})}{\bar{x}}^T (W_i^t x_{val}^j)
\end{equation}
$W_i^t$ is updated on the training dataset $\mathcal{D_{\mathrm{train}}} = \{X_{train}, y_{train}\}$, we have:
\begin{equation}
\begin{split}
\pder{\Ls}{W_i^t} &= \frac{p_i}{M}\sum_{k=1}^M \pder{\Ls(x^k_{train})}{\bar{x}} {x_{train}^k}^T \\
W_i^t &= W_i^{t-1} - \eta\pder{\Ls}{W_i^{t-1}} \\
	 &= W_i^{t-1} - \eta\frac{p_i^{t-1}}{M}\sum_{k=1}^N \pder{\Ls(x^k_{train})}{\bar{x}} {x_{train}^k}^T \\
\end{split}
\end{equation}
As a result, the gradient of $p_i$ is:
\begin{equation}
\begin{split}
&\pder{\Ls}{p_i^t} 
= \frac{1}{N}\sum_{j=1}^N \pder{\Ls(x^j_{val})}{\bar{x}}^T (W_i^{t-1} x_{val}^j) \\
&-  \frac{\eta p_i^{t-1}}{NM}\sum_{j=1}^N \sum_{k=1}^M (\pder{\Ls(x^j_{val})}{\bar{x}}^T \pder{\Ls(x^k_{train})}{\bar{x}}) ({x_{train}^k}^T x_{val}^j)\\
\end{split}
\label{gradmult}
\end{equation}
For bi-level optimization $W$ and $\alpha$ are computed on the different batches, if samples are different $\pder{\Ls(x^j_{val})}{\bar{x}}^T$ is independent $\pder{\Ls(x^k_{train})}{\bar{x}}$ and  ${x_{train}^k}$ is independent of $x_{val}^j$. Thus we have 
\begin{equation}
\label{equ:corr}
\sum_{j=1}^N \sum_{k=1}^M (\pder{\Ls(x^j_{val})}{\bar{x}}^T \pder{\Ls(x^k_{train})}{\bar{x}}) ({x_{train}^k}^T x_{val}^j) \approx 0
\end{equation}
Furthermore, on the early stage when iteration is not enough we have
\begin{equation}
\begin{split}
\pder{\Ls}{p_i^t} &\approx \frac{1}{N}\sum_{j=1}^N \pder{\Ls(x^j_{val})}{\bar{x}}^T (W_i^{t-1} x_{val}^j) \\
		          &\approx\frac{1}{N}\sum_{j=1}^N \pder{\Ls(x^j_{val})}{\bar{x}}^T (W_i^0 x_{val}^j) \\
\end{split}	
\end{equation}
	 
Learnable operations are initialized randomly thus they are actually making noise on input $x$ and fitting $x$ worse than non-learnable operations like skip-connect and pooling. If $W_i$ fit $x$ better, $\pder{\Ls}{p_i^t}$ would be smaller and $p_i$ would be increased. As a result, in the early stage, non-learnable operations' $p_i$ would be increased faster than learnable operations. Furthermore, for the biggest $\alpha_i$ (assume $\alpha_0$), it's more likely that the gradient of $\alpha_0$ smaller than others compared under softmax activation.
And large learning rate would expand the gap. It means DARTS is more easily converge to $\alpha_0$. 
 We have the following theorems.
\begin{theorem}\label{lemma1}
For function like $l=f(\sum_i  \frac{exp(\alpha_i)}{\sum_j exp(\alpha_j)} \vz_i)$, let $p_i = \frac{exp(\alpha_i)}{\sum_j exp(\alpha_j)}$, $\vz=\sum_i p_i\vz_i$, $i^*=\argmax_i \frac{\partial l}{\partial\vz}\vz_i$, we define margin $\delta=\min_{i\neq i^*}{\frac{\partial l}{\partial\vz}(\vz_{i^*} - \vz_i)} \geq 0$, if $\alpha_{i^*} \geq \alpha_i$, it holds that 
for any $\epsilon > 0$, function gradient achieves $p_{i^*} > 1 - \epsilon$ in iterations
\begin{equation}
 t \leq \frac{n \ln((1-\epsilon)n )}{\eta \delta}
\end{equation}
\end{theorem}
Please see proof in Appendix. However, for single-level optimization, we have
\begin{equation}
\begin{split}
\pder{\Ls}{p_i^t} 
%&= \frac{1}{N}\sum_{j=1}^N \pder{\Ls(x^j_{val})}{\bar{x}}^T ((W_i^{t-1} - \frac{\eta p_i^{t-1}}{M}\sum_{k=1}^M \pder{\Ls(x^k_{train})}{\bar{x}} {x_{train}^k}^T) x_{val}^j) \\
&\approx \frac{1}{N}\sum_{j=1}^N \pder{\Ls(x^j_{val})}{\bar{x}}^T (W_i^{t-1} x_{val}^j) \\
&-  \frac{\eta p_i^{t-1}}{N^2}\sum_{j=1}^N (\pder{\Ls(x^j_{train})}{\bar{x}}^T \pder{\Ls(x^{j}_{train})}{\bar{x}}) ({x_{train}^j}^T x_{train}^{j})\\
&\leq \frac{1}{N}\sum_{j=1}^N \pder{\Ls(x^j_{val})}{\bar{x}}^T (W_i^{t-1} x_{val}^j) \\
\end{split}
\end{equation}
It means for learnable operations, $\pder{\Ls}{p_i}$ could be decreased as training process goes. And after adequate iterations, learnable operations $\pder{\Ls}{p_i}$ would be smaller than non-learnable operations. Under gradient descent, $p_i$ would be increased faster than non-learnable operations until exceed them. 

\subsection{Methodology}
Above explanation shows that:\\
\begin{itemize}
\item Bi-level optimization make gradients of architecture parameters biased for that training data makes no direct effort on the gradients of architecture parameters. In fact, even doing bi-level optimization on the same dataset but not on the same data batch, the gradients could yet be biased. \\
\item Using softmax as $\alpha$'s activation function and inappropriate learning rate would expand the gap between non-learnable and learnable operations.
\end{itemize}
Therefore, we propose:
\begin{itemize}
\item Use single-level optimization meanwhile calculate gradients of $w$ and $\alpha$ at the same time to instead bi-level optimization.
\item Use sigmoid or other uncompetitive activation function to replace softmax.
\item Decrease the learning rate if there are too many non-learnable operations.
\end{itemize}

\section{Experiments}
\label{section:exp}
In this section, we introduce our experiments on different datasets and search spaces and mainly divided into two parts. One is to do experiments on NAS-Bench-201 to verify our hypothesis. %that bi-level optimization could cause gradient bias and softmax and big learning rate would cause the bias serious.
The other part is to compare our improved DARTS (single-level optimization with sigmoid activation function) with different algorithms. For the first part, specifically, we firstly analyze the gradient correlation \ref{equ:corr} and the gradients of $p_i$ in bi-level optimization and single-level optimization. Then we compare the effect of using softmax as activation function with other independent activation function. At last we compare different learning rates' effect. \\

%the biased gradients are affected by whether $\alpha$ and $w$ are updated on the same data batch.
For the second part, we do experiments in NAS-Benchmark-201 and DARTS space, and on Cifar10 and ImageNet-1K dataset. Experiments show that we could steadily search high-performance architectures. On NAS-Benchmark-201 space, we stably find out nearly the optimal architecture. On DARTS space, we search the state-of-the-art architecture with 77.0\% top1 accuracy (training setting follows PDARTS and without any additional module) on ImageNet-1K and steadily search architectures up-to 76.5\% top1 accuracy which is comparable with current reported best result.

\subsection{Verification On NAS Benchmark 201}
\label{nas201}
\subsubsection{Search space.} 
NAS-Bench-201 \cite{dong2020bench} builds a cell-based search space, where a cell could be seen as a directed acyclic graph consisting of 4 nodes and 6 edges. Each network is stacked by 15 cells. Each edge represents an operation selected from (1) zeroize, (2) skip connection, (3) 1-by-1 convolution, (4) 3-by-3 convolution, and (5) 3-by-3 average pooling. The search space has 15,625 neural cell candidates in total. And all the candidates are given training accuracy/valid accuracy/test accuracy on three datasets: CIFAR-10, CIFAR-100 and ImageNet-16-120. In order to facilitate the analysis, in this section we set optimizers of network weights and architecture parameters as SGD.

\subsubsection{Analysis on Gradient Correlation}
\begin{figure}[]
\begin{center}
%\fbox{\rule[-.5cm]{0cm}{4cm} \rule[-.5cm]{4cm}{0cm}}
\includegraphics[width=0.3\linewidth]{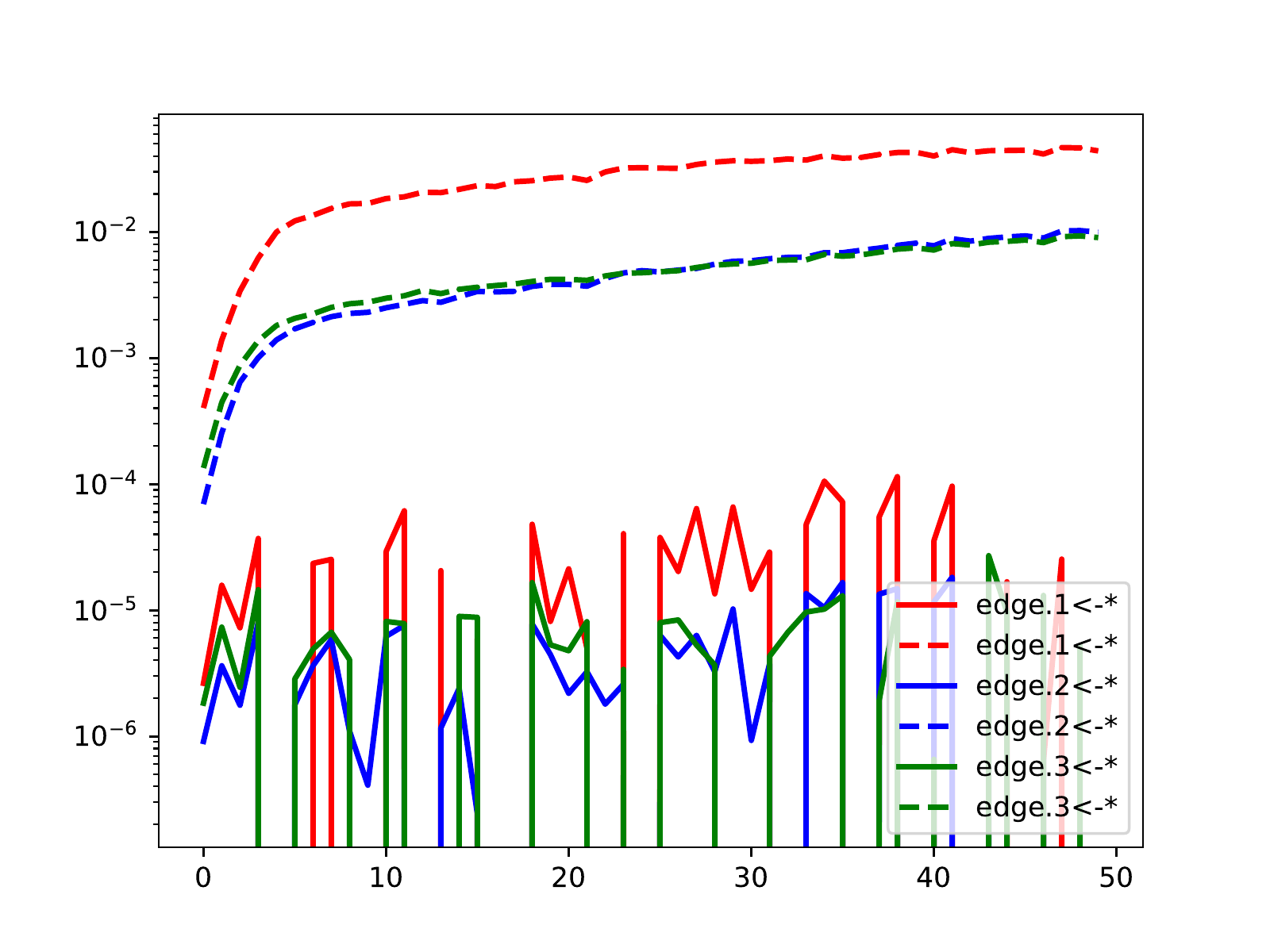}
\includegraphics[width=0.3\linewidth]{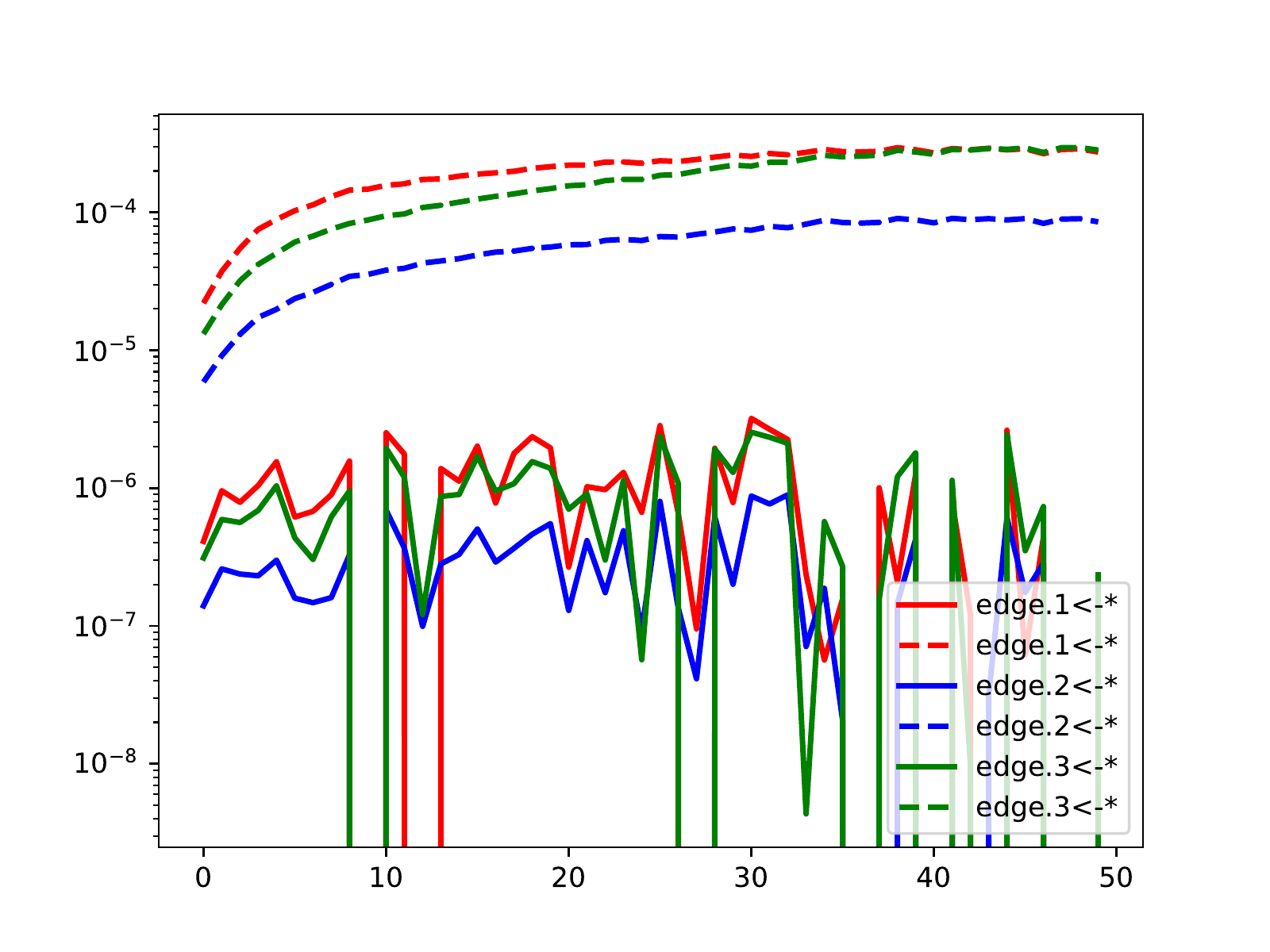}
\includegraphics[width=0.3\linewidth]{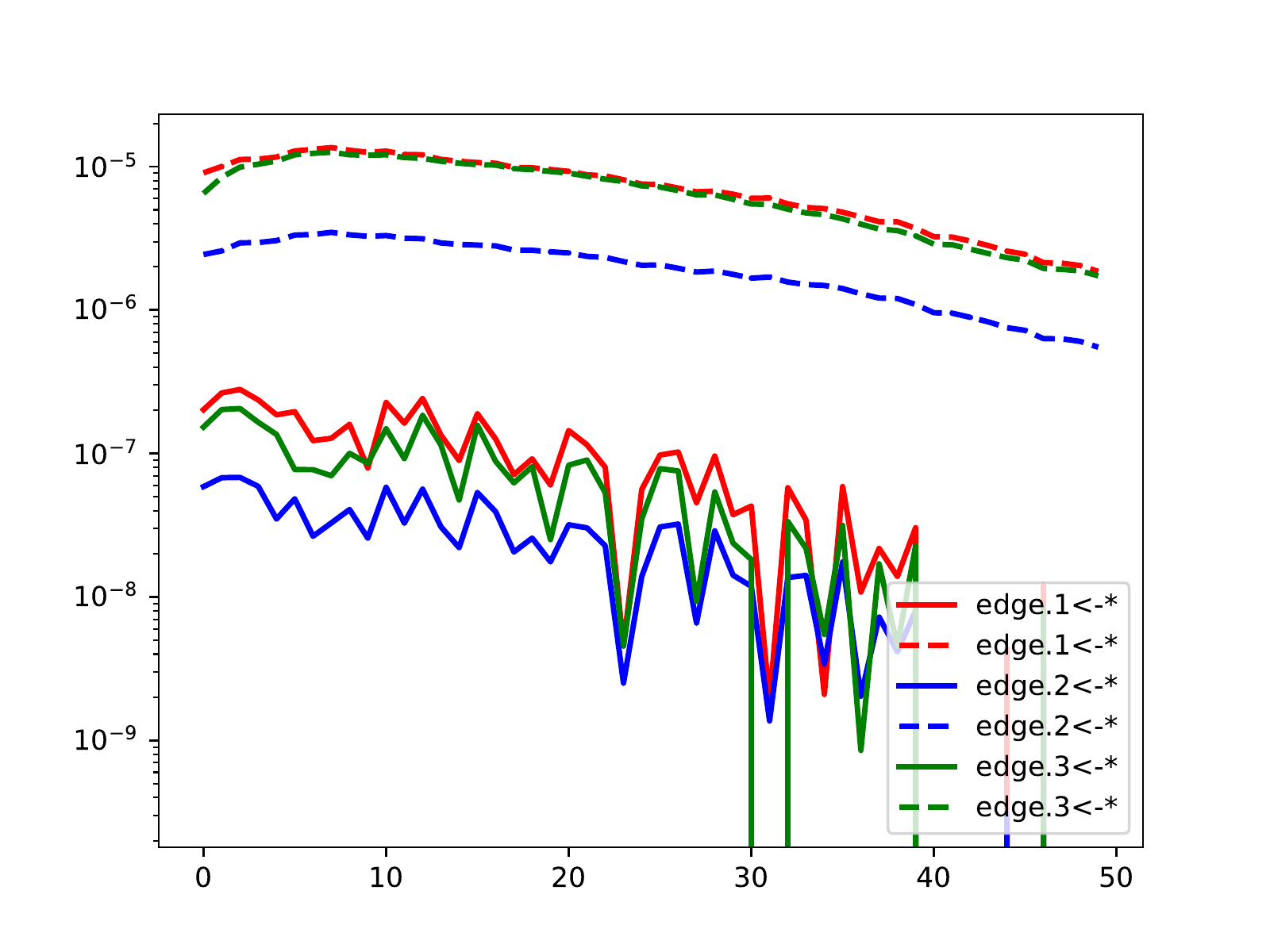}
\caption{Comparison of gradient correlation \ref{equ:corr} between bi-level and single-level optimization on Cifar10 dataset and in NAS-Benchmark-201 search space.  It shows the gradient correlation on different nodes of first, middle and last cells. Solid lines represent bi-level and dashed lines represent single-level. Gradient correlation in bi-level optimization is nearly zero and far smaller than in single-level optimization.} 
\label{fig:corr1}
\end{center}
\end{figure}

In last section, we analyze the gradient correlation between bi-level optimization and single-level optimization. Fig \ref{fig:corr1} compares that the gradients correlation \ref{equ:corr} in bi-level and single-level optimization in different cells. For bi-level optimization it's nearly zero and far less than that gradient-correlation in single-level.  In fact, even from the same dataset but not the same batch, their gradients are still irrelevant. Table \ref{tab:batch} compares the effect of whether backward propagations gradients of $w$ and $\alpha$ in the same batch. The results show that even the gradients are from the same dataset but from different batches,  the search progress collapses yet for the bias exists. The training phase lasts 50 epochs. $w$ and $\alpha$ is optimized with SGD optimizer,  learning rate is 0.005 (cosine scheduler). Original learning rate in \cite{dong2020bench} is 0.025, but it's too large and we show that it results badly in next section. 

\subsubsection{Analysis on Gradient Bias}
\begin{figure}[]
%\begin{center}
\centering
%\fbox{\rule[-.5cm]{0cm}{4cm} \rule[-.5cm]{4cm}{0cm}}
 \begin{subfigure}[b]{0.3\linewidth}
 \centering
\includegraphics[width=\textwidth]{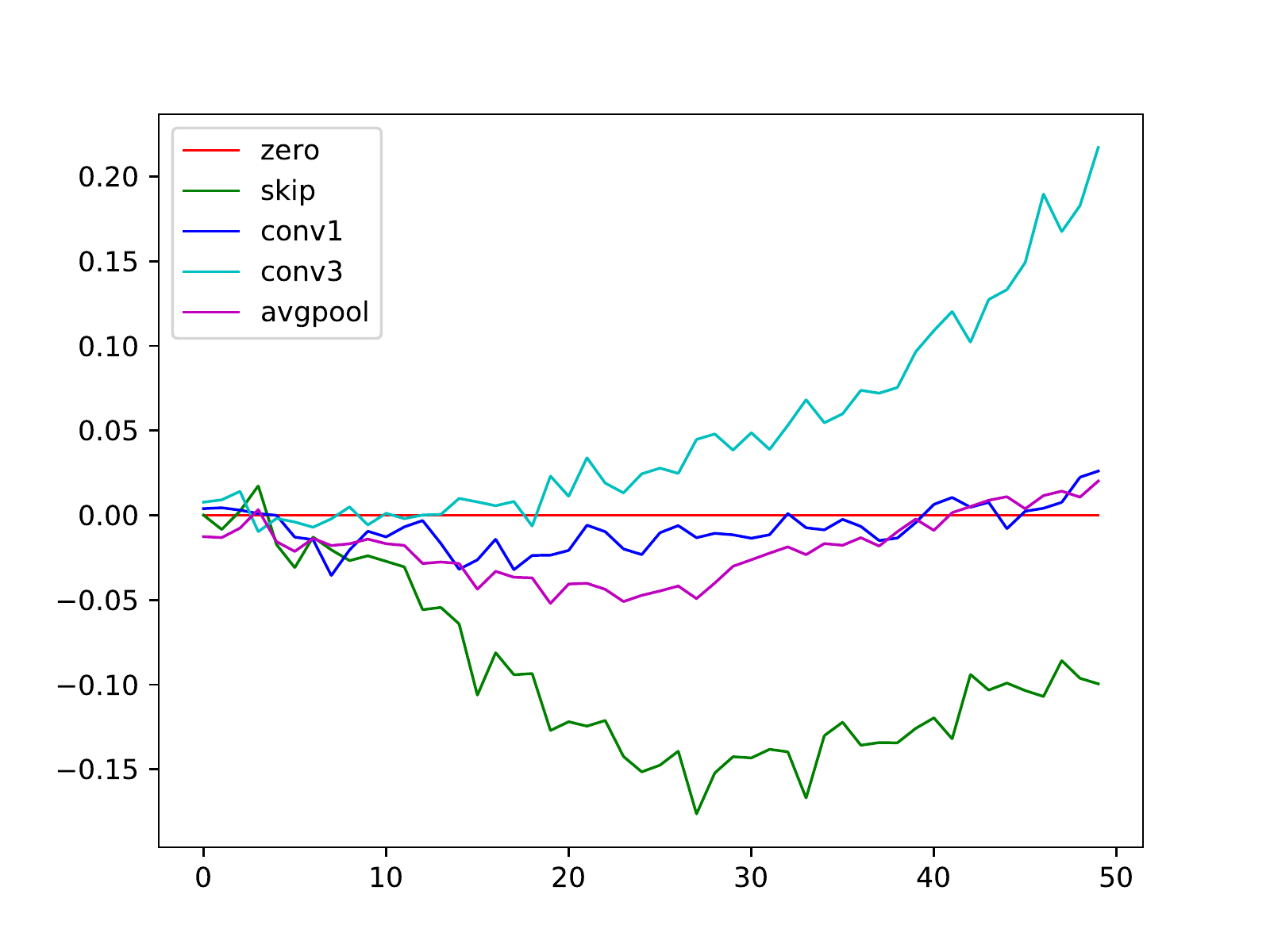}
 \caption{edge.3$\leftarrow$0}
\end{subfigure}
\hfill
 \begin{subfigure}[b]{0.3\linewidth}
 \centering
\includegraphics[width=\textwidth]{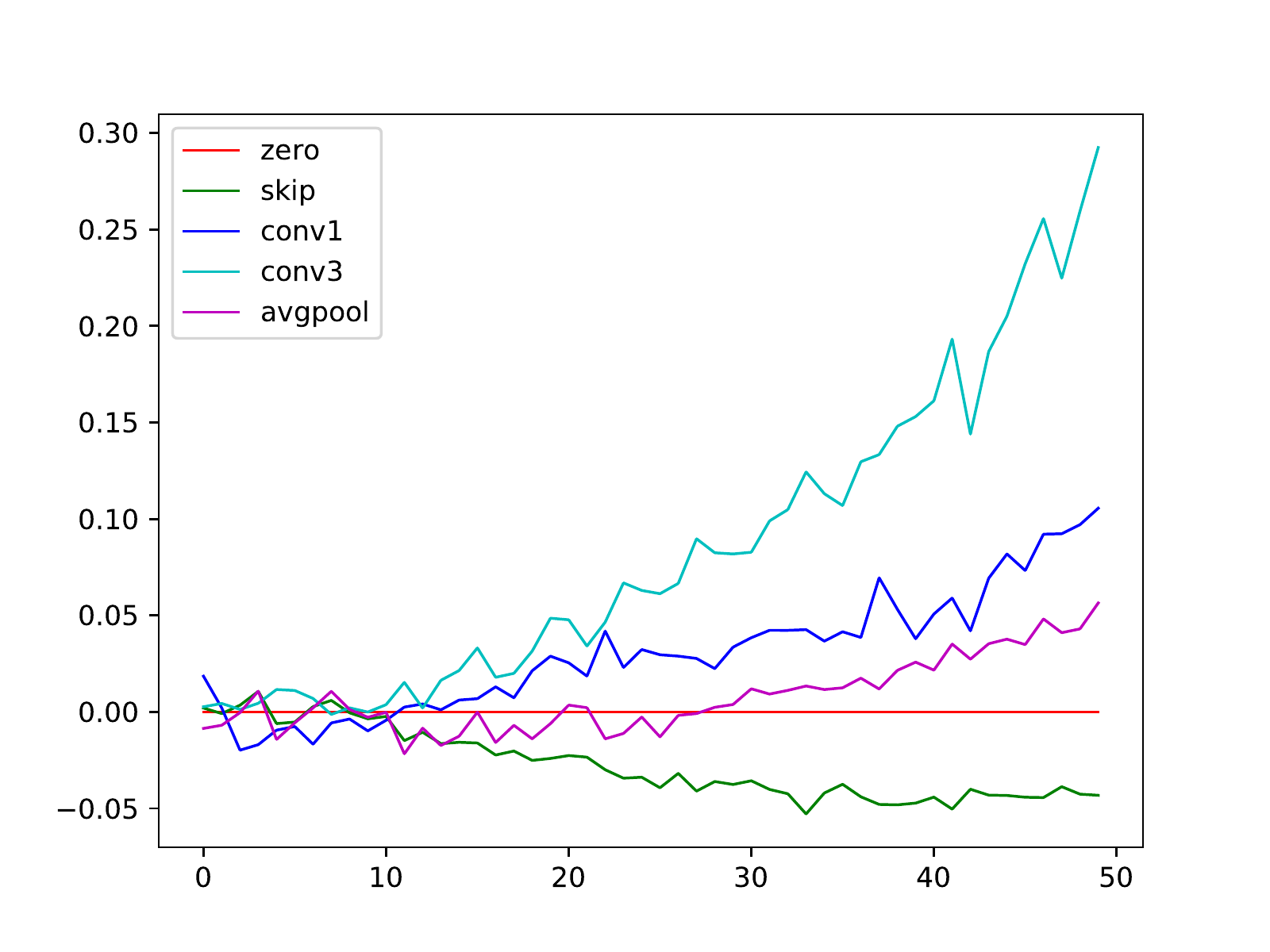}
 \caption{edge.3$\leftarrow$1}
\end{subfigure}
\hfill
 \begin{subfigure}[b]{0.3\linewidth}
 \centering
\includegraphics[width=\textwidth]{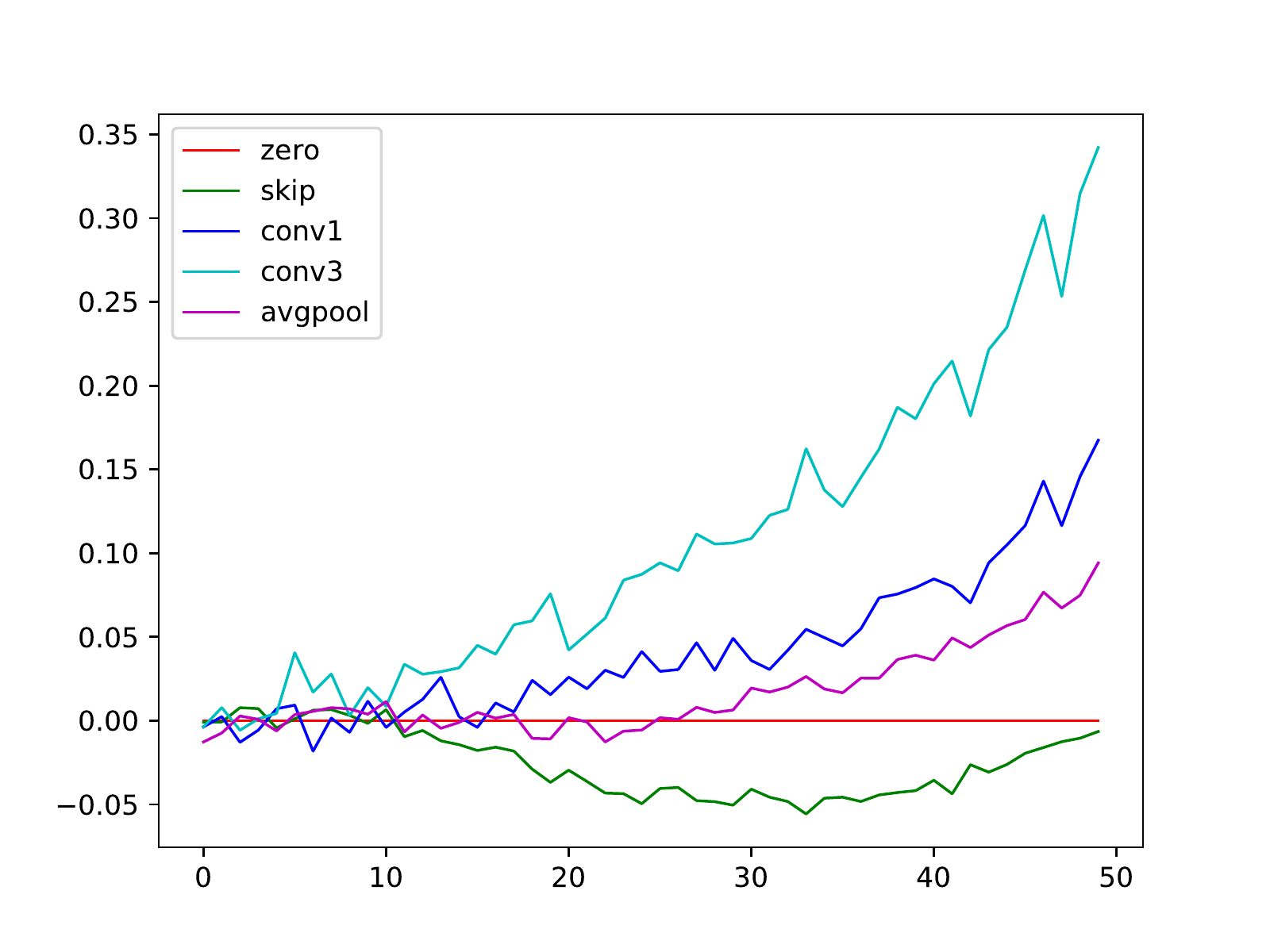}
 \caption{edge.3$\leftarrow$2}
\end{subfigure}
\hfill
\quad

\begin{subfigure}[b]{0.3\linewidth}
\centering
\includegraphics[width=\textwidth]{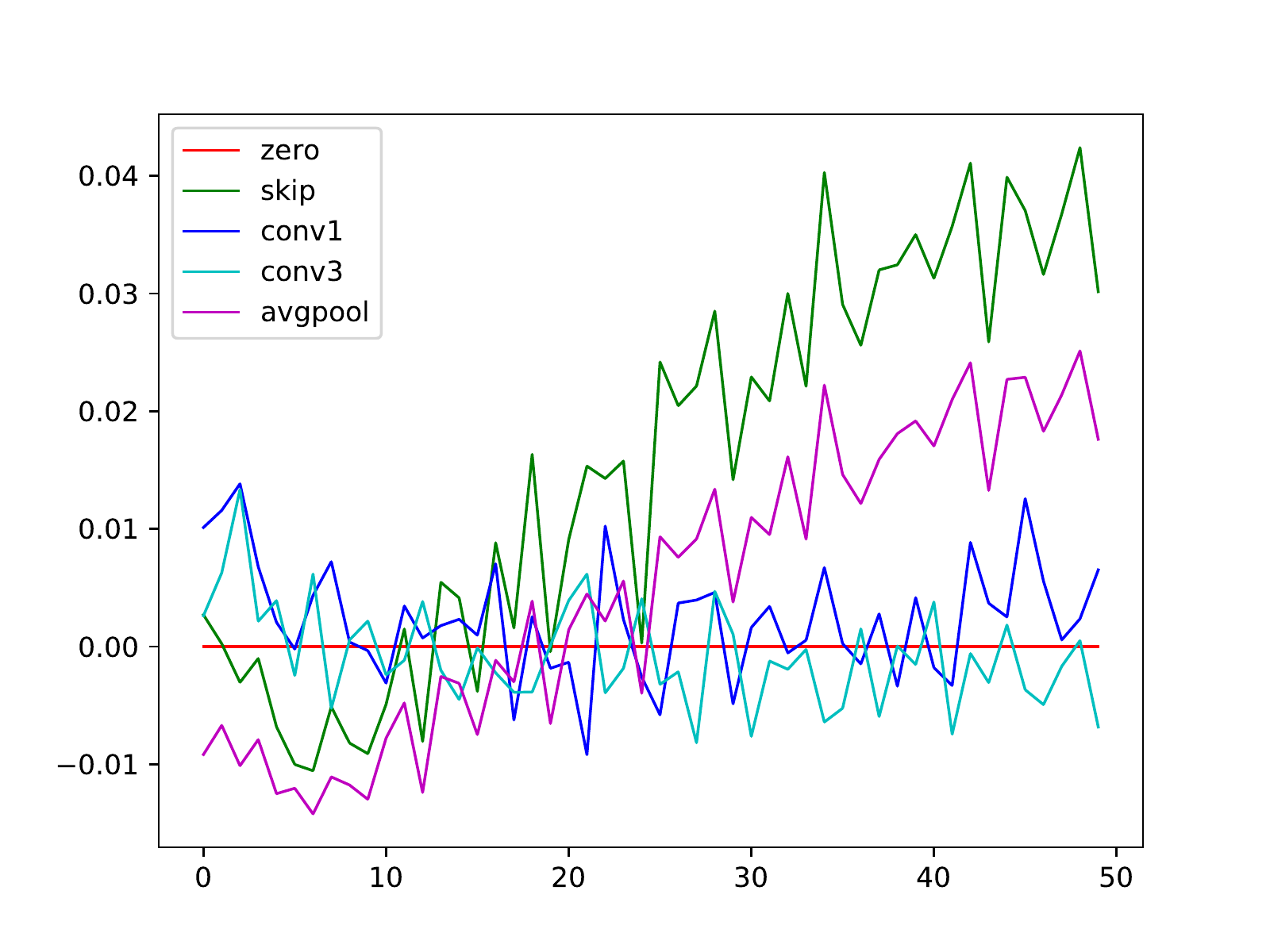}
\caption{edge.3$\leftarrow$0}
\end{subfigure}
\hfill
\begin{subfigure}[b]{0.3\linewidth}
\centering
\includegraphics[width=\textwidth]{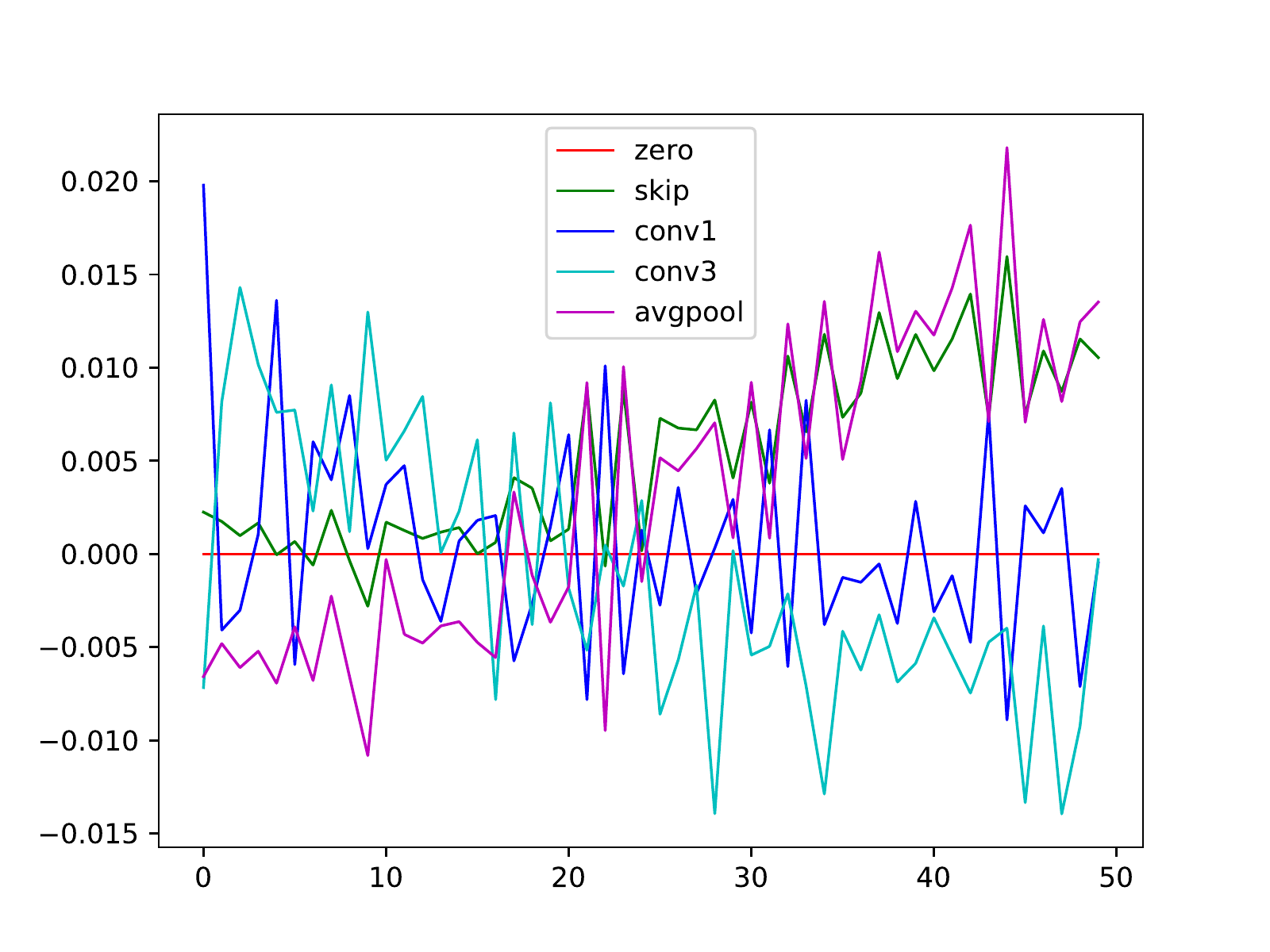}
 \caption{edge.3$\leftarrow$1}
\end{subfigure}
\hfill
 \begin{subfigure}[b]{0.3\linewidth}
 \centering
\includegraphics[width=\textwidth]{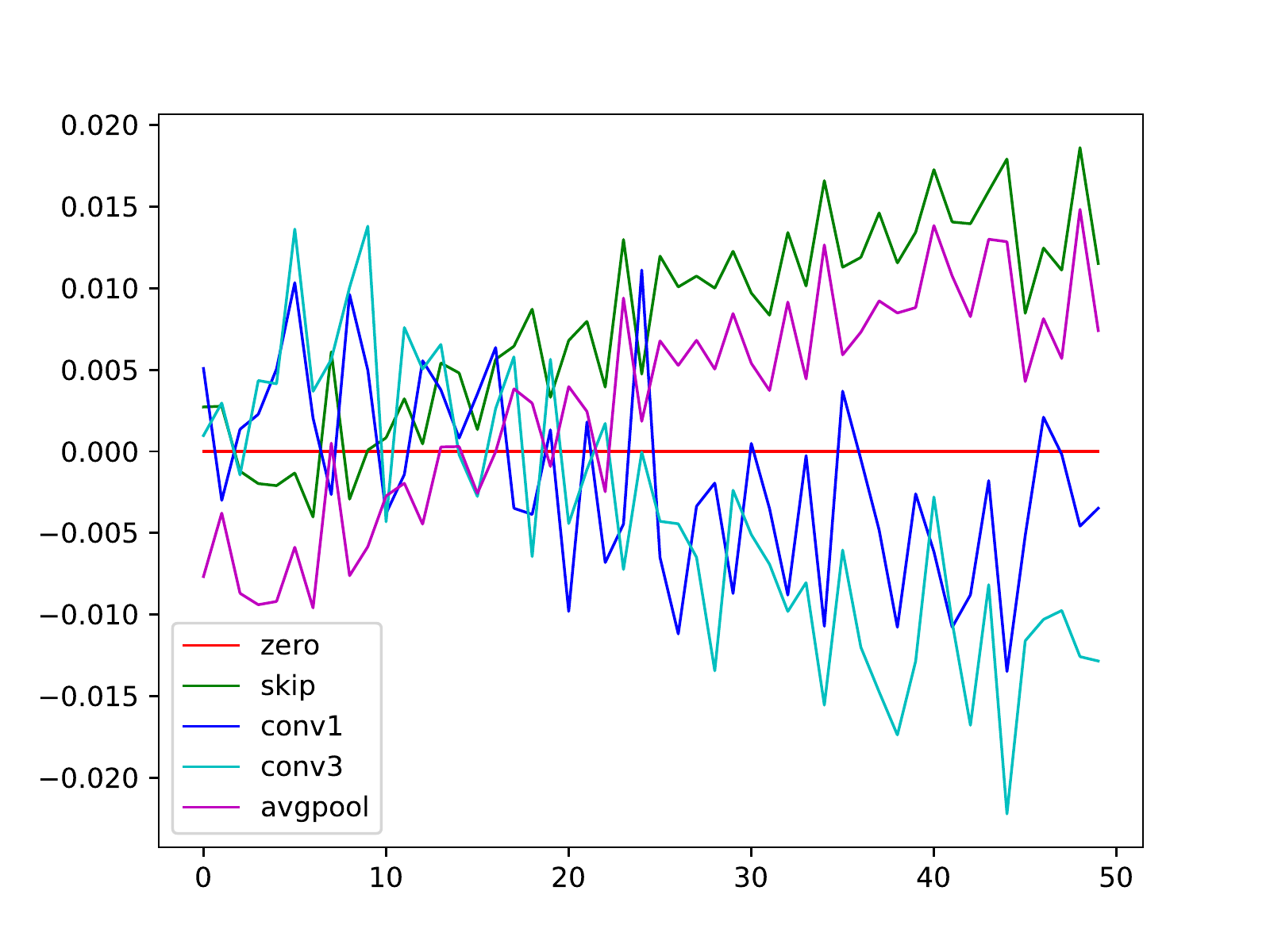}
 \caption{edge.3$\leftarrow$2}
\end{subfigure}
\hfill
\quad

\caption{Comparison of gradient of $p_i$ in the edges 3$\leftarrow$* in the 15th cell between bi-level and single-level optimization. 
Figures (a)(b)(c) are on bi-level optimization and (d)(e)(f) are on single-level optimization.
%Fig (a)(b) is on bi-level optimization with softmax and sigmoid as activation function. 
%Fig (c)(d) is on single-level optimization with softmax and sigmoid as activation function. 
} 
\label{fig:grad_pi}
\end{figure}

We compare $\pder{\Ls}{p_i}$ in the last cell between bi-level and single-level optimization in Fig \ref{fig:grad_pi}. For bi-level optimization, due to gradient bias, $\pder{\Ls}{p_i}$ of non-learnable operations are smaller than learnable operations and the gap is increased as the training process goes. Thus non-learnable operations' architecture parameter would be much larger than learnable operations'. For single-level optimization, $\pder{\Ls}{p_i}$ of learnable operations compete against non-learnable operations but finally get dominated. More comparison of different cells are shown in Appendix. 

\begin{table*}[t]
\caption{Comparison the effect of whether gradients are calculated on the same data. Each setting is run 3 independent times to get average values. It shows that even on the same dataset but not the same mini batch, DARTS could not search good architectures for different batches are also low correlated. The experiments are under softmax as activation function. For learning rate damages search phase (it would be discussed bellow), the analysis is based on learning rate set as 0.005.}
\label{tab:batch}
\begin{center}
\begin{tabular}{|l|l|l|l|l|}
\hline
 \textbf{same dataset} & \textbf{same batch} & \textbf{CIFAR10} & \textbf{CIFAR100} & \textbf{ImageNet16-120} \\ \hline
False                 & False               &61.88$\pm$10.72 &25.13$\pm$13.47 &17.98$\pm$2.35             \\ \hline
True                  & False               & 84.34$\pm$0.02       & 54.92$\pm$0.06        & 25.97$\pm$0.49              \\ \hline
True                  & True                &94.27$\pm$0.13 &73.01$\pm$0.71 &46.10$\pm$0.34              \\ \hline
\end{tabular}
\end{center}
\end{table*}

\subsubsection{Analysis on Activation Function and Learning Rate}
%we analysis the margin on \ref{tab:margin}.  The margin of bi-level optimization is bigger then single-level optimization. Then we compare the learning rate effect on convergence rate on table \ref{tab:lr}. 
We compare the effects of learning rate and activation function on the searched results. Table \ref{tab:lr_act} shows that larger learning rate would cause search phase collapsed, especially in the bi-level optimization framework. In the single-level optimization, the search phase performs more steadily. But too large learning rate also has bad effects. In addition, using sigmoid performs more steadily than softmax under different learning rate and optimization level. 
In this section, for the convenience of analysis, we use SGD as network weights' and architecture parameters' optimizer. 
In general NAS tasks, we advice to use SGD as network weights' optimizer and Adam as architecture parameters' optimizer. We could adjust network parameters' learning rate according to the searched non-parametric operations' ratio. If it's over a threshold such as 0.3, we should decrease it. And the process commonly doesn't cost too much time.

\begin{table}[]
\caption{Comparison of effects of different activation function and learning rate. Each setting is run 3 independent times to get average values. Larger learning rate would cause search phase collapsed, especially in the bi-level optimization framework. In the single-level optimization, the search phase performs more steadily. And using sigmoid performs more steadily. }
\label{tab:lr_act}
\begin{center}
\begin{tabular}{|l|l|l|l|l|}
\hline
\textbf{optim} & \textbf{activation} & \textbf{lr} & \textbf{CIFAR10} & \textbf{CIFAR100}  \\ \hline
bi &softmax &0.001 &91.54$\pm$1.26 &68.21$\pm$1.15  \\ \hline
bi &softmax &0.005 &61.88$\pm$10.72 &25.13$\pm$13.47  \\ \hline
bi &softmax &0.025 &84.97$\pm$1.03 &56.01$\pm$1.71  \\ \hline
bi &sigmoid &0.001 &92.53$\pm$1.62 &69.62$\pm$2.05  \\ \hline
bi &sigmoid &0.005 &80.57$\pm$0.00 &47.93$\pm$0.00  \\ \hline
bi &sigmoid &0.025 &74.27$\pm$14.12 &41.86$\pm$18.56  \\ \hline
single &softmax &0.001 &94.36$\pm$0.00 &73.51$\pm$0.00  \\ \hline
single &softmax &0.005 &94.27$\pm$0.13 &73.01$\pm$0.71  \\ \hline
single &softmax &0.025 &86.58$\pm$0.00 &58.33$\pm$0.00  \\ \hline
single &sigmoid &0.001 &94.36$\pm$0.00 &73.51$\pm$0.00 \\ \hline
single &sigmoid &0.005 &94.36$\pm$0.00 &73.51$\pm$0.00 \\ \hline
single &sigmoid &0.025 &93.10$\pm$0.00 &69.24$\pm$0.00 \\ \hline
\end{tabular}
\end{center}
\end{table}

\subsection{Comparison between algorithms}
\subsubsection{On NAS-201}
At last, we compare our improvement with other algorithms. We use single-level to replace bi-level optimization, use sigmoid to replace softmax and $\alpha$ initialized as $-\ln(n-1)$. It shows single-level optimization could steadily get the SOTA and nearly the optimal architecture in NAS-201 space. On search phase, we train the super model for 50 epochs, using SGD optimizer with a momentum of 0.9, cosine scheduler, batch size as 64, learning rate as 0.005, weight decay as $3\times10^{-4}$. For architecture parameters, we use Adam optimizer with a fixed learning rate of $3\times10^{-4}$, a momentum (0.5, 0.999). $\alpha$ is initialized as $-\ln(4)$ such that $sigmoid(\alpha)=0.2$. And weight decay is set 0 to avoid extra gradients on zero-op.  Results \ref{tbl:nas201} show single-level optimization search steadily near the best architecture in NAS-201-Benchmark. 
\begin{table*}[t]
\caption{Comparison of different NAS algorithms on NAS-Bench-201. }
\begin{center}
\begin{tabular}{|l|l|l|l|l|l|l|}
\hline
\multicolumn{1}{|c|}{\multirow{2}{*}{Method}} & \multicolumn{2}{c|}{CIFAR-10}                               & \multicolumn{2}{c|}{CIFAR-100}                              & \multicolumn{2}{c|}{ImageNet-16-120}                        \\ \cline{2-7} 
\multicolumn{1}{|c|}{}                        & \multicolumn{1}{c|}{validation} & \multicolumn{1}{c|}{test} & \multicolumn{1}{c|}{validation} & \multicolumn{1}{c|}{test} & \multicolumn{1}{c|}{validation} & \multicolumn{1}{c|}{test} \\ \hline
RSPS\cite{li2020random}                                          & 84.16$\pm$1.69                      & 87.66$\pm$1.69                & 59.00$\pm$4.60                      & 58.33$\pm$4.34                & 31.56$\pm$3.28                      & 31.14$\pm$3.88                \\ \hline
DARTS\cite{liu2018darts}                                         & 39.77$\pm$0.00                      & 54.30$\pm$0.00                & 15.03$\pm$0.00                      & 15.61$\pm$0.00                & 16.43$\pm$0.00                      & 16.32$\pm$0.00                \\ \hline
GDAS\cite{dong2019searching}                                          & 90.00$\pm$0.21                      & 93.51$\pm$0.13                & 71.14$\pm$0.27                      & 70.61$\pm$0.26              & 41.70$\pm$1.26                      & 41.84$\pm$0.90                \\ \hline
SETN \cite{dong2019one}                                         & 82.25$\pm$5.17                      & 86.19$\pm$4.63                & 56.86$\pm$7.59                      & 56.87$\pm$7.77                & 32.54$\pm$3.63                      & 31.90$\pm$4.07                \\ \hline
ENAS \cite{pham2018efficient}                                         & 39.77$\pm$0.00                      & 54.30$\pm$0.00                & 15.03$\pm$0.00                      & 15.61$\pm$0.00                & 16.43$\pm$0.00                      & 16.32$\pm$0.00                \\ \hline
CDARTS \cite{yu2020cyclic}    &91.13$\pm$0.44   &94.02$\pm$0.31     &72.12$\pm$1.23  &71.92 $\pm$1.30    &{\bf 45.09$\pm$0.61}  & 45.51$\pm$0.72 \\ \hline
DARTS-  \cite{chu2020darts} &91.03$\pm$0.44   &93.80$\pm$0.40    &71.36$\pm$1.51   &71.53$\pm$1.51   &44.87$\pm$1.46    & 45.12$\pm$0.82 \\ \hline
single-level+sigmoid   &{\bf91.55$\pm$0.00}   & {\bf 94.36$\pm$0.00}   &{\bf 73.49$\pm$0.00}  &{\bf 73.51$\pm$0.00}   & 44.87$\pm$0.00  &{\bf 46.34$\pm$0.00} \\  
\hline
optimal   &91.61      & 94.37      &73.49        &73.51         &46.77   &47.31  \\  
\hline
\end{tabular}
\end{center}
\label{tbl:nas201}
\end{table*}

\subsubsection{On DARTS}
\label{darts}
DARTS \cite{dong2020bench} is also a cell-based search space. Each cell contains 6 nodes and each node has to select 2 edges to connect with previous 2 nodes. Each edge has 8 operations: $3\times3$ and $5\times5$ separable convolution, $3\times3$ and $5\times5$ dilated separable convolution, $3\times3$ max-pooling, $3\times3$ average-pooling, skip-connect (identity), and zero (none). And the stacked networks have normal cells and  reduction cells. The search space covers  $10^{18}$ architectures which is quite large.\\
\begin{table}[h!]
  \caption{Comparison with SOTA architectures on ImageNet in DARTS space. $\dagger$ denotes the average results of searched architectures (not the average results of retraining searched best architecture) and  $*$ denotes the best result. For single-level optimization and DARTS(ours), we search 3 times directly on the ImageNet-1K. Different from PDARTS training setting that re-train the searched architecture for 250 epochs,  DropNAS train it for 600 epochs and DARTS+ 800 epochs.}
  \label{table:darts-imagenet}
  \centering
  \begin{tabular}{lccc}
  	\toprule
    \textbf{Architecture} &  \textbf{FLOPs}
 & \textbf{Params} & \textbf{Top-1.}   \\
 %\multirow{2}{8em}{\textbf{Transfer Learning}} \\
  & (M)&  (M)  &  (\%) \\
    \midrule[1.5pt]
    NASNet-A (\cite{zoph2018learning}) & 564 & 5.3 & 74.0 \\
    %NASNet-B (\cite{zoph2018learning}) &  72.8 & 91.3  & 564M & \cmark \\
    %NASNet-C (\cite{zoph2018learning}) &  72.5 & 91.1  & 564M & \cmark \\
    AmoebaNet-C (\cite{real2019regularized}) &570 & 6.4 & 75.7   \\
    PDARTS (\cite{chen2019progressive}) & 557 & 4.9 &75.6 \\
    PC-DARTS (\cite{xu2019pc}) & 597 & 5.3 & 75.8  \\ 
    DARTS (\cite{liu2018darts}) & 574 & 4.7 & 73.3  \\
    DARTS(ours)*          & 677 &5.93 &74.6 \\
    DARTS(ours)$\dagger$    & 629 &5.51 & 74.0 \\
    DARTS+(\cite{liang2019darts}) & 591 & 5.1 & 76.3  \\
    CDARTS$\dagger$(\cite{yu2020cyclic}) &732 & 6.1 & 76.3  \\
    CDARTS*(\cite{yu2020cyclic}) &704 & 6.3 & 76.6  \\    
    DropNAS(\cite{hong2020dropnas} & 597 & 5.4 & 76.6 \\
    \midrule[1.5pt]
    Single-level+softmax* & 706 & 6.56 & {\bf 76.7}  \\
    Single-level+softmax$\dagger$ &710 &6.58 & 76.4  \\ 
    Single-level+sigmoid* &714 & 6.60 & {\bf 77.0}  \\
    Single-level+sigmoid$\dagger$ &707 &6.55 & 76.7  \\
    \midrule[1.5pt]
  \end{tabular}
\end{table}
We directly do search process on the ImgeNet-1k dataset. Specifically, input images are downsampled third times by convolution layers at the beginning of the super model to reduce spatial resolution which follows \cite{unnas}. On search phase, we train the super model for 50 epochs, using SGD optimizer with a momentum of 0.9, cosine scheduler, batch size as 360, learning rate as 0.025, weight decay as $3\times10^{-4}$. For architecture parameters, we use Adam optimizer with a fixed learning rate of $3\times10^{-4}$, a momentum (0.5, 0.999).  For softmax as architecture parameters' activation function, $\alpha$ is initialized with $Gaussian(0, 10^{-3})$ and weight decay is $10^{-3}$. For sigmoid, $\alpha$ is initialized as $-\ln(7)$ such that $sigmoid(\alpha)=0.125$. And weight decay is set 0 to avoid extra gradients on zero-op. The search phase costs 4 GPUs for about 28 hours on NVIDIA GeForce RTX 2080ti. On retrain phase, we train the searched architecture following fully PDARTS setting and without any additional module. It lasts 250 epochs, using SGD optimizer with a momentum of 0.9, batch size as 768, an initial learning rate of 0.375 (decayed down to zero linearly), and a weight decay of $3\times10^{-5}$. Additional enhancements are adopted including label smoothing and an auxiliary loss tower during training as in PDARTS. Learning rate warm-up is applied for the first 5 epochs. The retrain phase costs 8 GPUs for about 3.5 days on NVIDIA GeForce RTX 2080ti.
We search the state-of-the-art architecture with 77.0\% top1 accuracy (training setting follows PDARTS and without any additional module) on ImageNet-1K and steadily search architectures up-to 76.5\% top1 accuracy. More details are shown in Appendix.

\section{conclusion}
In this paper, we do analysis on the bi-level optimization in DARTS. Experiments show that bi-level optimization could cause gradients bias of architecture parameters and expand the gap between learnable and non-learnable operations. Furthermore, architecture parameters' activation function and large learning rate would aggregate the gap. We give a theoretical explanation and conduct experiments to verify it. As a result, we propose to use single-level optimization as an instead. And we use uncompetitive activation function like sigmoid to replace softmax. Meanwhile, we do normalization for the initialization of sigmoid. Experiments on NAS-Benchmark-201 and DARTS space show that single-level optimization could steadily find out high-performance architectures. 
\\

{\small
\bibliographystyle{ieee_fullname}
\bibliography{single_level}
}

\appendix
%\title{Appendix}
%\maketitle
\section{Proof of theorem 3.1}
%\textbf{Proof of Theorem 3.1}
\begin{proof}
It's easily to see that $p_i \geq p_j$ if $\alpha_i \geq \alpha_j$ and $p_{i^*} \geq \frac{1}{n}$ for $\sum_i p_i = 1$.
Consider
\begin{equation}
\begin{split}
\frac{\partial l}{\partial\alpha_i} &= \frac{\partial l}{\partial\vz}\frac{\partial\vz}{\partial{\alpha_i}} \\
			&= \frac{\partial l}{\partial\vz}(p_i(1-p_i)\vz_i - p_i\sum_{j\neq i}p_j\vz_j) \\
			&= \frac{\partial l}{\partial\vz}\sum_{j} p_i p_j (\vz_i - \vz_j) \\
			&= p_i \sum_{j} p_j\frac{\partial l}{\partial\vz} (\vz_i-\vz_j) \\
\end{split}
\end{equation}
%For $\alpha_{i^*}$, it holds that $\frac{\partial l}{\partial\alpha_{i^*}} \geq \delta p_{i^*}$. And due to $\sum_i{p_i} = 1
%$, thus $p_{i*} \geq p_i$
For $i^*$ and $\forall i \neq i^*$, because $\frac{\partial l}{\partial\vz}(\vz_{i^*}-\vz_j) \geq \frac{\partial l}{\partial\vz}(\vz_{i}-\vz_j)$ and $p_{i^*} \geq p_i \geq 0$, thus $\frac{\partial l}{\partial\alpha_{i^*}} \geq \frac{\partial l}{\partial\alpha_i}$.\\
As a result
\begin{equation}
\begin{split}
%\begin{center}
&\frac{\partial l}{\partial\alpha_{i^*}} - \frac{\partial l}{\partial\alpha_{i}}\\
&= p_{i^*} \sum_{j} p_j\frac{\partial l}{\partial\vz} (\vz_{i^*}-\vz_j) - p_i \sum_{j} p_j\frac{\partial l}{\partial\vz} (\vz_i-\vz_j) \\
&= (p_{i^*} - p_i) \sum_{j} p_j\frac{\partial l}{\partial\vz} (\vz_{i^*}-\vz_j) + p_i (\sum_{j} p_j\frac{\partial l}{\partial\vz} (\vz_{i^*}-\vz_j) \\
&- \sum_{j} p_j\frac{\partial l}{\partial\vz} (\vz_i-\vz_j)) \\
&= (p_{i^*} - p_i) \sum_{j} p_j\frac{\partial l}{\partial\vz} (\vz_{i^*}-\vz_j) + p_i\sum_{j} p_j\frac{\partial l}{\partial\vz} (\vz_{i^*}-\vz_i) \\
&\geq (p_{i^*} - p_i) \delta + p_i \delta \\
&= p_{i^*} \delta \\
&\geq \frac{\delta}{n} \\
\end{split}
\end{equation}
Under gradient ascent, on update step $t$ we have, for any $i \neq i^*$,
\begin{equation}
\begin{split}
\alpha_{i^*}^t - \alpha_{i}^t &= \alpha_{i^*}^{t-1} - \alpha_{i}^{t-1} + \eta(\frac{dl_{t-1}}{d\alpha_{i^*}^{t-1}} - \frac{dl_{t-1}}{d\alpha_i^{t-1}}) \\
&\geq \alpha_{i^*}^{t-1} - \alpha_{i}^{t-1} + \eta p_{i^*}^t\delta_t \\
& \geq \alpha_{i^*}^{0} - \alpha_{i}^{0} + \eta \sum_t p_{i^*}^t\delta_t \\
& \geq \frac{\eta t \delta}{n}  \\
\end{split}
\end{equation}
When $t=\frac{n \ln((1-\epsilon)n )}{\eta \delta}$, we have \\
\begin{equation}
	\alpha_{i^*} - \alpha_i \geq \ln((1-\epsilon)n)
\end{equation}
Thus
\begin{equation}
 p_{i^*} = \frac{1}{\sum_i \exp({\alpha_i - \alpha_{i^*}})} \geq \frac{1}{\sum_i \exp(-\ln((1-\epsilon)n))} = 1 - \epsilon \\
\end{equation}
Under gradient descent, for $i^*=\argmin_i \frac{\partial l}{\partial\vz}\vz_i$, we have the same conclusion.
\end{proof}

\section{More comparison of gradient of $p_i$ in section 4.2.3}
\begin{figure}[h]
%\begin{center}
\centering
%\fbox{\rule[-.5cm]{0cm}{4cm} \rule[-.5cm]{4cm}{0cm}}
 \begin{subfigure}[b]{0.3\linewidth}
 \centering
\includegraphics[width=\textwidth]{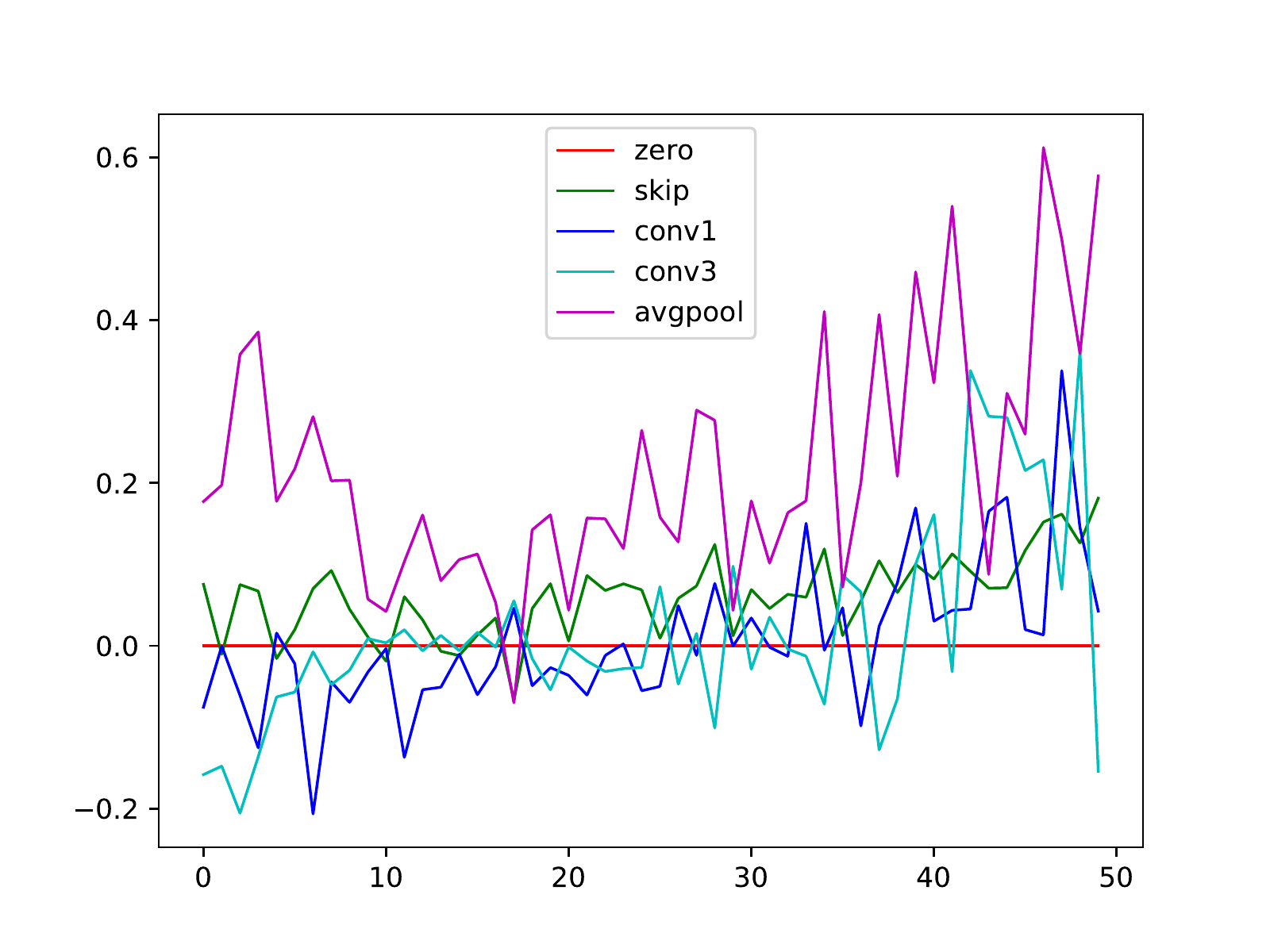}
 \caption{edge.1$\leftarrow$0}
\end{subfigure}
\hfill
 \begin{subfigure}[b]{0.3\linewidth}
 \centering
\includegraphics[width=\textwidth]{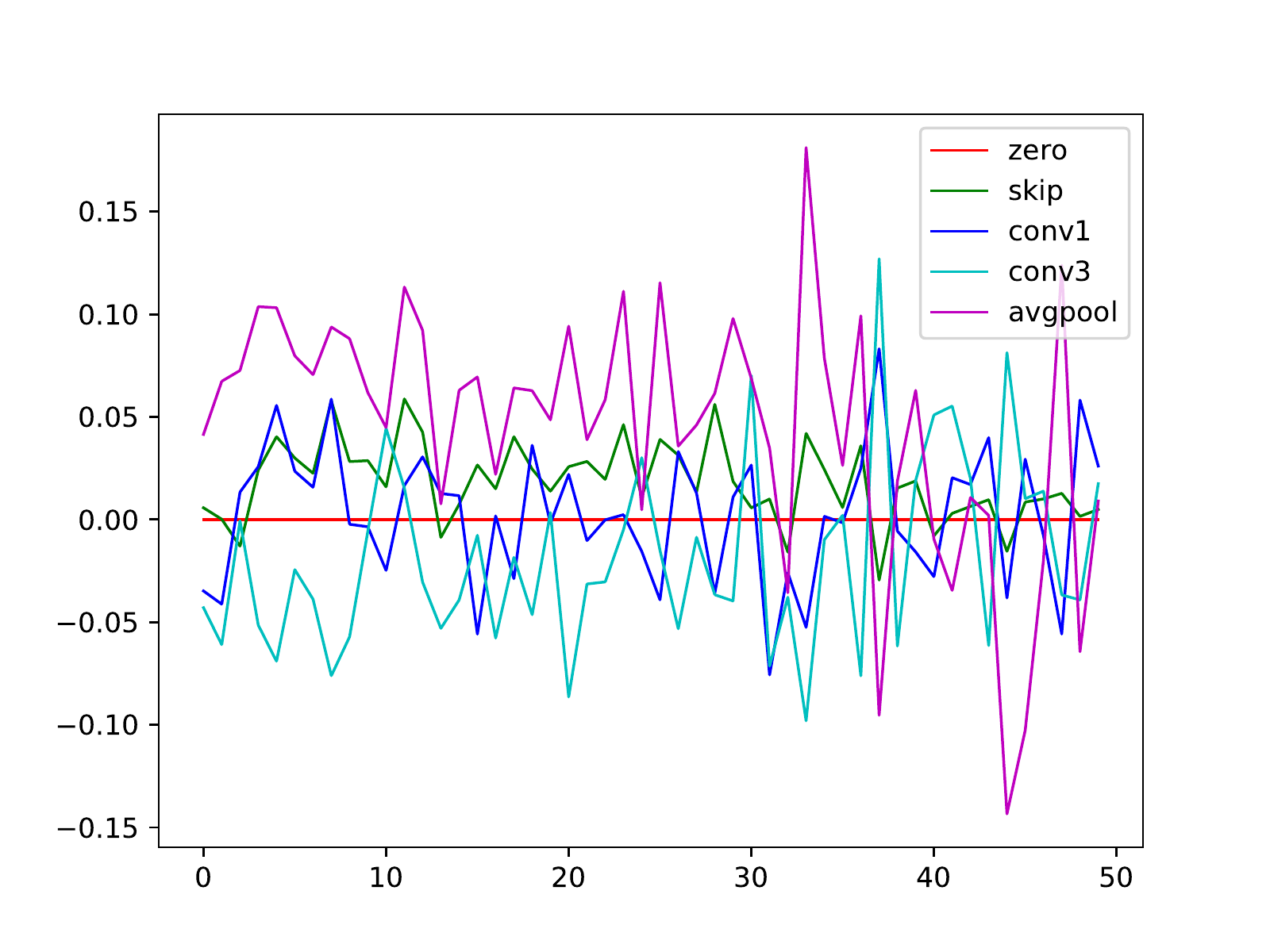}
 \caption{edge.2$\leftarrow$0}
\end{subfigure}
\hfill
 \begin{subfigure}[b]{0.3\linewidth}
 \centering
\includegraphics[width=\textwidth]{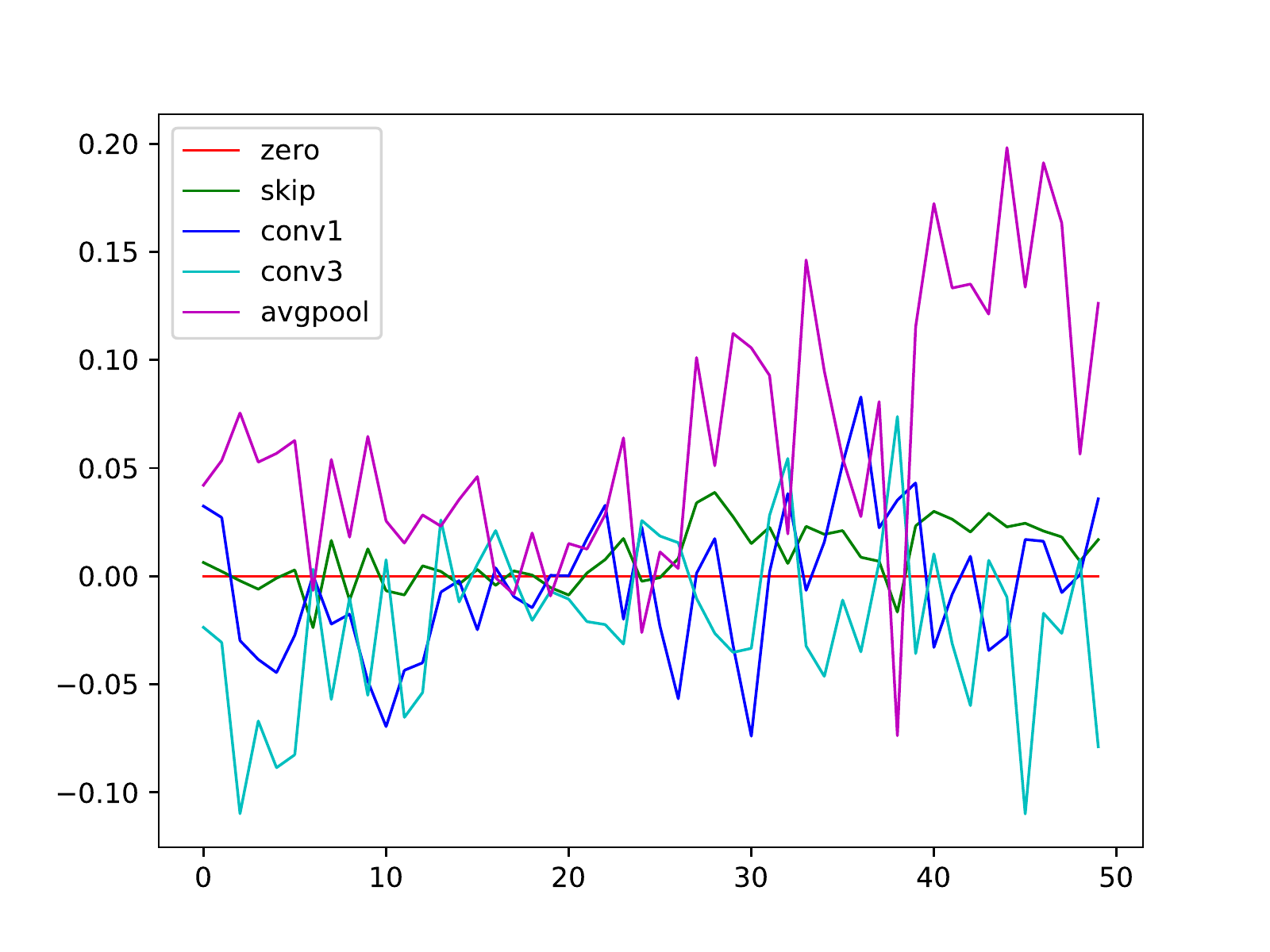}
 \caption{edge.2$\leftarrow$1}
\end{subfigure}
\hfill
\quad
 \begin{subfigure}[b]{0.3\linewidth}
 \centering
\includegraphics[width=\textwidth]{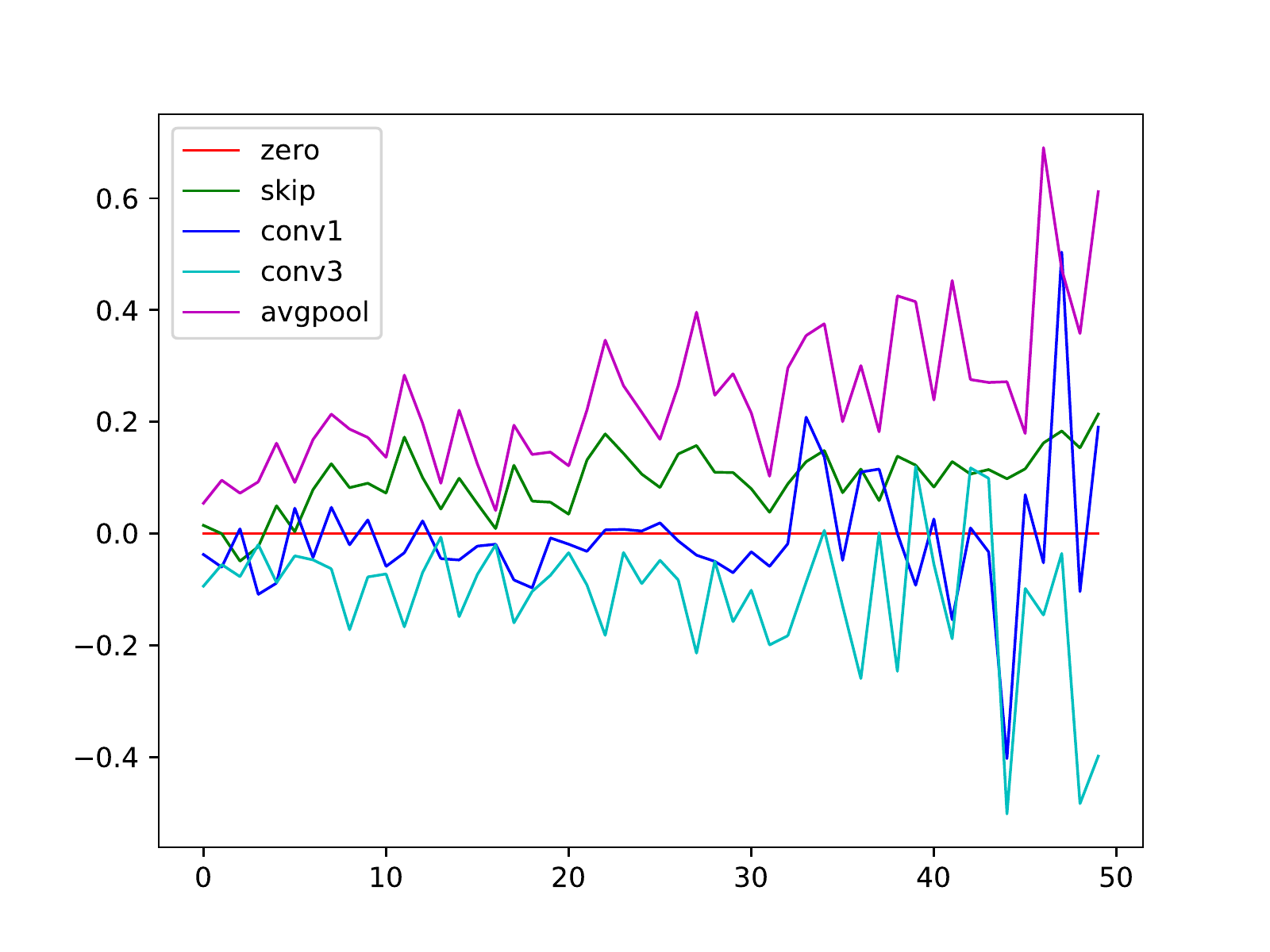}
 \caption{edge.3$\leftarrow$0}
\end{subfigure}
\hfill
 \begin{subfigure}[b]{0.3\linewidth}
 \centering
\includegraphics[width=\textwidth]{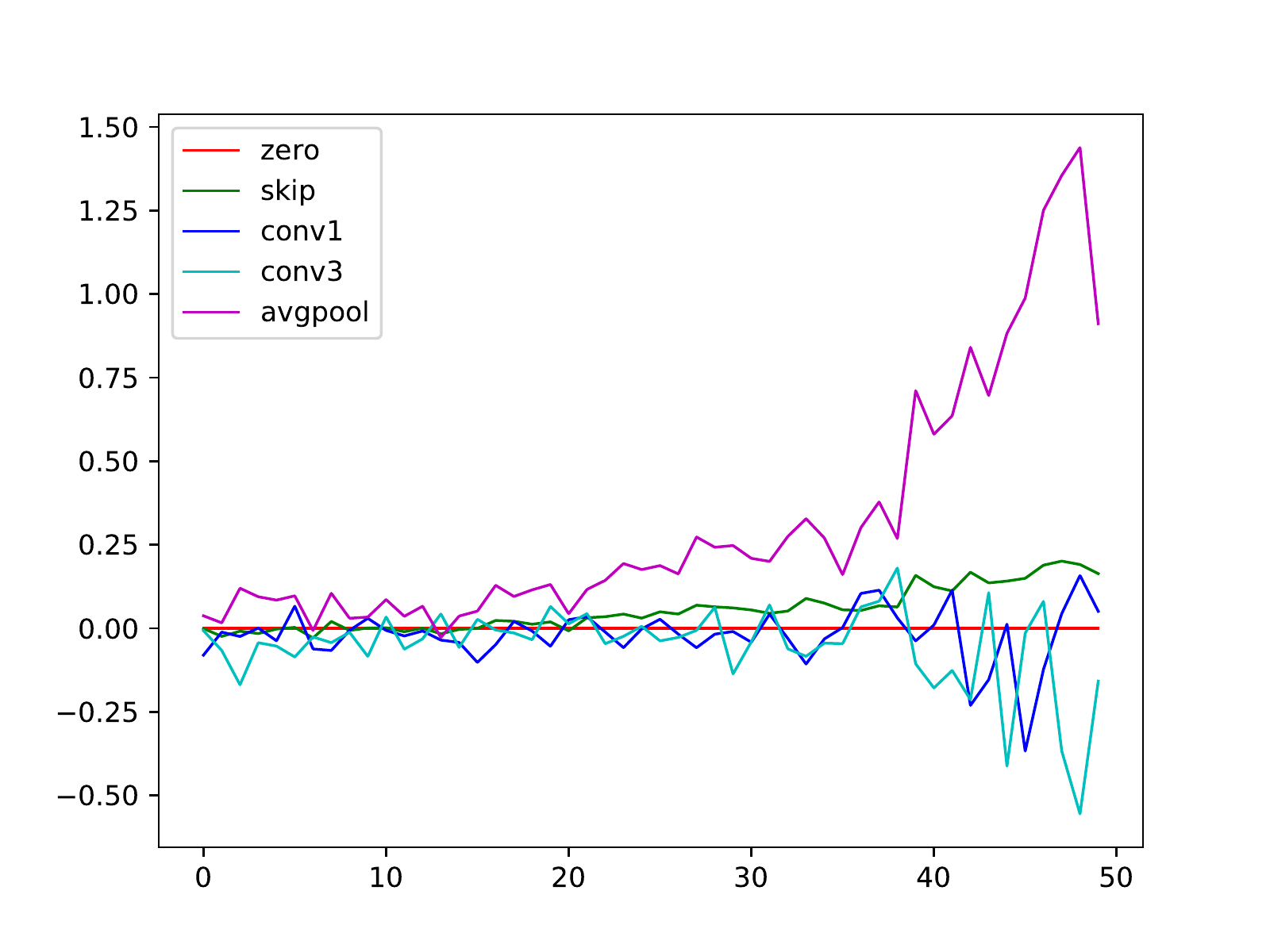}
 \caption{edge.3$\leftarrow$1}
\end{subfigure}
\hfill
 \begin{subfigure}[b]{0.3\linewidth}
 \centering
\includegraphics[width=\textwidth]{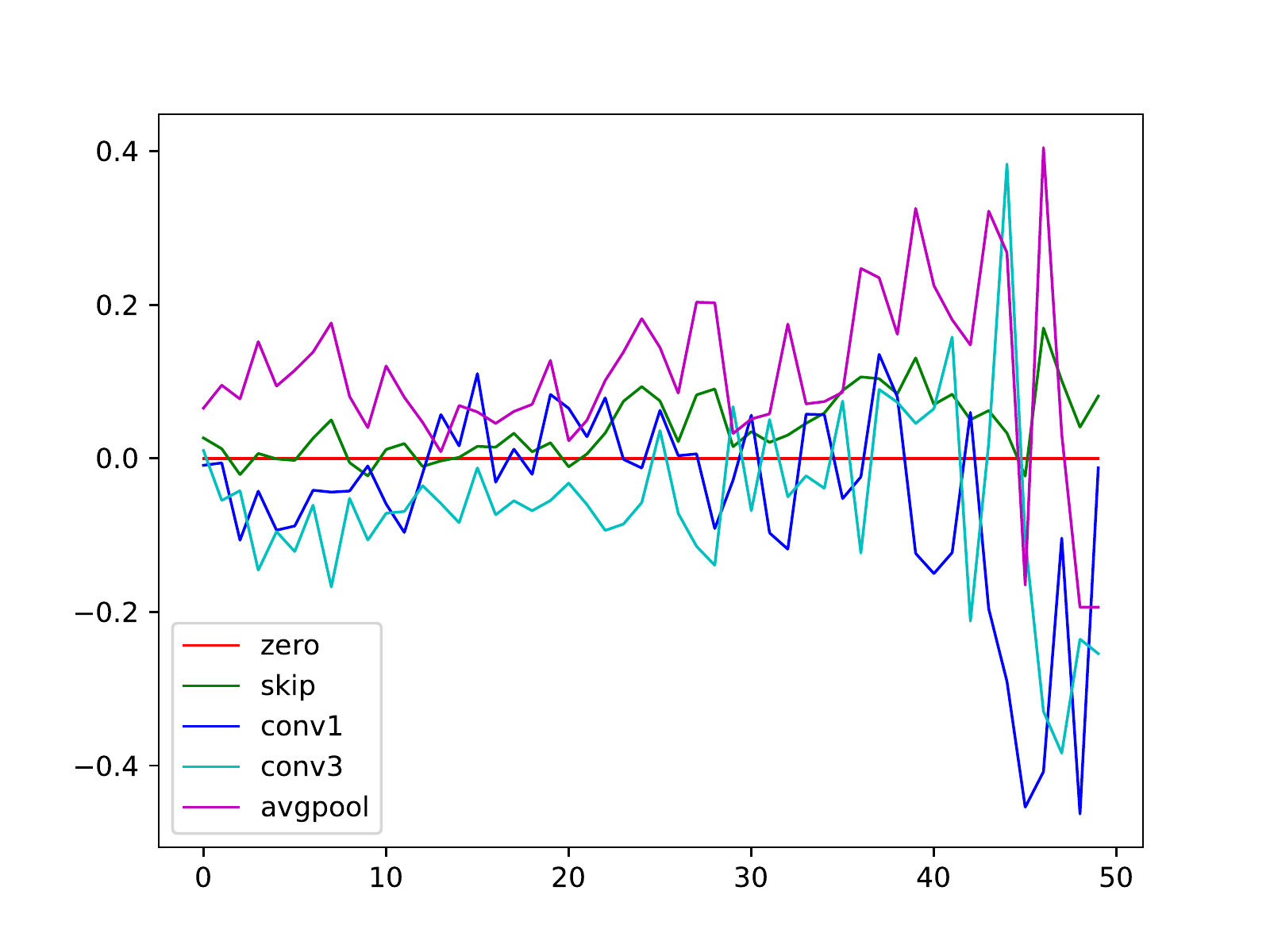}
 \caption{edge.3$\leftarrow$2}
\end{subfigure}
\hfill
\caption{Bi-level optimization. On the 0th cell.}
\end{figure}

\begin{figure}[h]
 \begin{subfigure}[b]{0.3\linewidth}
 \centering
\includegraphics[width=\textwidth]{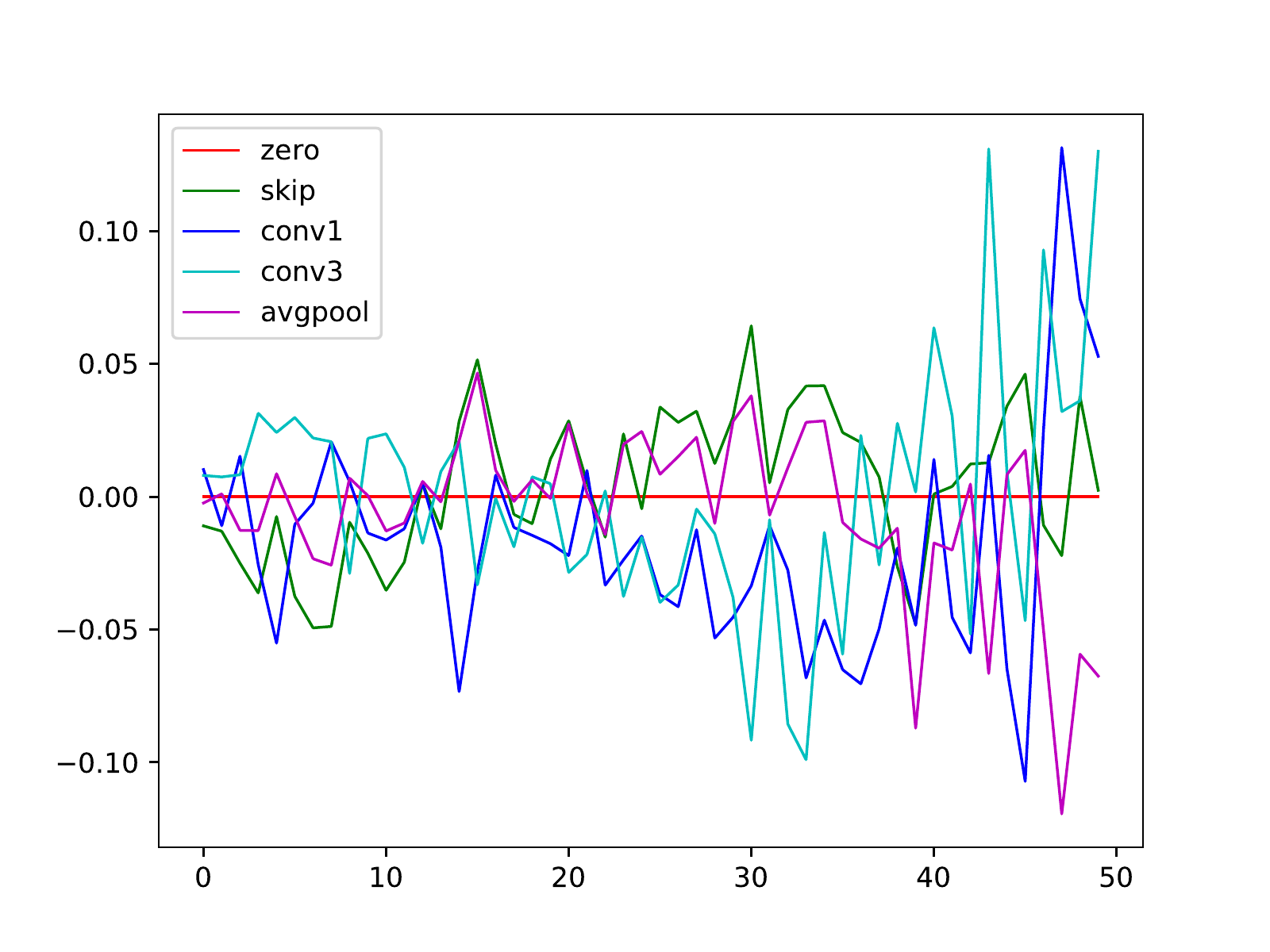}
 \caption{edge.1$\leftarrow$0}
\end{subfigure}
\hfill
 \begin{subfigure}[b]{0.3\linewidth}
 \centering
\includegraphics[width=\textwidth]{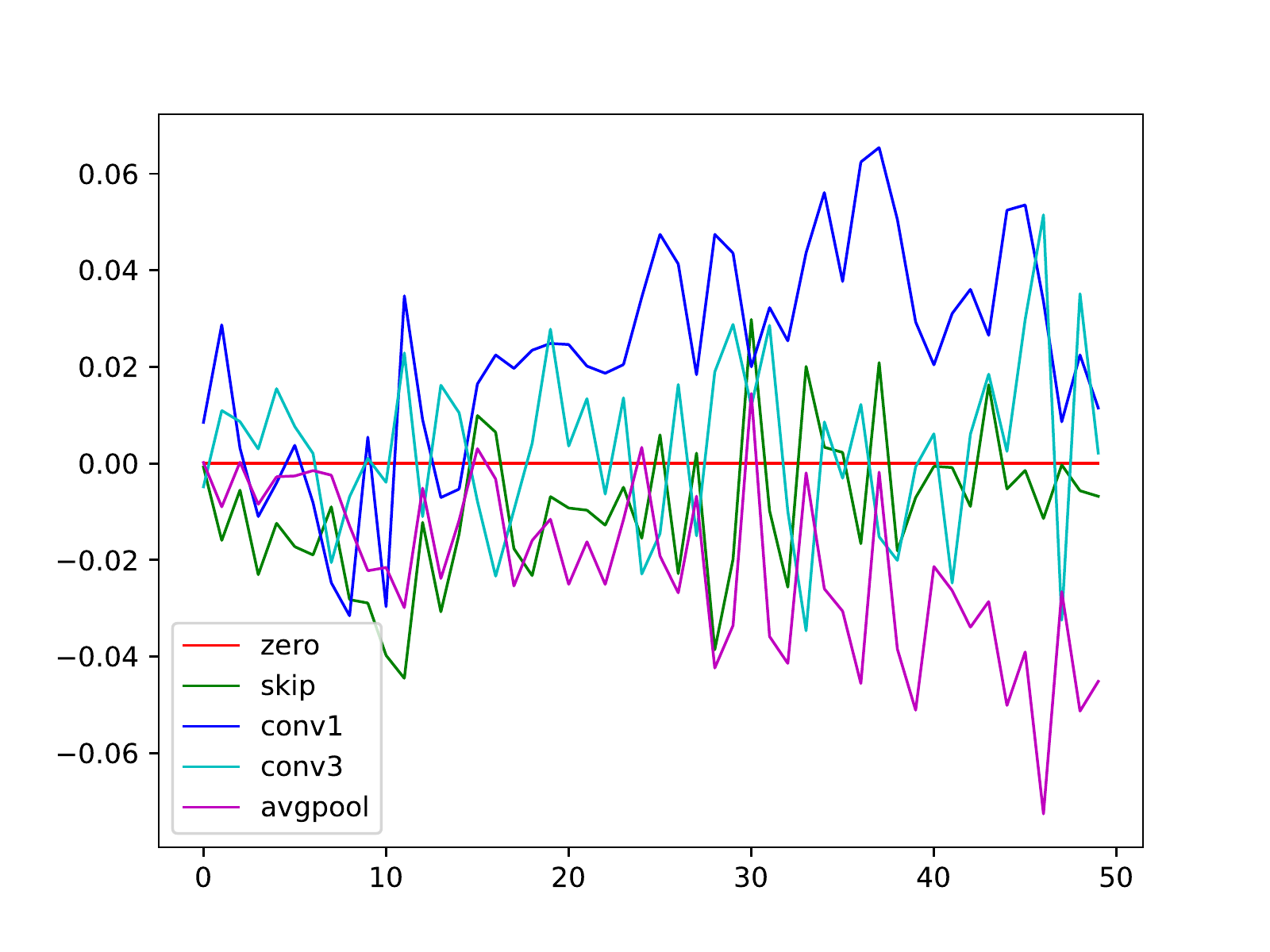}
 \caption{edge.2$\leftarrow$0}
\end{subfigure}
\hfill
 \begin{subfigure}[b]{0.3\linewidth}
 \centering
\includegraphics[width=\textwidth]{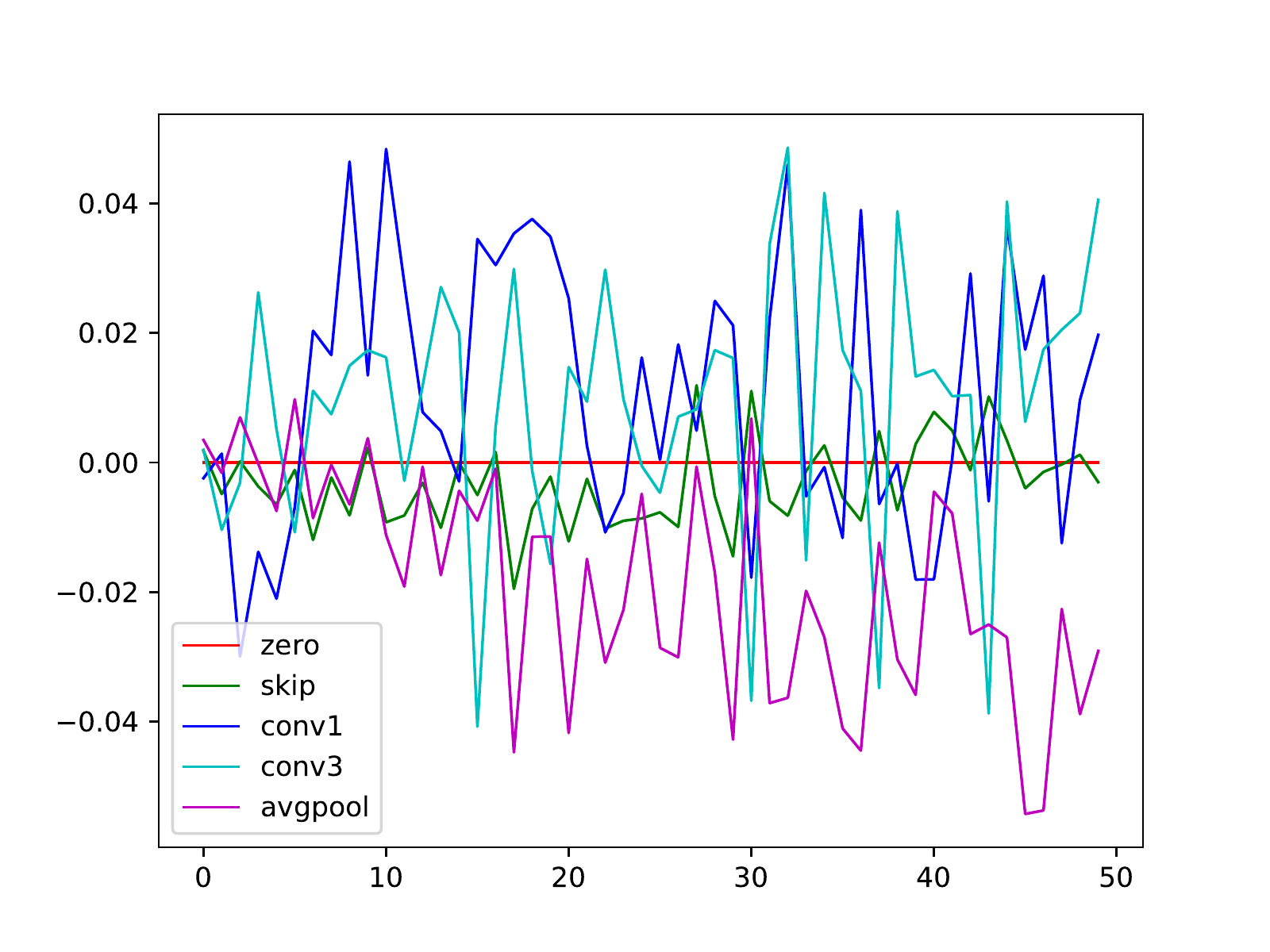}
 \caption{edge.2$\leftarrow$1}
\end{subfigure}
\hfill
\quad
 \begin{subfigure}[b]{0.3\linewidth}
 \centering
\includegraphics[width=\textwidth]{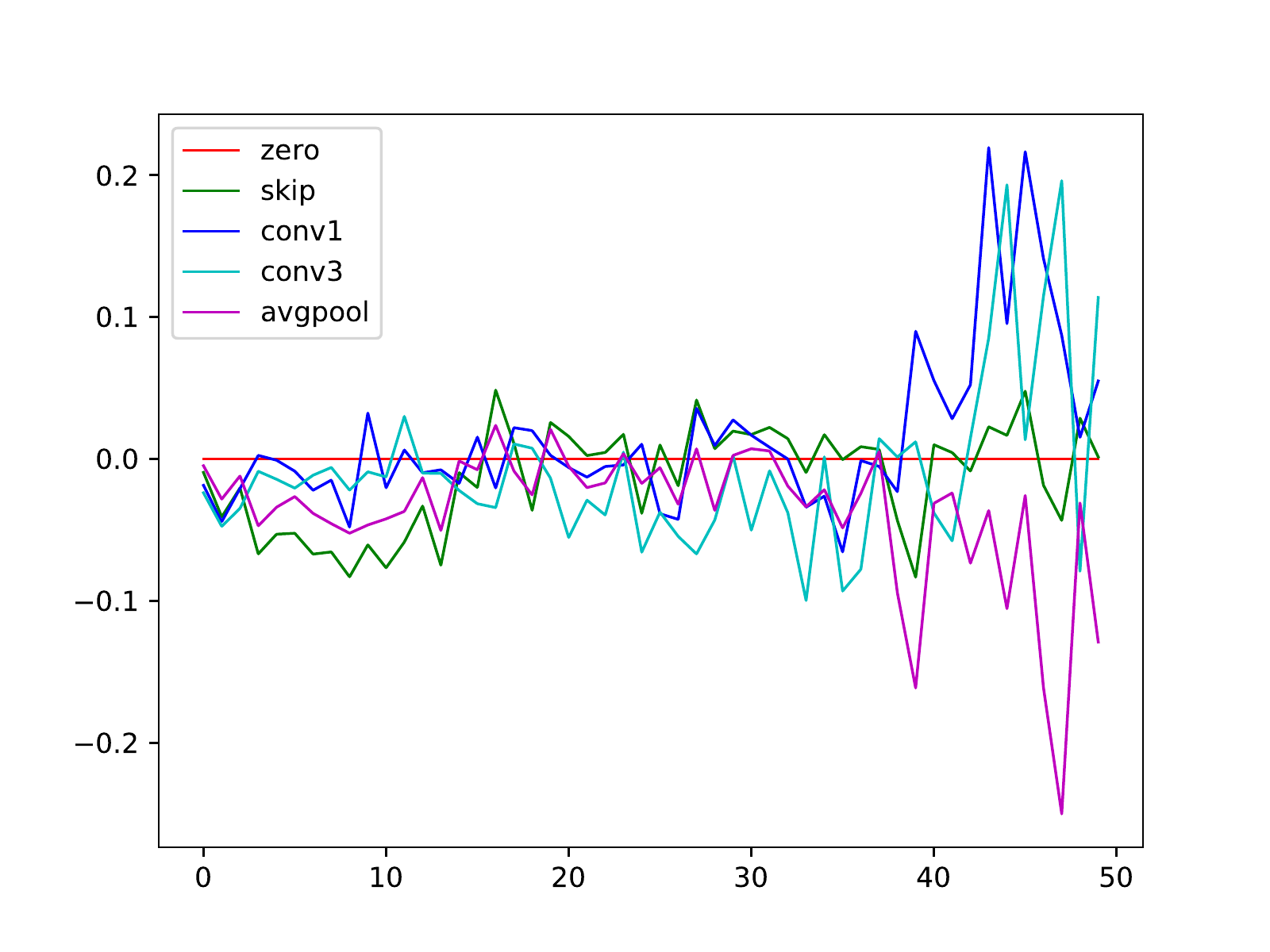}
 \caption{edge.3$\leftarrow$0}
\end{subfigure}
\hfill
 \begin{subfigure}[b]{0.3\linewidth}
 \centering
\includegraphics[width=\textwidth]{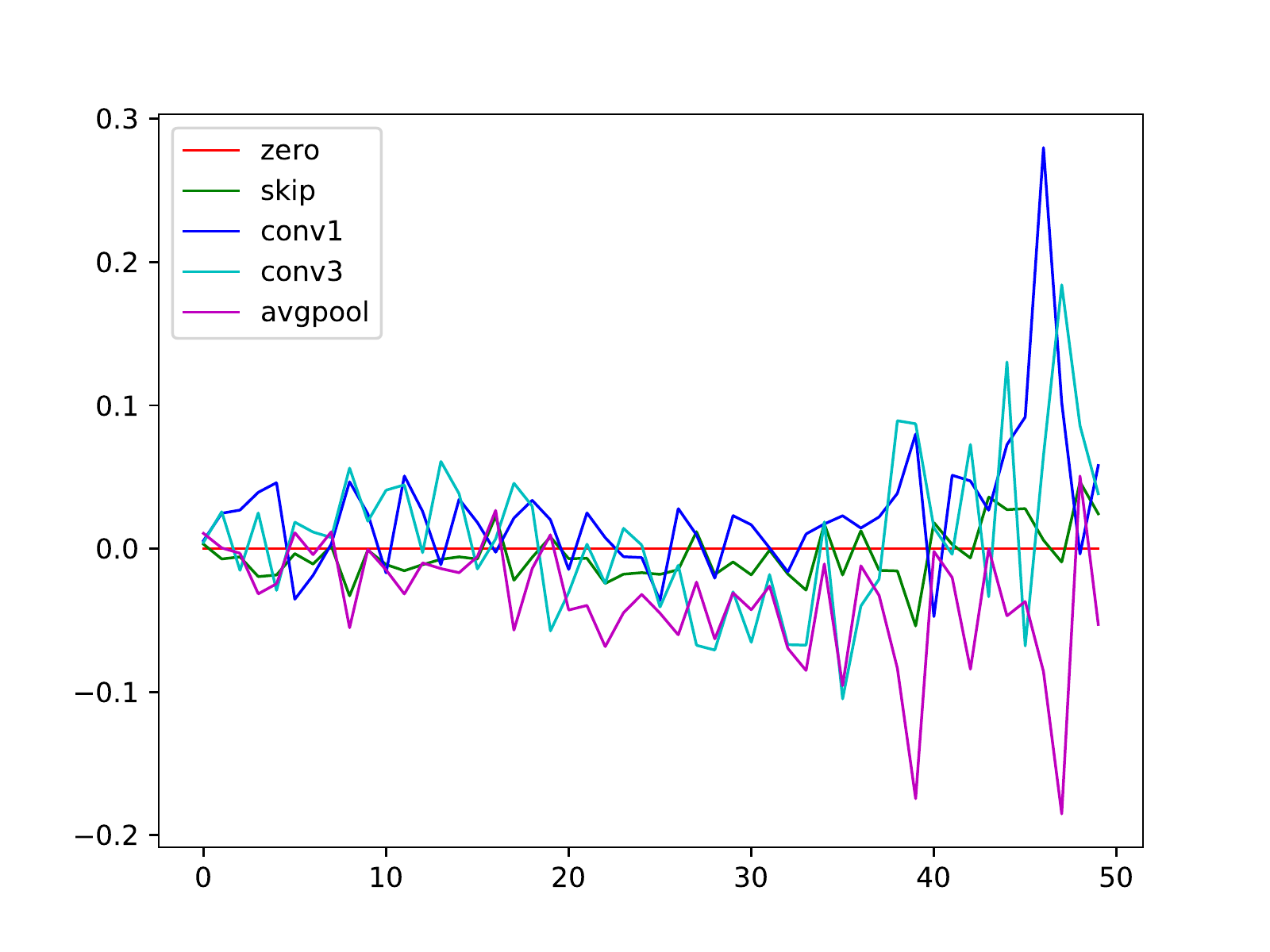}
 \caption{edge.3$\leftarrow$1}
\end{subfigure}
\hfill
 \begin{subfigure}[b]{0.3\linewidth}
 \centering
\includegraphics[width=\textwidth]{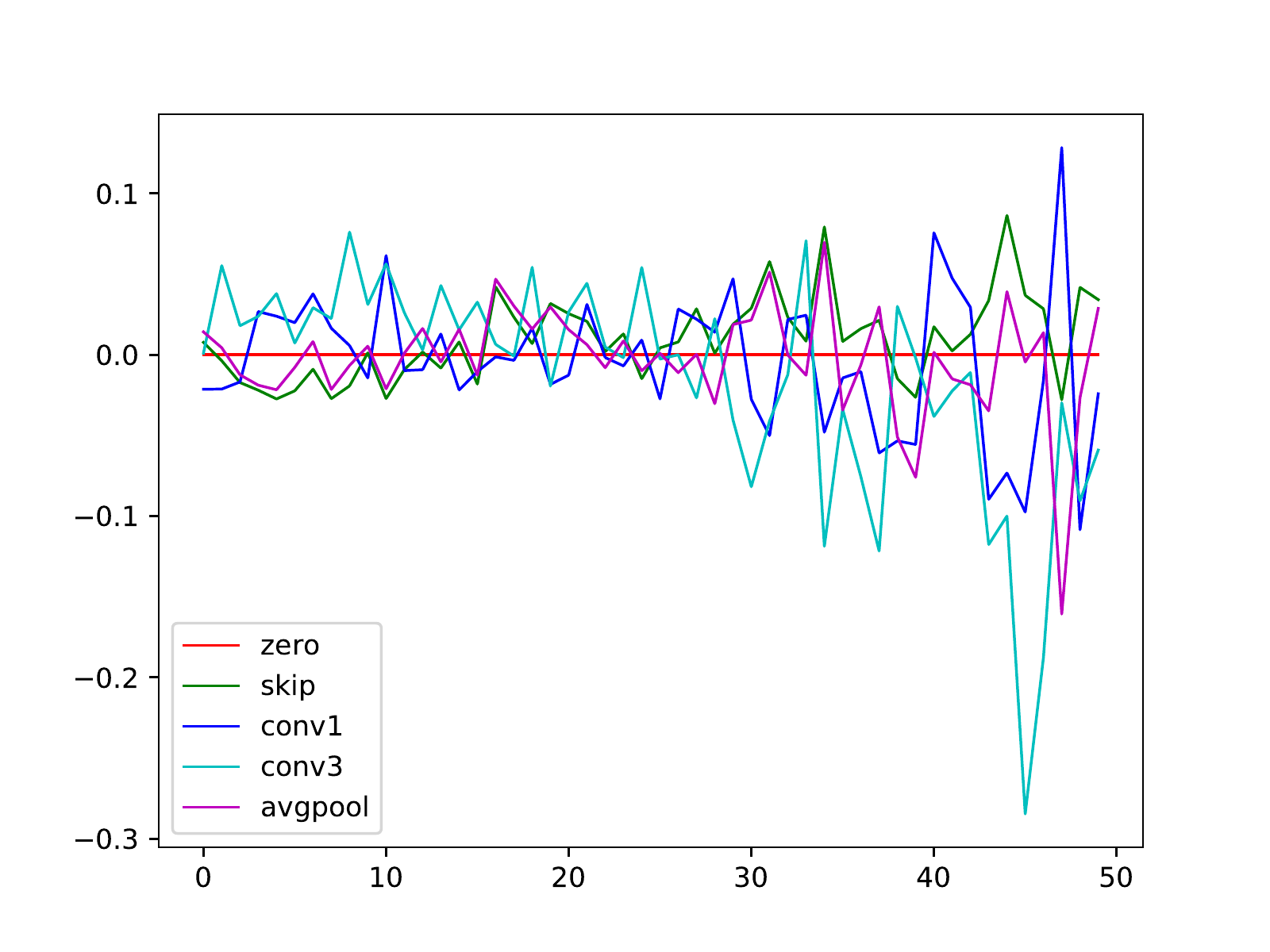}
 \caption{edge.3$\leftarrow$2}
\end{subfigure}
\hfill
\caption{Bi-level optimization. On the 8th cell.}
\end{figure}

\begin{figure}[h]
 \begin{subfigure}[b]{0.3\linewidth}
 \centering
\includegraphics[width=\textwidth]{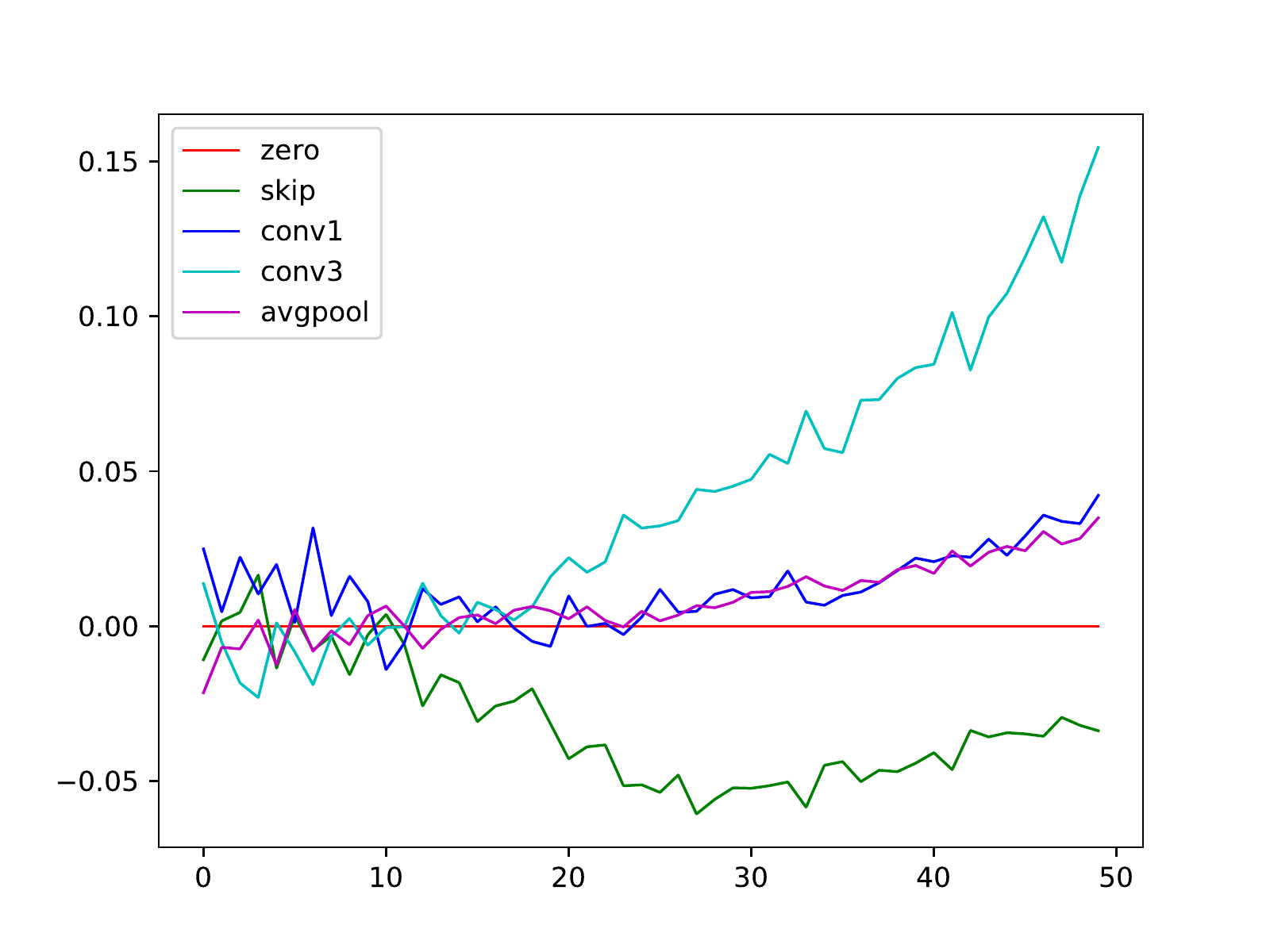}
 \caption{edge.1$\leftarrow$0}
\end{subfigure}
\hfill
 \begin{subfigure}[b]{0.3\linewidth}
 \centering
\includegraphics[width=\textwidth]{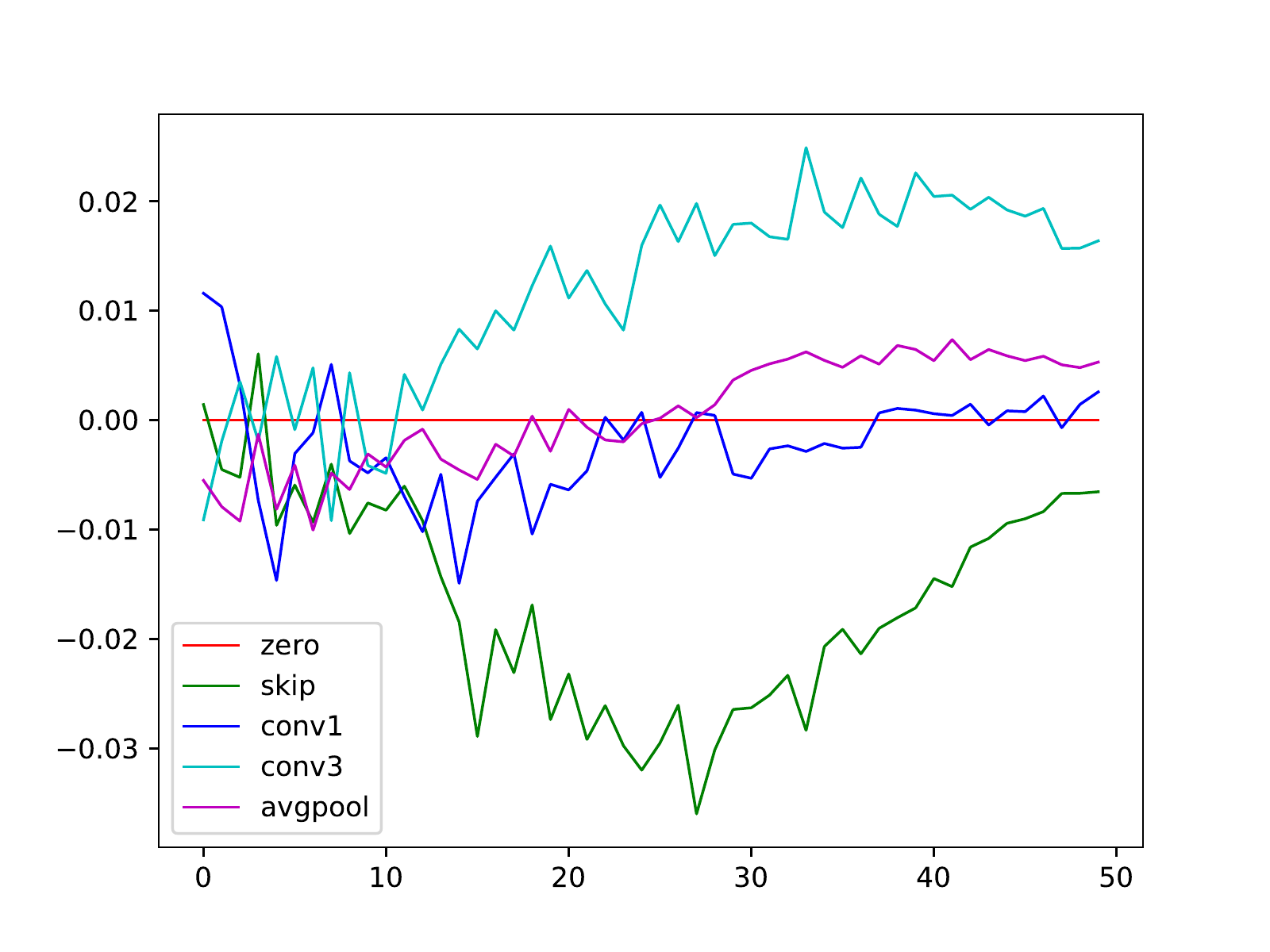}
 \caption{edge.2$\leftarrow$0}
\end{subfigure}
\hfill
 \begin{subfigure}[b]{0.3\linewidth}
 \centering
\includegraphics[width=\textwidth]{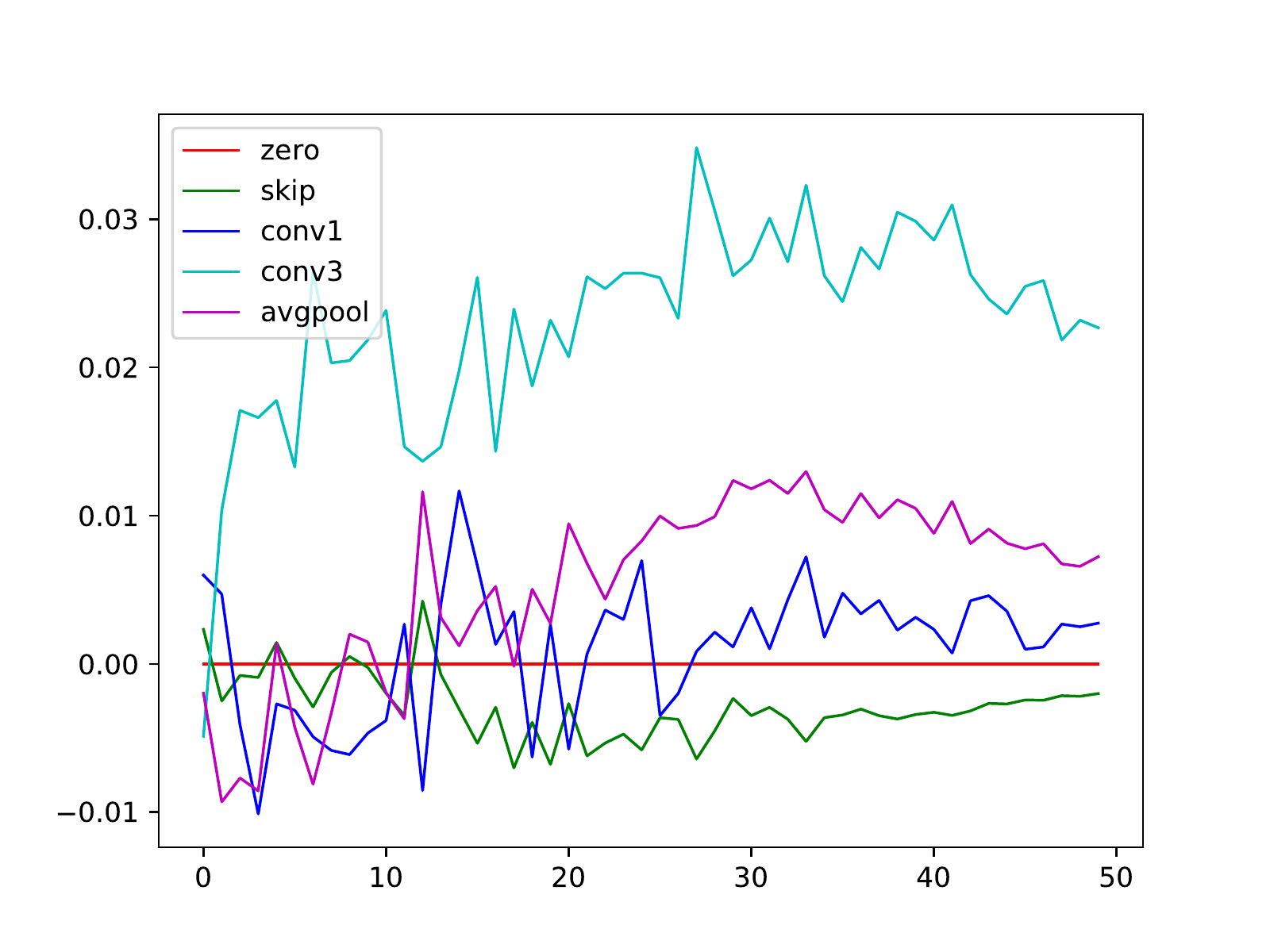}
 \caption{edge.2$\leftarrow$1}
\end{subfigure}
\hfill
\quad
\begin{subfigure}[b]{0.3\linewidth}
 \centering
\includegraphics[width=\textwidth]{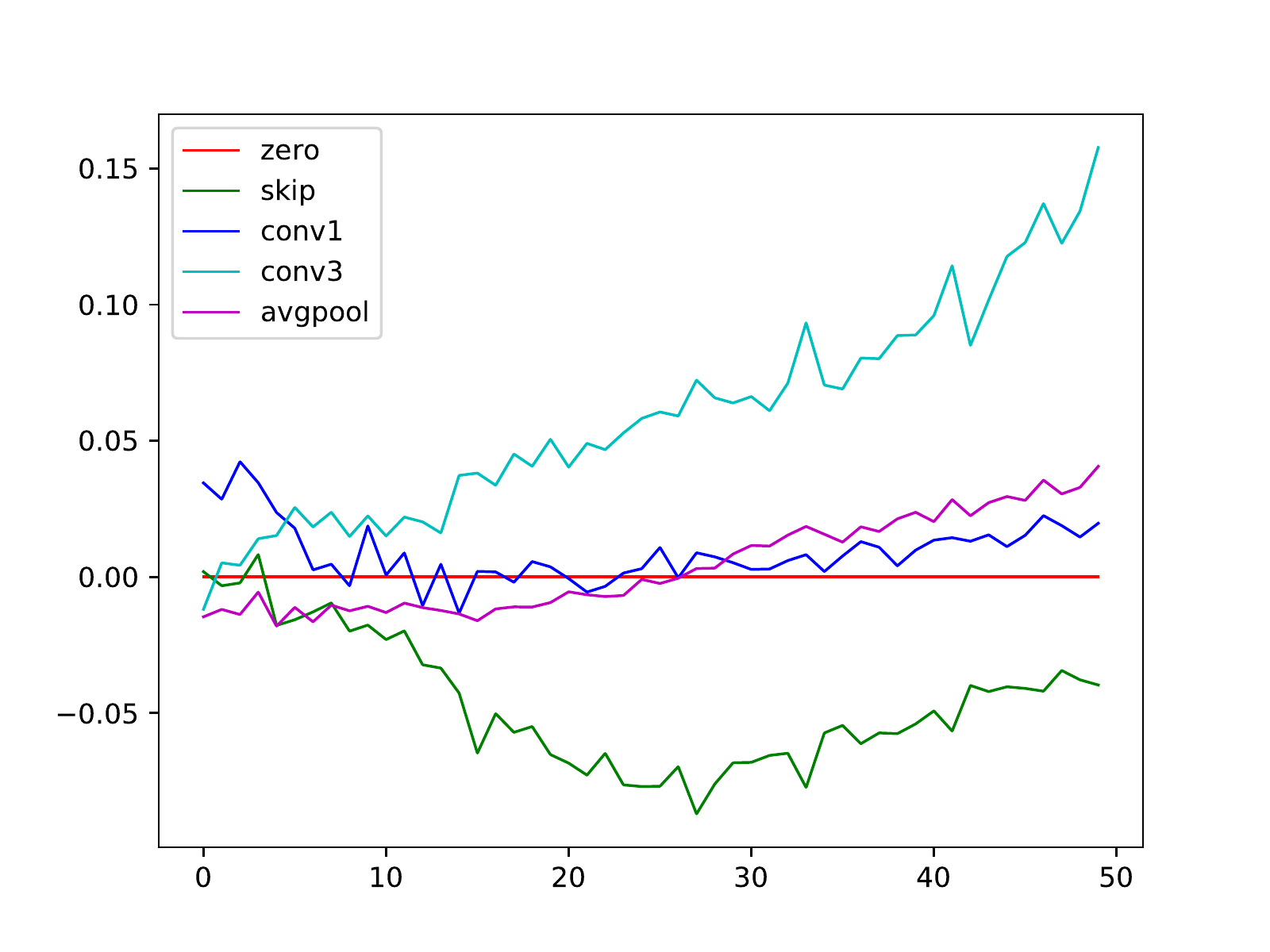}
 \caption{edge.3$\leftarrow$0}
\end{subfigure}
\hfill
 \begin{subfigure}[b]{0.3\linewidth}
 \centering
\includegraphics[width=\textwidth]{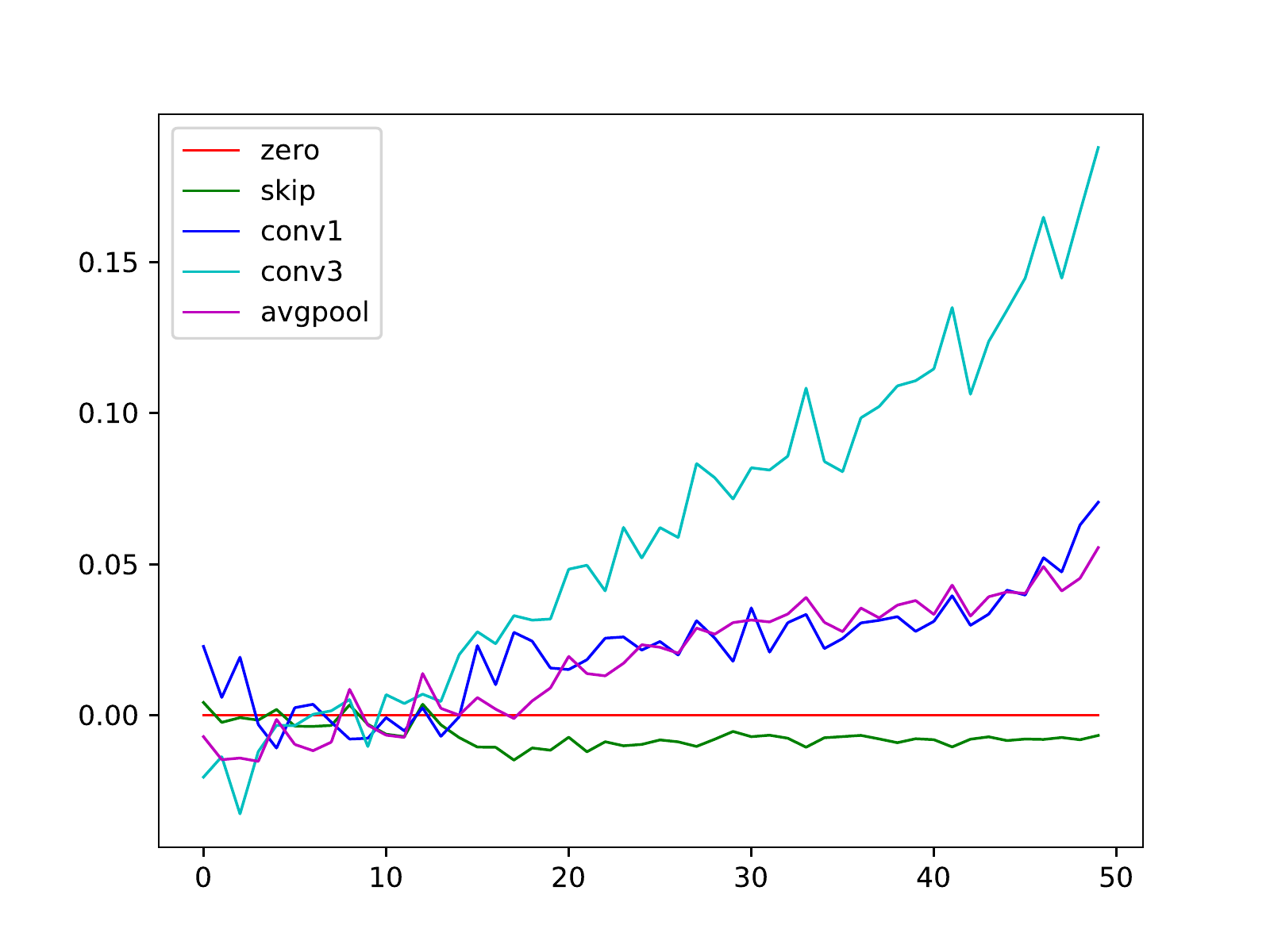}
 \caption{edge.3$\leftarrow$1}
\end{subfigure}
\hfill
 \begin{subfigure}[b]{0.3\linewidth}
 \centering
\includegraphics[width=\textwidth]{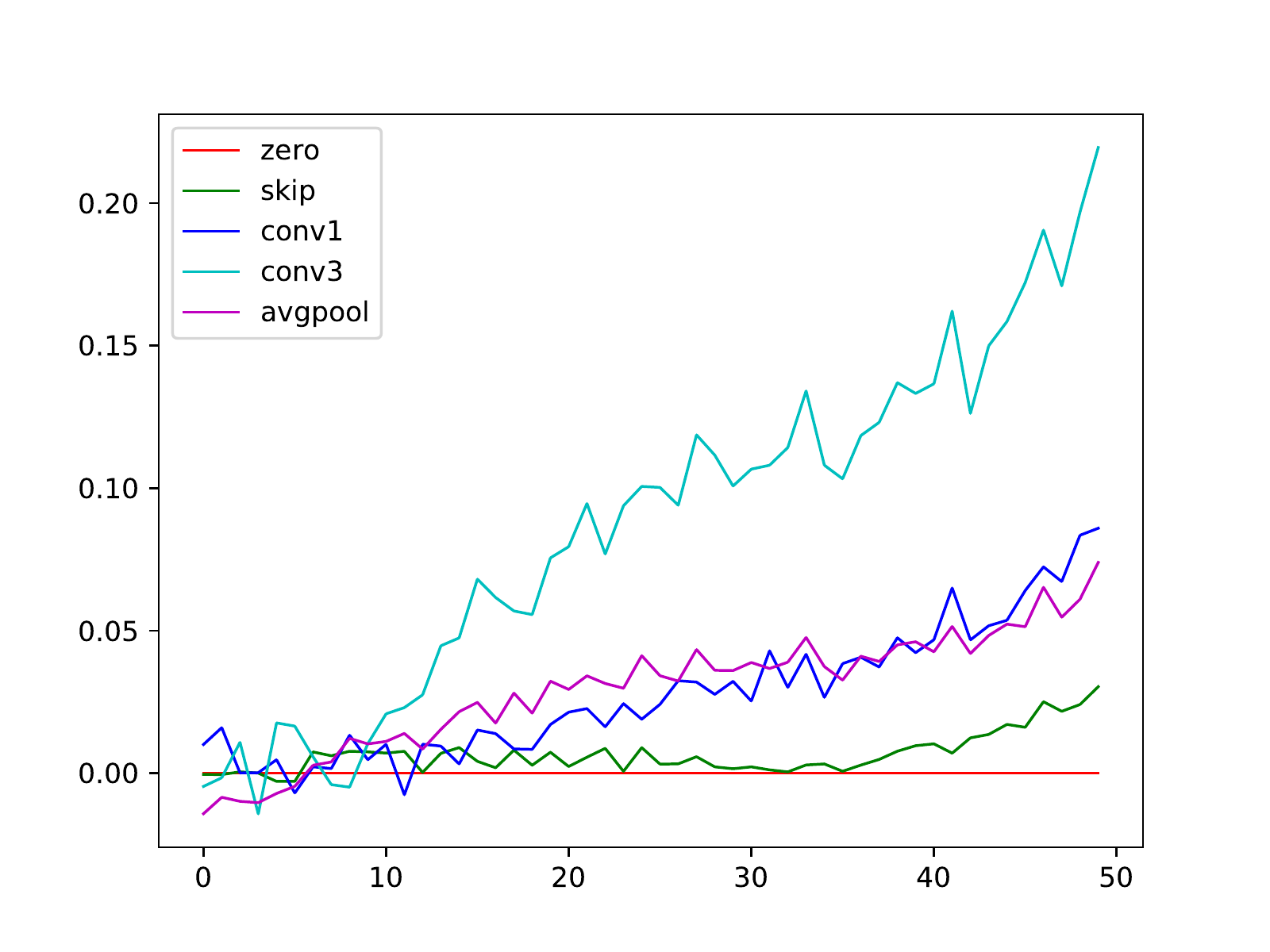}
 \caption{edge.3$\leftarrow$2}
\end{subfigure}
\hfill
\caption{Bi-level optimization. On the 16th cell.}
\end{figure}
\begin{figure}[h]
%\begin{center}
\centering
 \begin{subfigure}[b]{0.3\linewidth}
 \centering
\includegraphics[width=\textwidth]{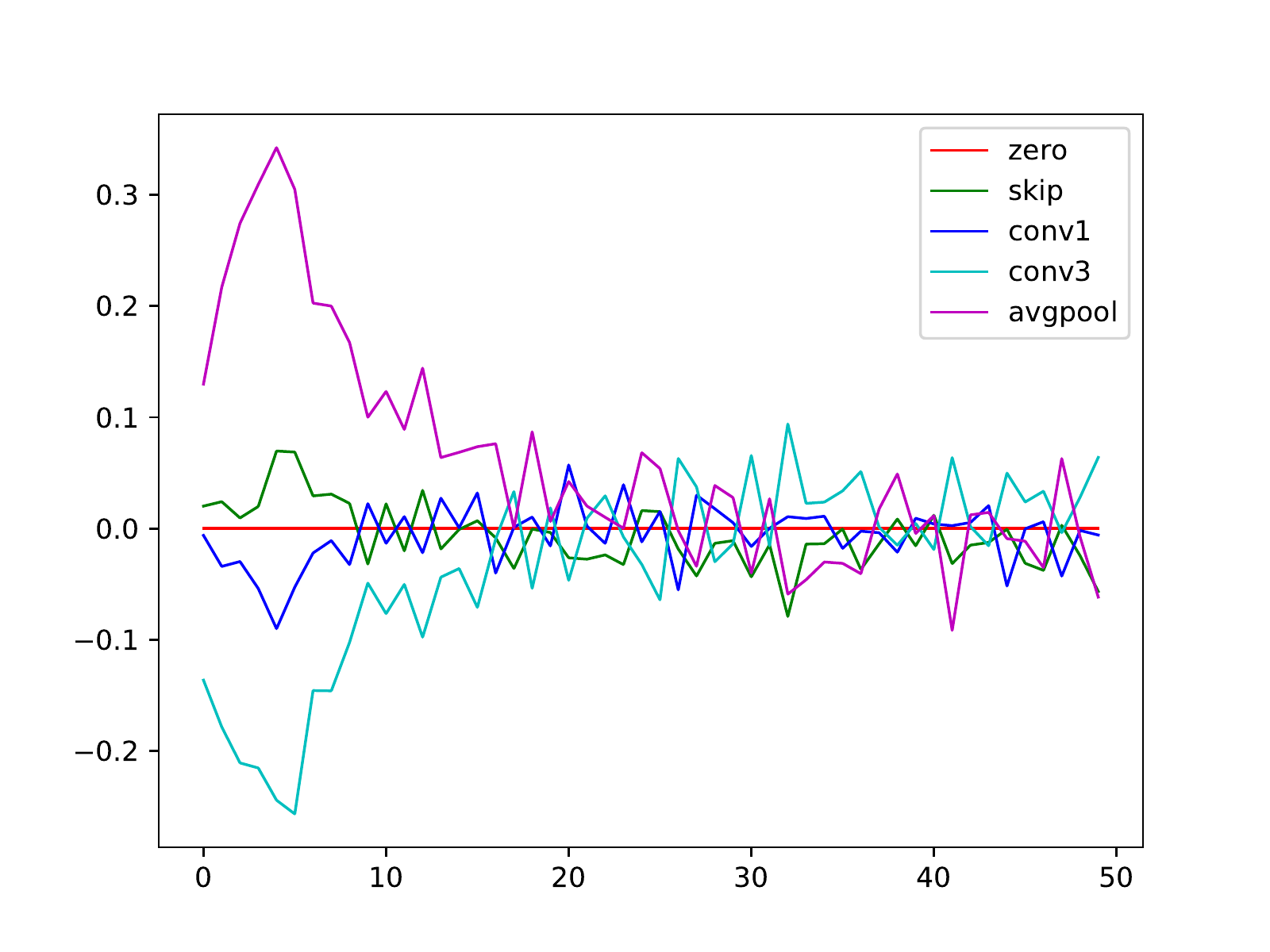}
 \caption{edge.1$\leftarrow$0}
\end{subfigure}
\hfill
 \begin{subfigure}[b]{0.3\linewidth}
 \centering
\includegraphics[width=\textwidth]{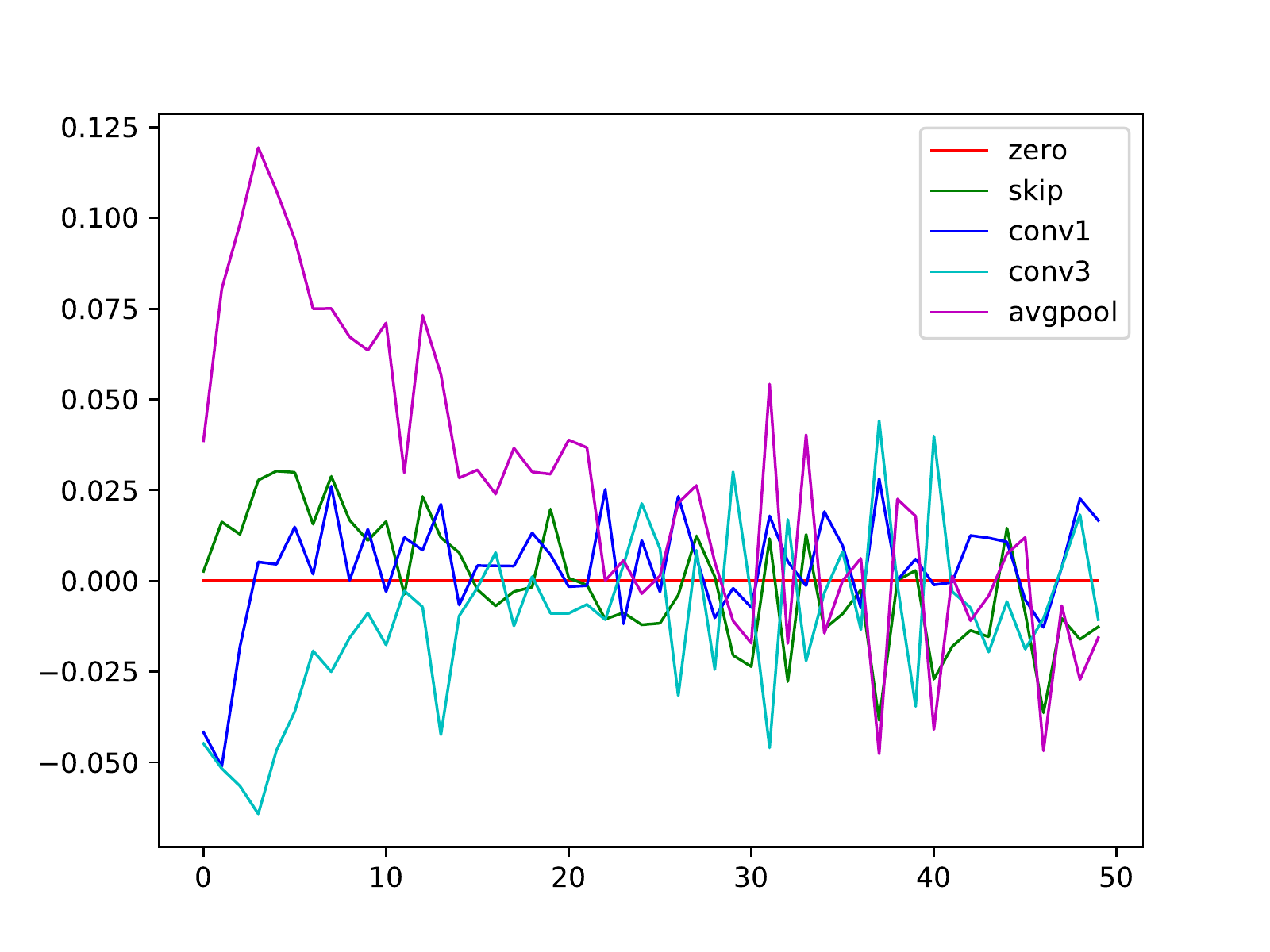}
 \caption{edge.2$\leftarrow$0}
\end{subfigure}
\hfill
 \begin{subfigure}[b]{0.3\linewidth}
 \centering
\includegraphics[width=\textwidth]{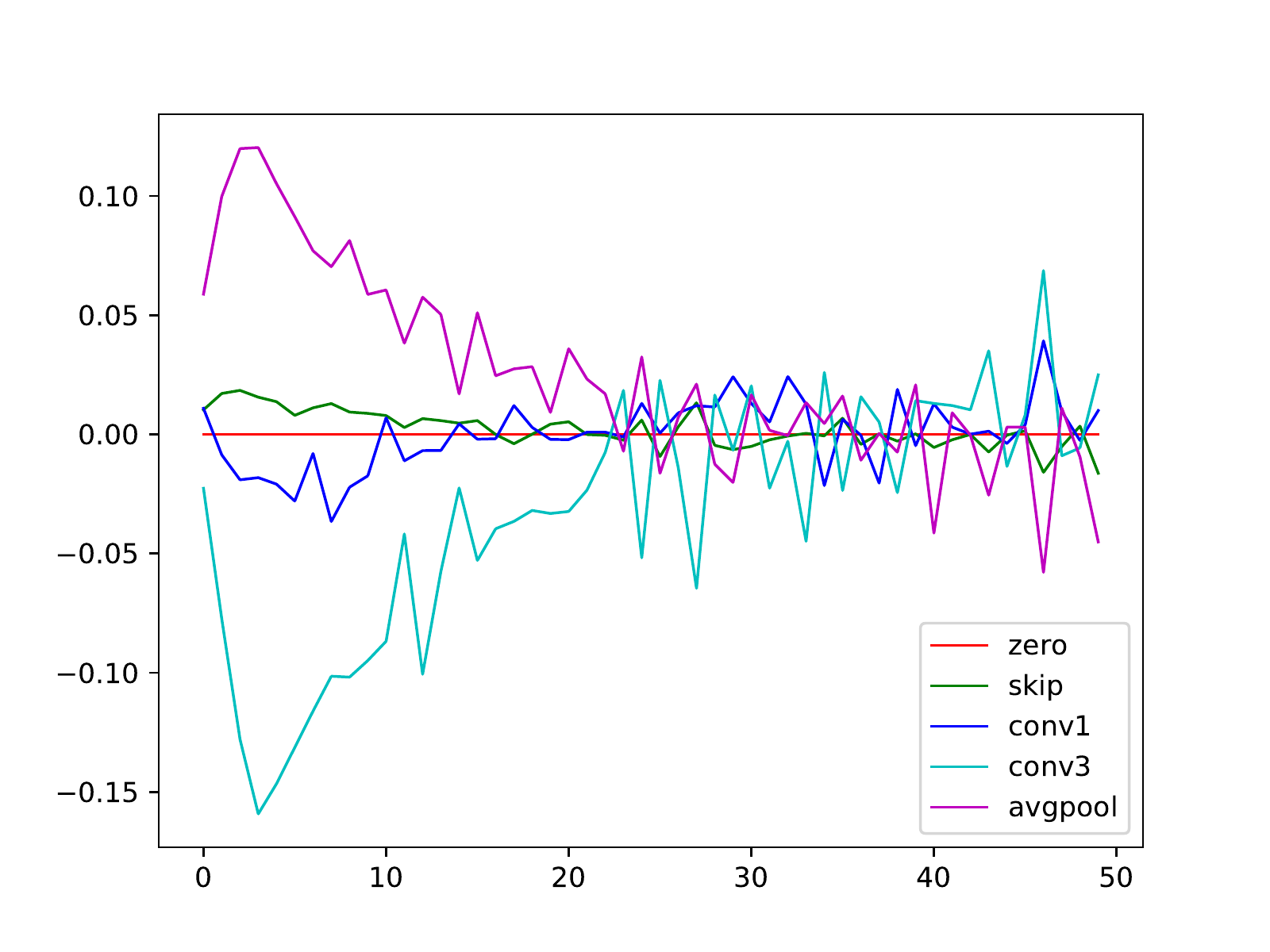}
 \caption{edge.2$\leftarrow$1}
\end{subfigure}
\hfill
\quad
 \begin{subfigure}[b]{0.3\linewidth}
 \centering
\includegraphics[width=\textwidth]{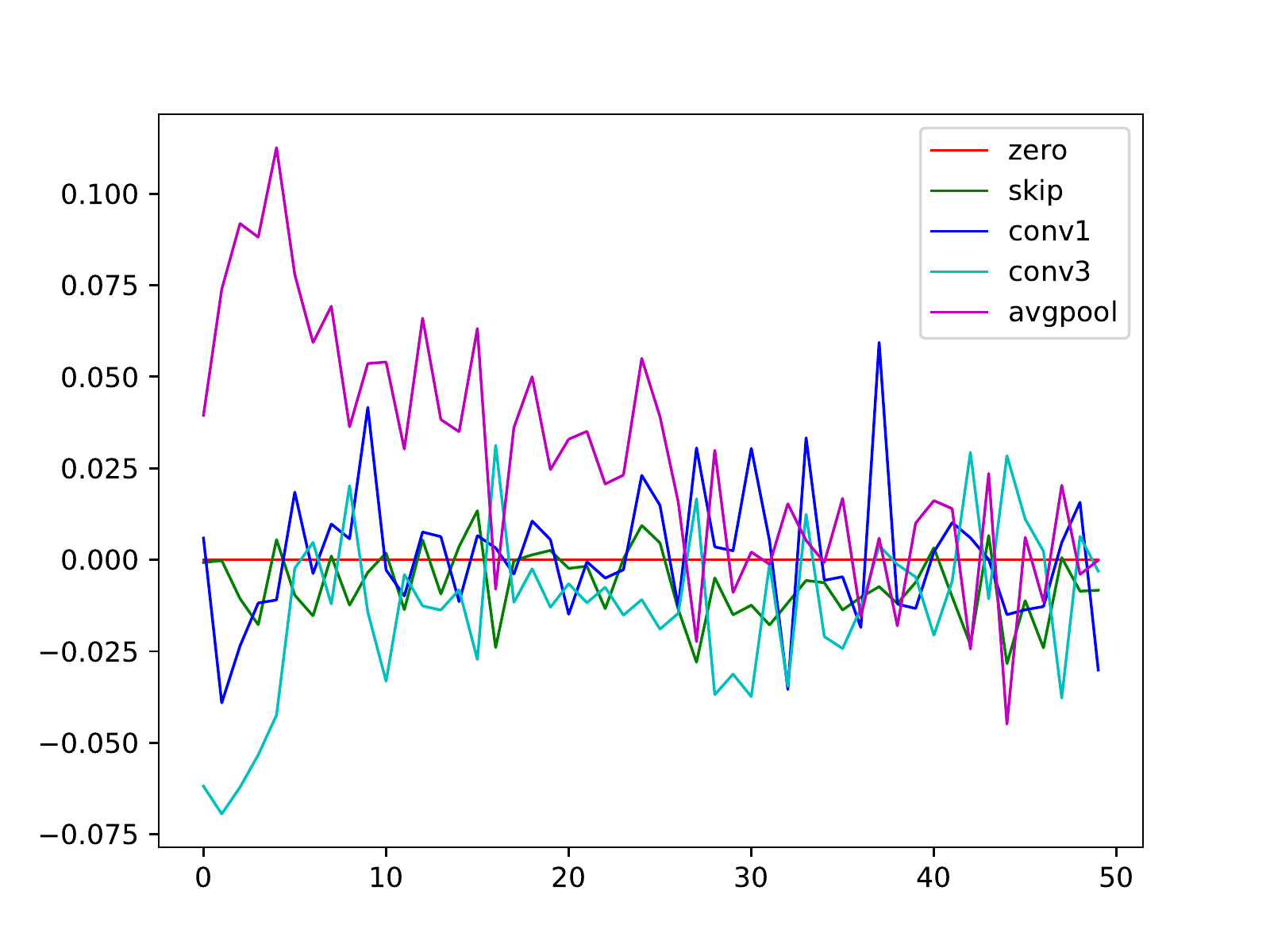}
 \caption{edge.3$\leftarrow$0}
\end{subfigure}
\hfill
 \begin{subfigure}[b]{0.3\linewidth}
 \centering
\includegraphics[width=\textwidth]{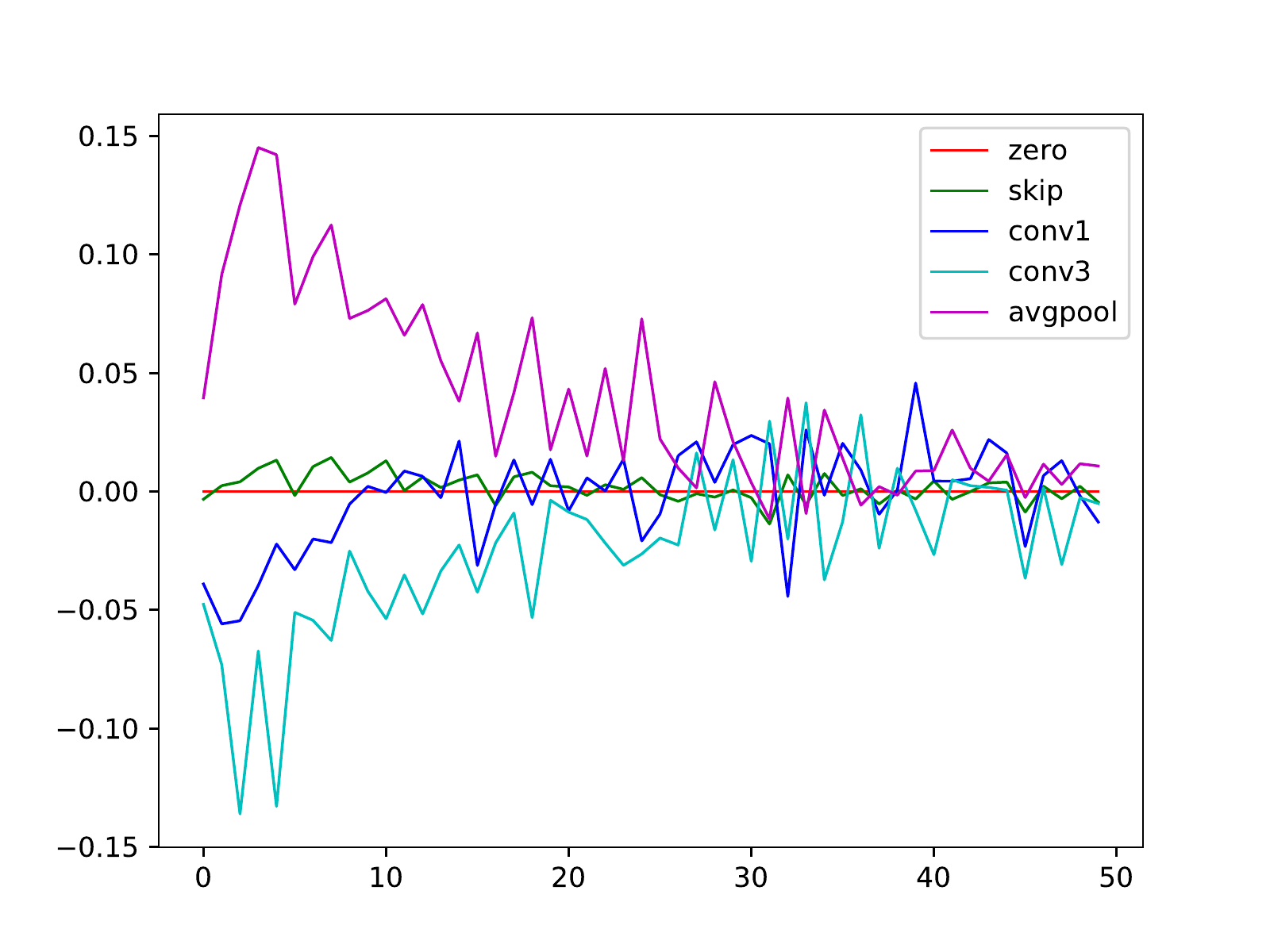}
 \caption{edge.3$\leftarrow$1}
\end{subfigure}
\hfill
 \begin{subfigure}[b]{0.3\linewidth}
 \centering
\includegraphics[width=\textwidth]{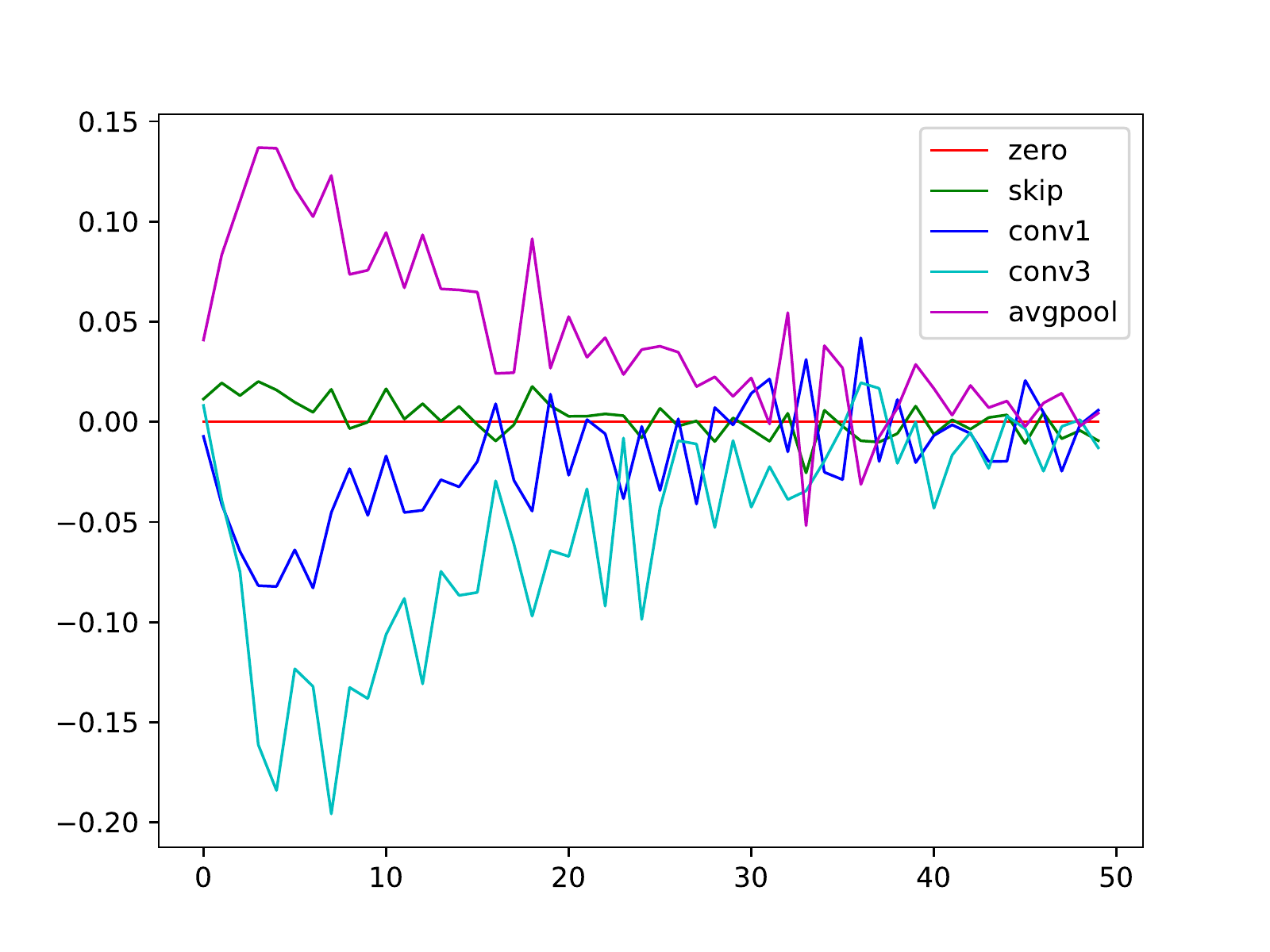}
 \caption{edge.3$\leftarrow$2}
\end{subfigure}
\hfill
\caption{Single-level optimization. On the 0th cell.}
\end{figure}

\begin{figure}[h]
 \begin{subfigure}[b]{0.3\linewidth}
 \centering
\includegraphics[width=\textwidth]{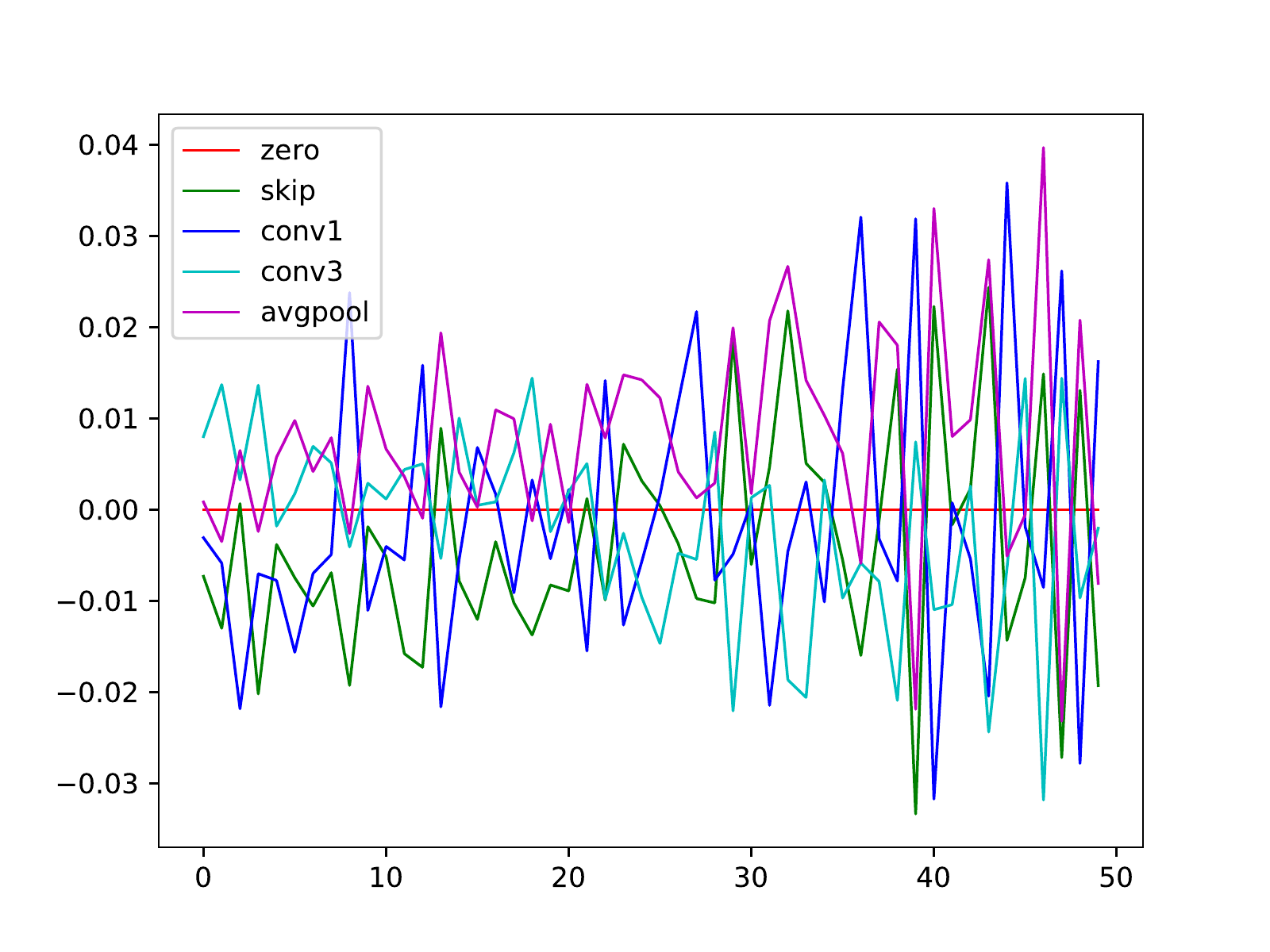}
 \caption{edge.1$\leftarrow$0}
\end{subfigure}
\hfill
 \begin{subfigure}[b]{0.3\linewidth}
 \centering
\includegraphics[width=\textwidth]{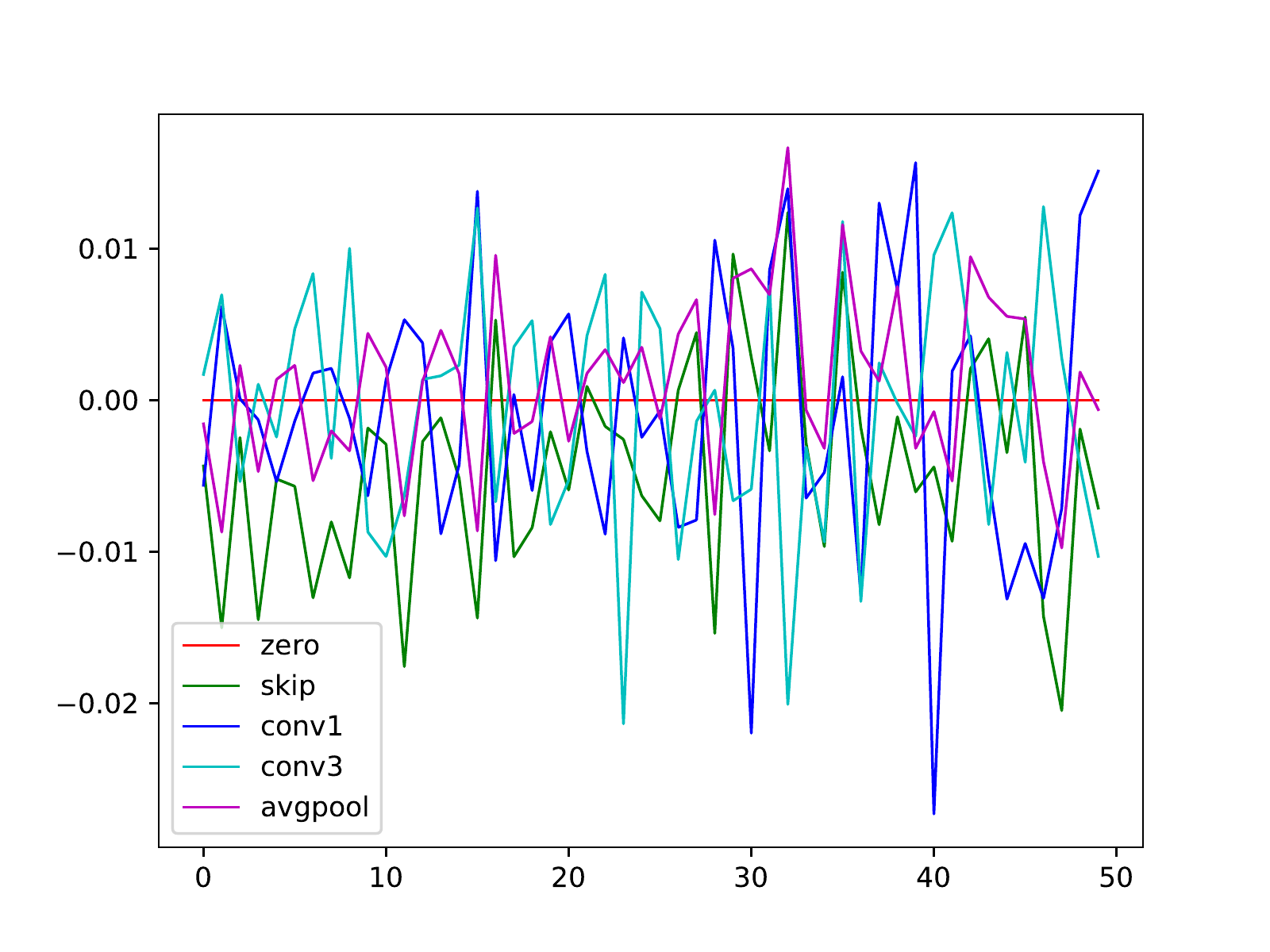}
 \caption{edge.2$\leftarrow$0}
\end{subfigure}
\hfill
 \begin{subfigure}[b]{0.3\linewidth}
 \centering
\includegraphics[width=\textwidth]{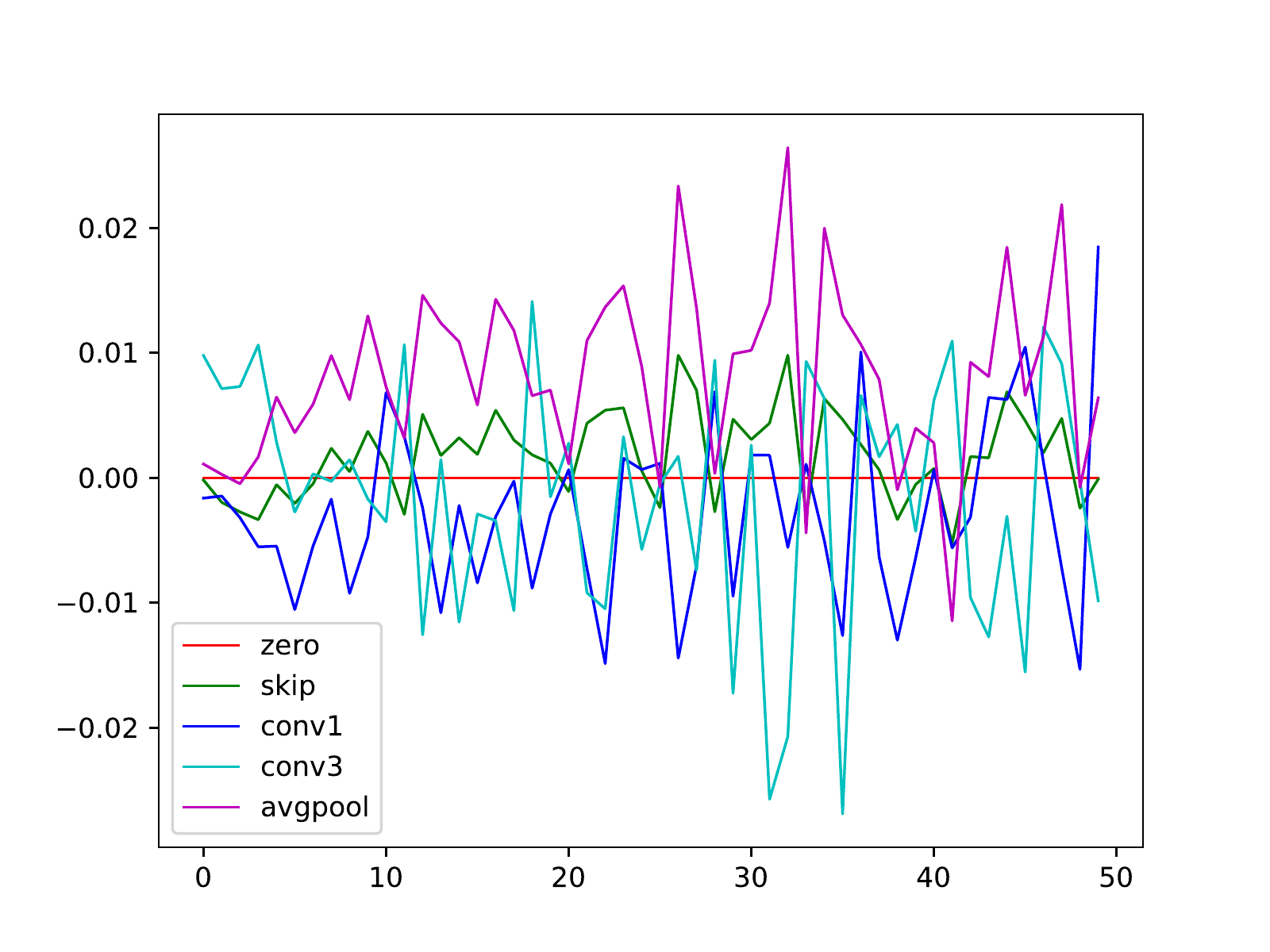}
 \caption{edge.2$\leftarrow$1}
\end{subfigure}
\hfill
\quad
 \begin{subfigure}[b]{0.3\linewidth}
 \centering
\includegraphics[width=\textwidth]{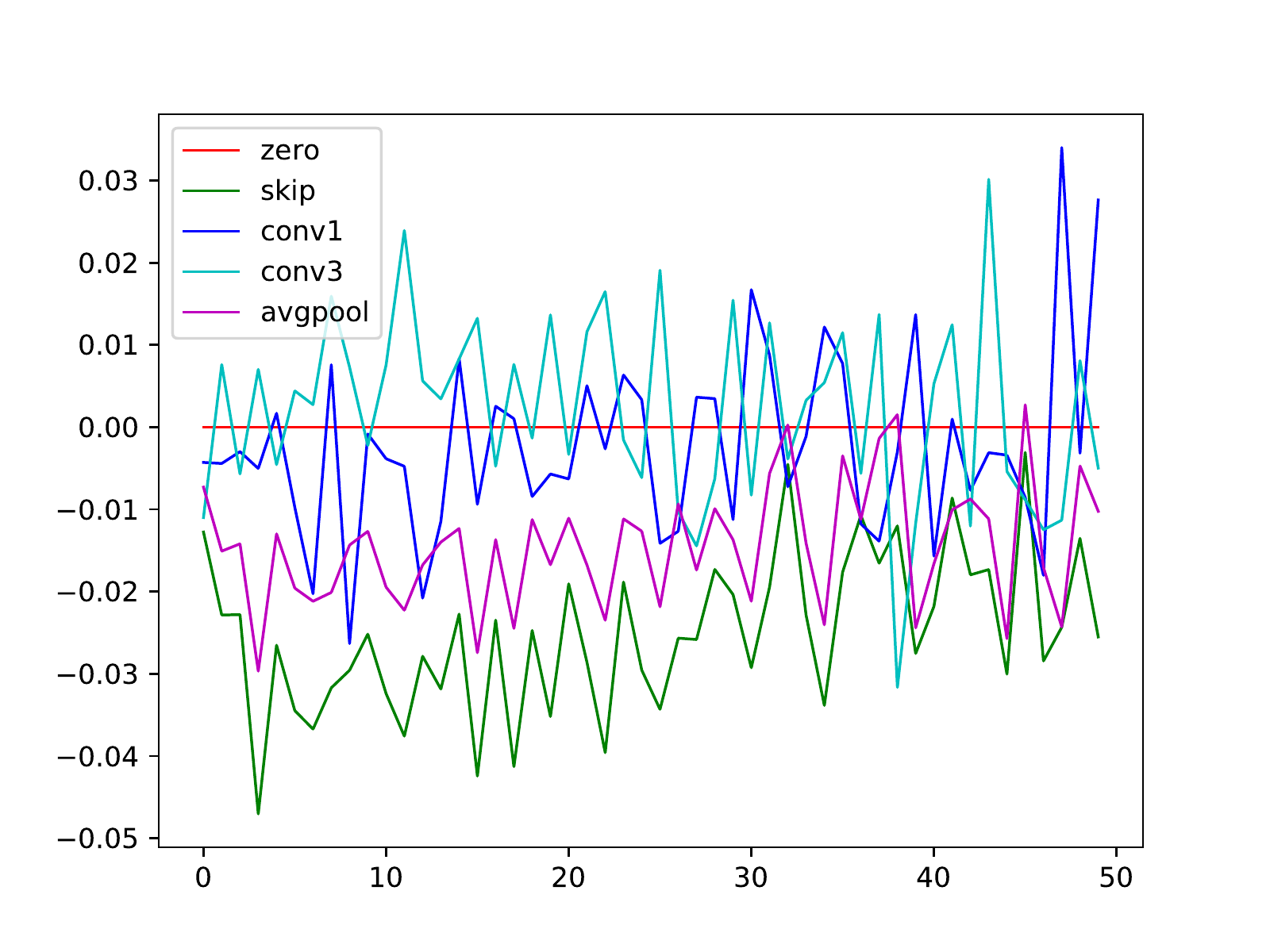}
 \caption{edge.3$\leftarrow$0}
\end{subfigure}
\hfill
 \begin{subfigure}[b]{0.3\linewidth}
 \centering
\includegraphics[width=\textwidth]{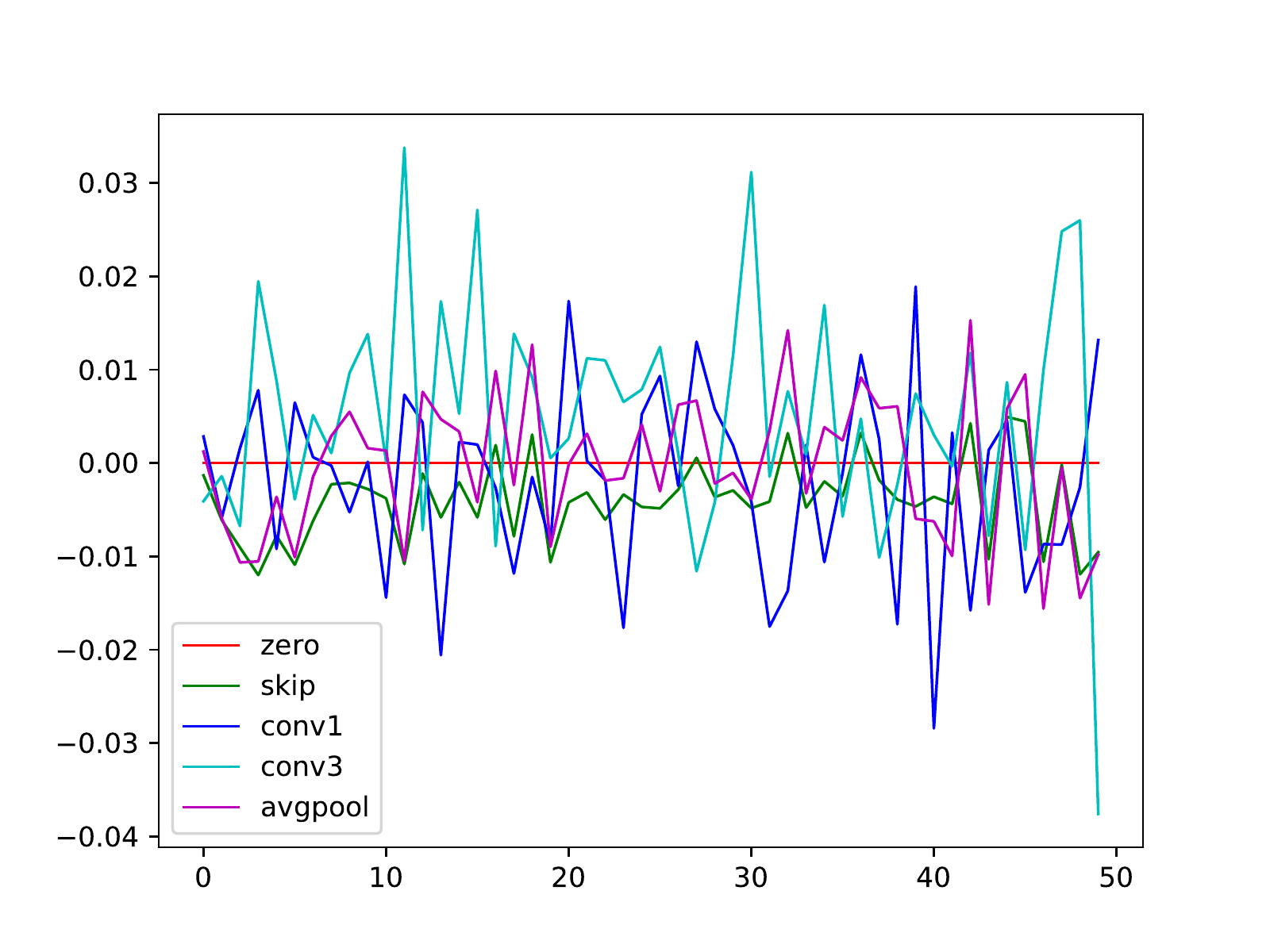}
 \caption{edge.3$\leftarrow$1}
\end{subfigure}
\hfill
 \begin{subfigure}[b]{0.3\linewidth}
 \centering
\includegraphics[width=\textwidth]{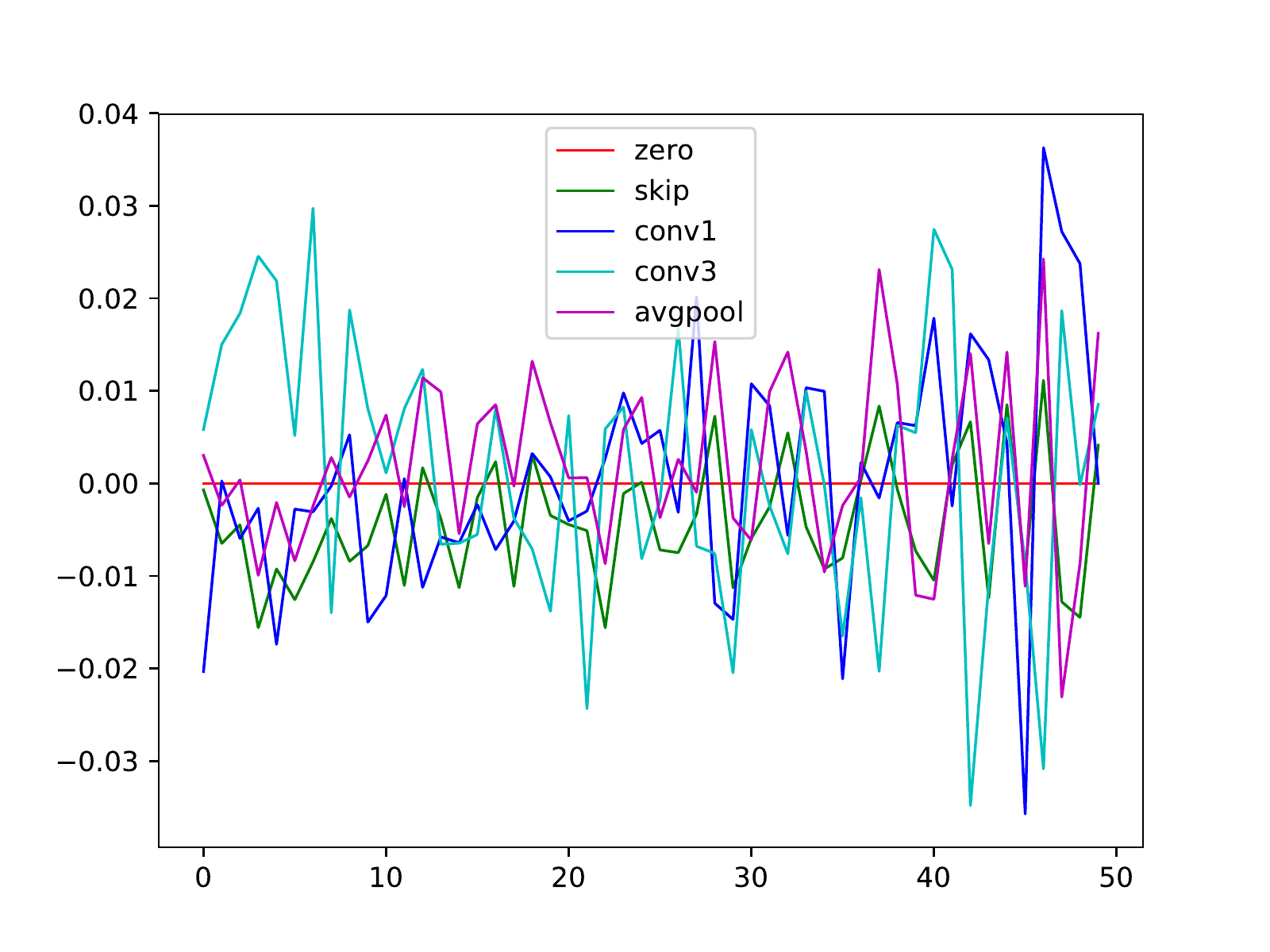}
 \caption{edge.3$\leftarrow$2}
\end{subfigure}
\hfill
\caption{Single-level optimization. On the 8th cell.}
\end{figure}

\begin{figure}[h]
 \begin{subfigure}[b]{0.3\linewidth}
 \centering
\includegraphics[width=\textwidth]{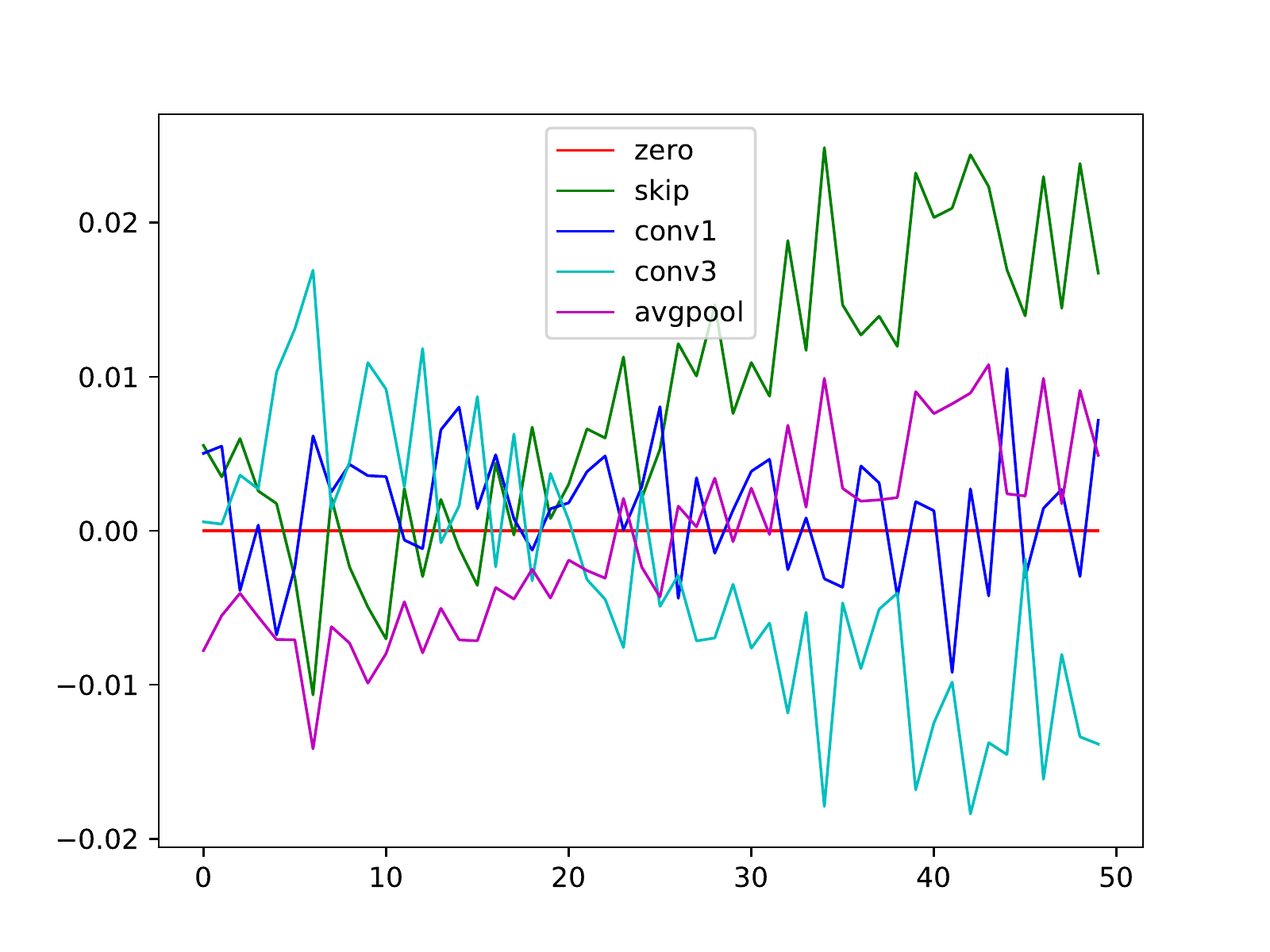}
 \caption{edge.1$\leftarrow$0}
\end{subfigure}
\hfill
 \begin{subfigure}[b]{0.3\linewidth}
 \centering
\includegraphics[width=\textwidth]{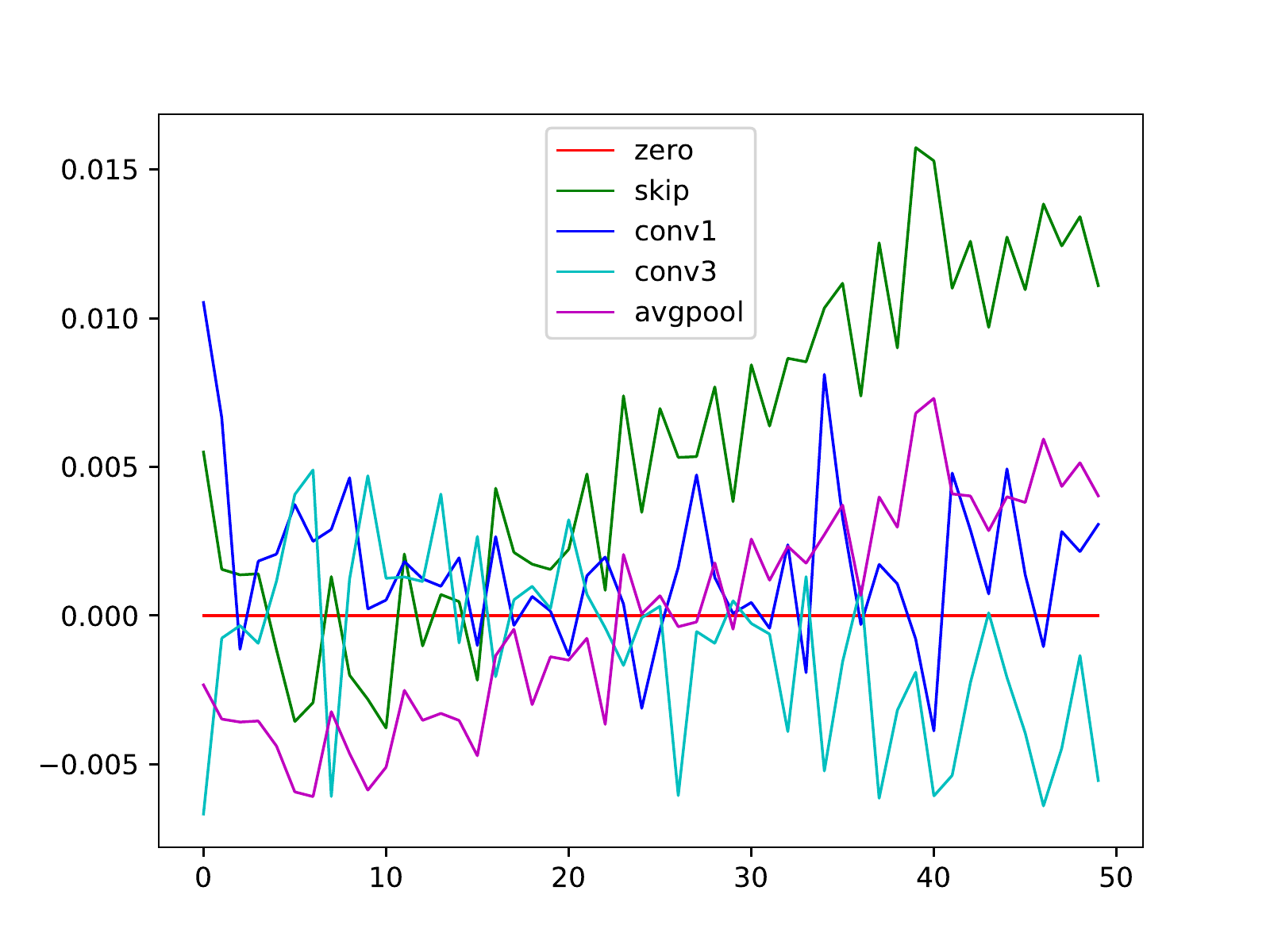}
 \caption{edge.2$\leftarrow$0}
\end{subfigure}
\hfill
 \begin{subfigure}[b]{0.3\linewidth}
 \centering
\includegraphics[width=\textwidth]{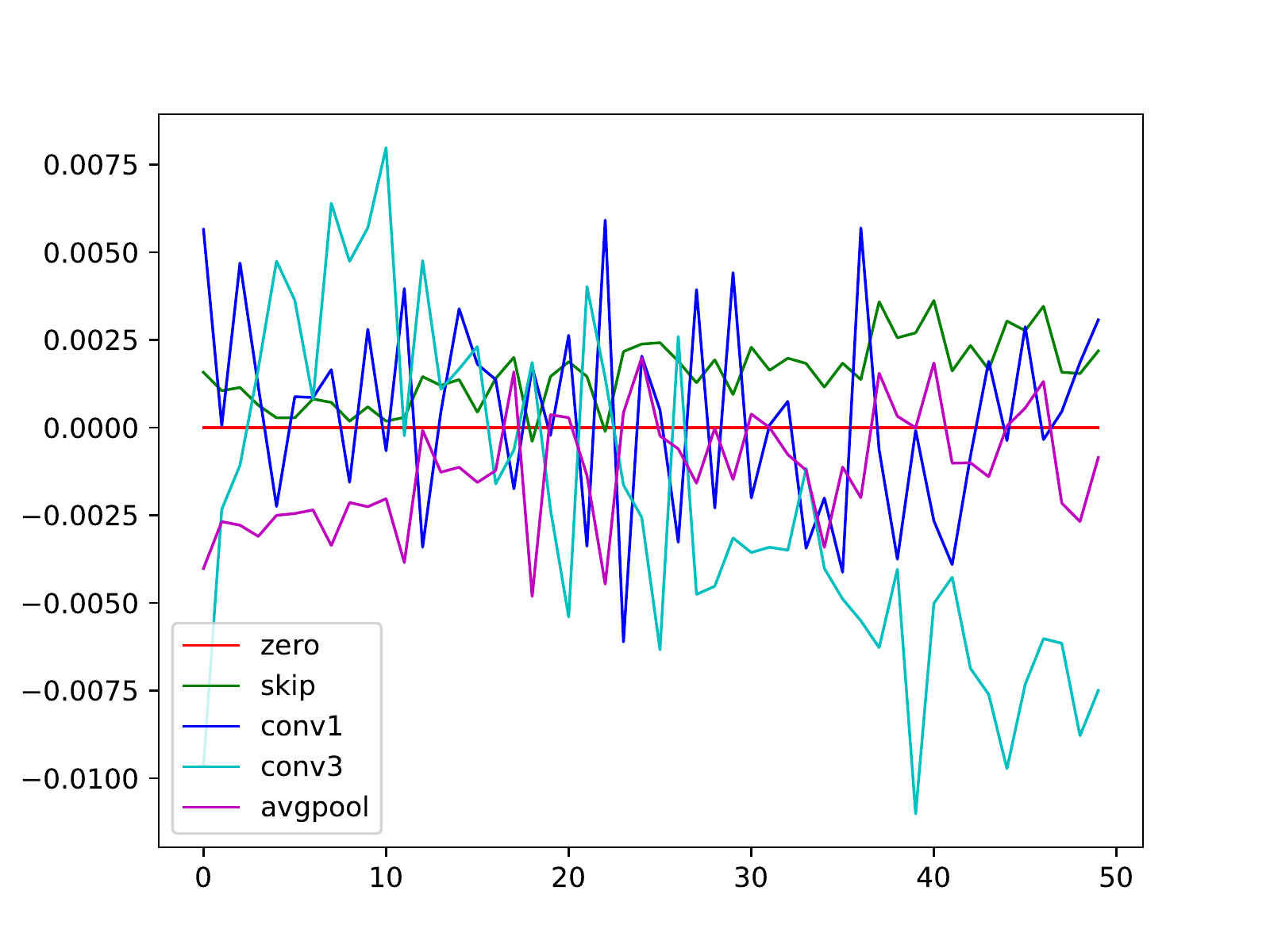}
 \caption{edge.2$\leftarrow$1}
\end{subfigure}
\hfill
\quad
\begin{subfigure}[b]{0.3\linewidth}
 \centering
\includegraphics[width=\textwidth]{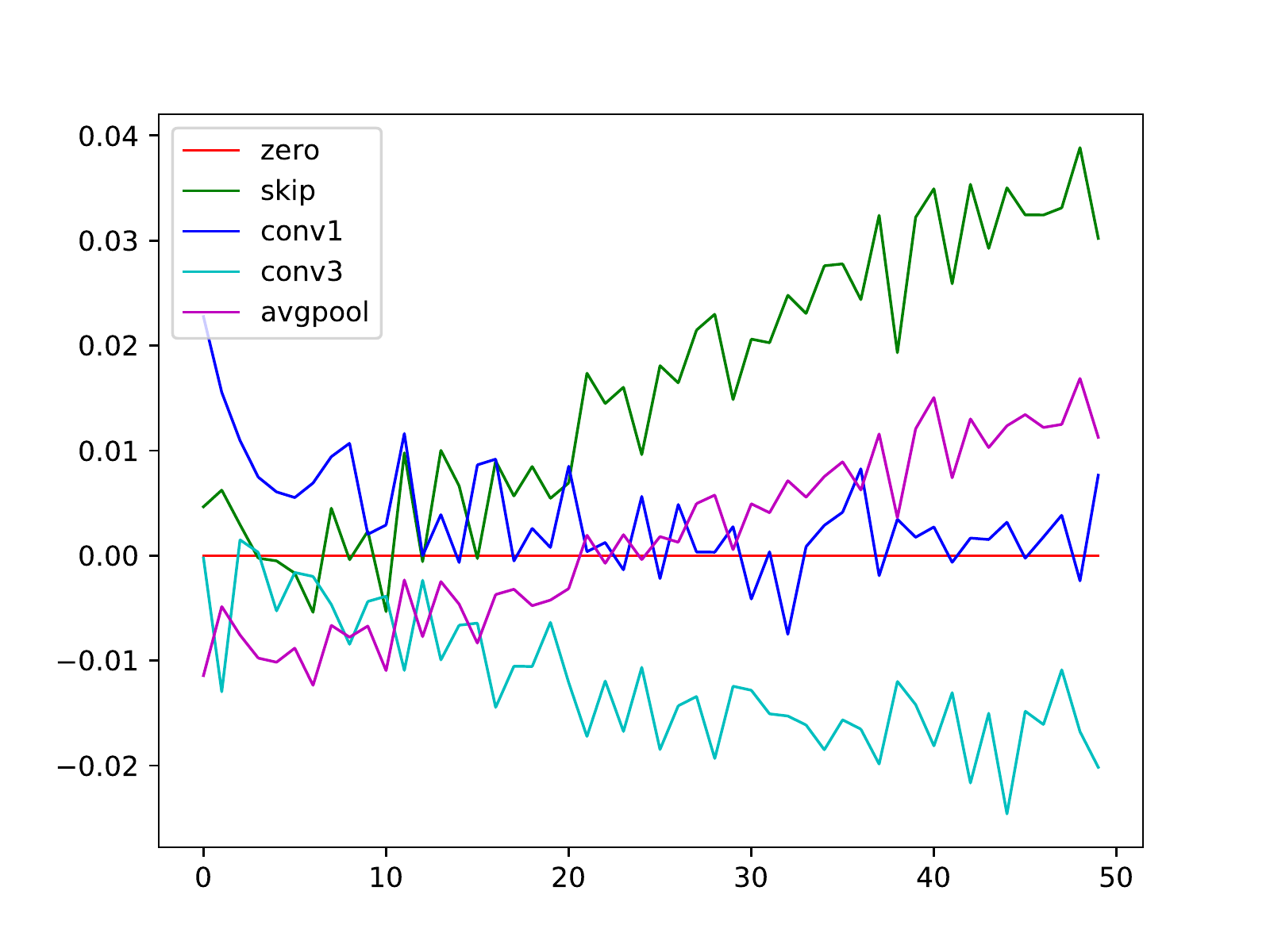}
 \caption{edge.3$\leftarrow$0}
\end{subfigure}
\hfill
 \begin{subfigure}[b]{0.3\linewidth}
 \centering
\includegraphics[width=\textwidth]{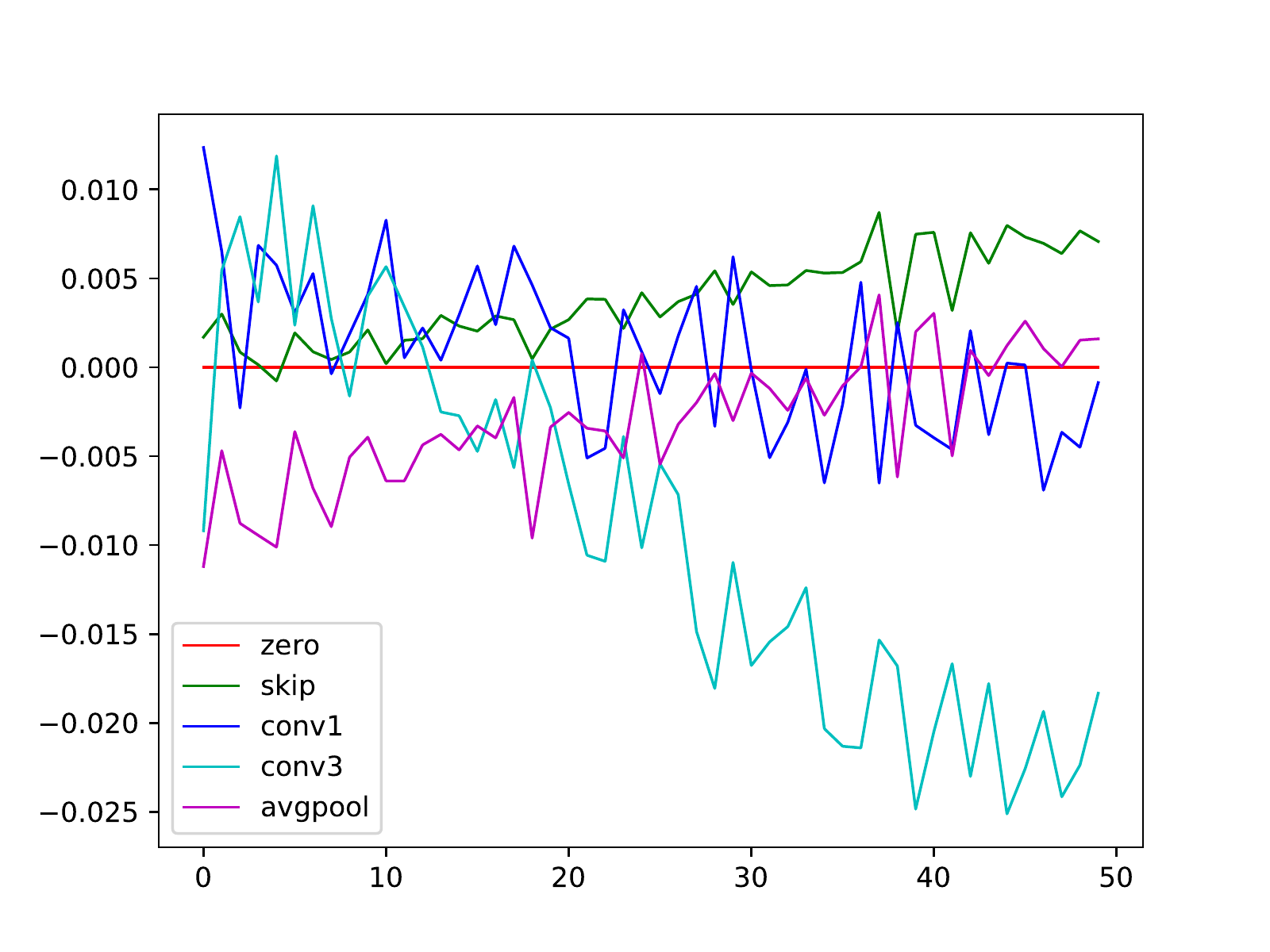}
 \caption{edge.3$\leftarrow$1}
\end{subfigure}
\hfill
 \begin{subfigure}[b]{0.3\linewidth}
 \centering
\includegraphics[width=\textwidth]{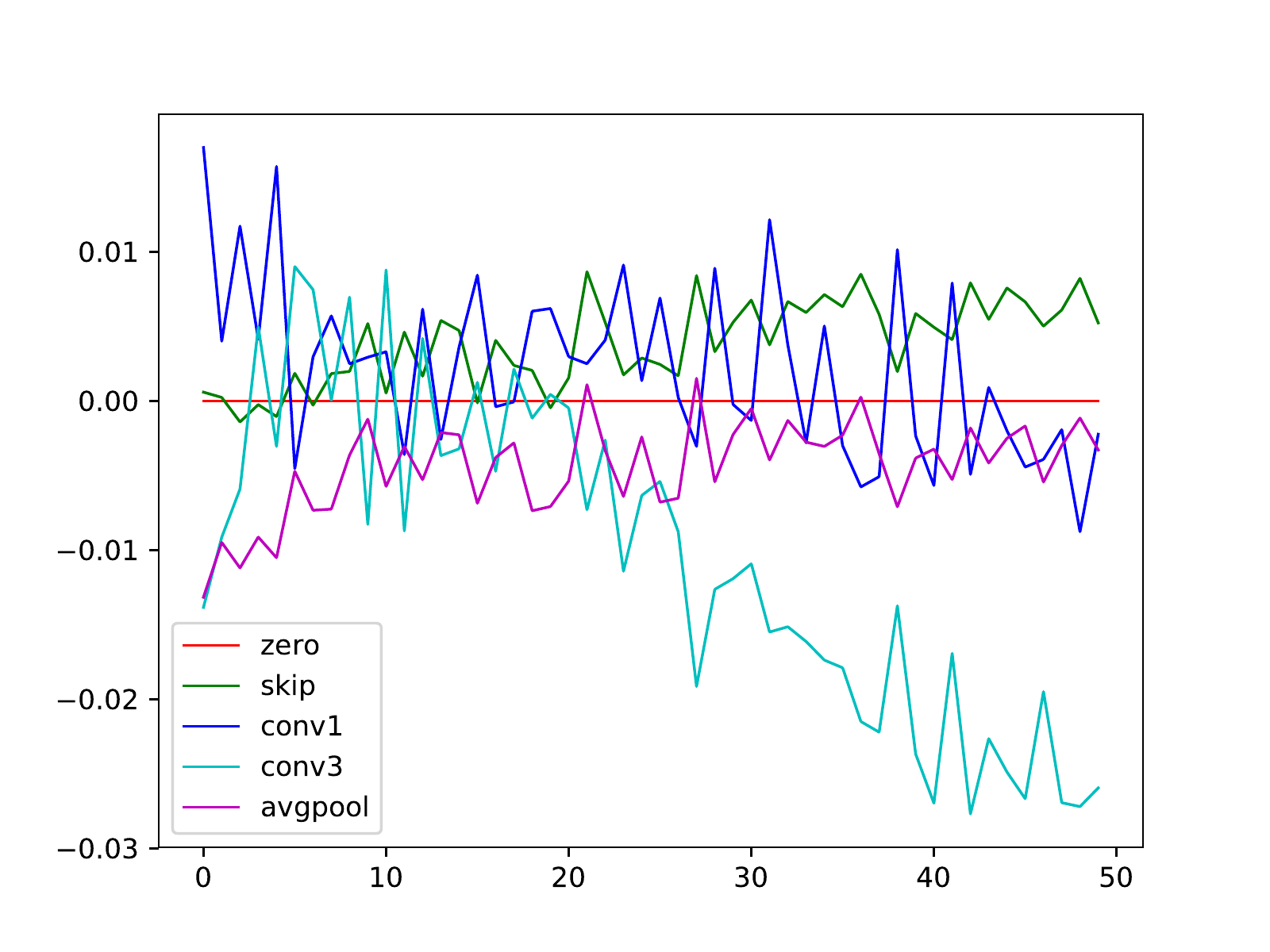}
 \caption{edge.3$\leftarrow$2}
\end{subfigure}
\hfill
\caption{Single-level optimization. On the 16th cell.}
\end{figure}

\section{Searched results in DARTS}
We use single-level optimization to directly search on the ImageNet-1K dataset in DARTS space. Initialize $\alpha_i$ as $-\ln(7)$ so that $\softmax(\alpha_i)=0.125$ would have more better results. It's first time that searched architecture in DARTS space get 77\% top-1 accuracy on ImageNet-1K dataset. The training setting follows PDARTS and without any additional tricks. In addition, for single-level optimization, using half of dataset could search promising results as well.
\begin{table}[h]
  \caption{$*$ denotes $\alpha$ is initialized as $-\ln(7)$. 'Data' means use full or half of data to do search.}
  \label{table:darts-imagenet}
  \centering
  \begin{tabular}{lccccc}
  	\toprule
   \textbf{Activation} & \textbf{Data} & \textbf{Seed}  & \textbf{FLOPs} & \textbf{Params} & \textbf{Top-1.}   \\
 %\multirow{2}{8em}{\textbf{Transfer Learning}} \\
  & (M)&  (M)  &  (\%) \\
    \midrule[1.5pt]
    softmax & full & 0 & 714.72 & 6.58 & 76.28  \\
    softmax & full & 1 & 712.92 & 6.56 & 76.71  \\ 
    softmax & full & 2 & 722.25 &6.61 & 76.27  \\
    sigmoid & full & 0 & 738.21 & 6.69 & 76.12 \\
    sigmoid & full & 1 & 738.21 & 6.69 & 76.58 \\
    sigmoid & full & 2 & 721.35 & 6.60 & 76.29 \\
    sigmoid* & full & 0 & 707.89 & 6.50 & 76.26  \\
    sigmoid* & full & 1  & 721.35 &6.60 &  77.0 \\
    sigmoid* & full & 2 & 700.61 & 6.40 & 76.95 \\
    softmax & half & 0 & 712.01 & 6.55 & 76.51  \\
    softmax & half & 1 & 692.18 & 6.36 & 76.31	  \\ 
    softmax & half & 2 & 709.04 & 6.45 & 76.51  \\
    sigmoid* & half & 0 & 720.44 & 6.59 & 76.67  \\
    sigmoid* & half & 1 & 692.18 & 6.36 &  76.78 \\
    sigmoid* & half & 2 & 707.89 & 6.50 & 76.54 \\
    \midrule[1.5pt]
  \end{tabular}
\end{table}

\section{Visualization of architectures}
\begin{figure}[h]
 \begin{subfigure}[b]{0.49\linewidth}
  \parbox[][2.2cm][c]{\linewidth}{
 \centering
\includegraphics[width=\textwidth]{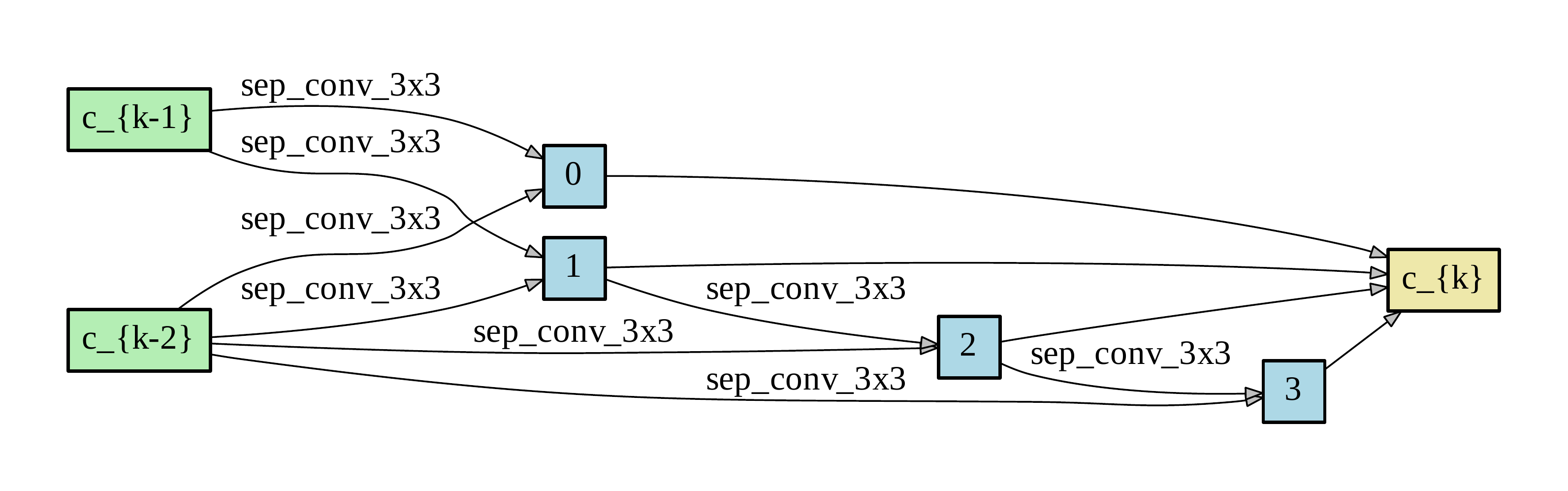}
 }
 \caption{normal}
\end{subfigure}
\hfill
 \begin{subfigure}[b]{0.49\linewidth}
 \centering
\includegraphics[width=\textwidth]{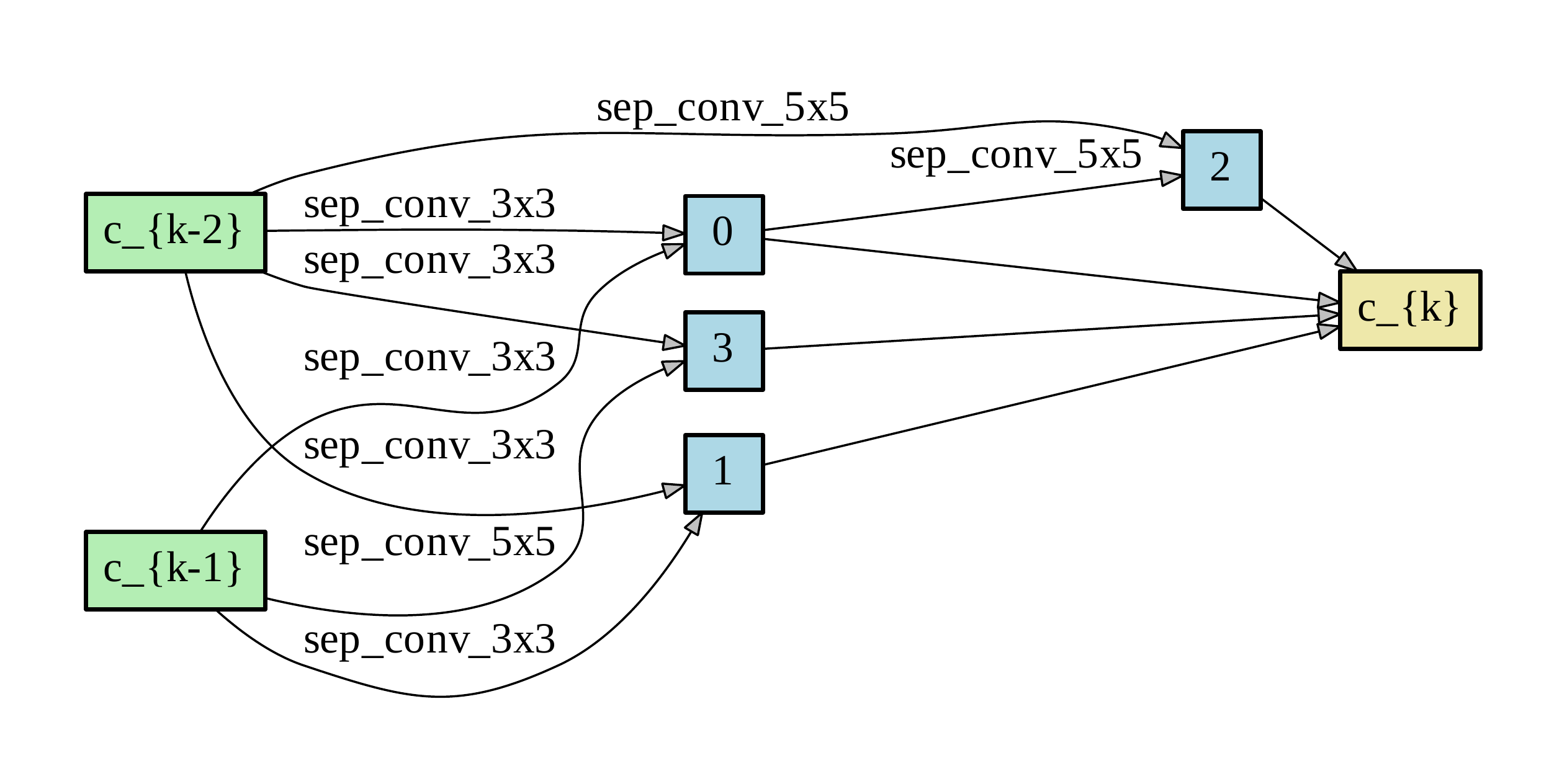}
 \caption{reduction}
\end{subfigure}
\hfill
\quad
\caption{activation=softmax, data=full, seed=0}
\end{figure}

\begin{figure}[h]
 \begin{subfigure}[b]{0.49\linewidth}
 \centering
\includegraphics[width=\textwidth]{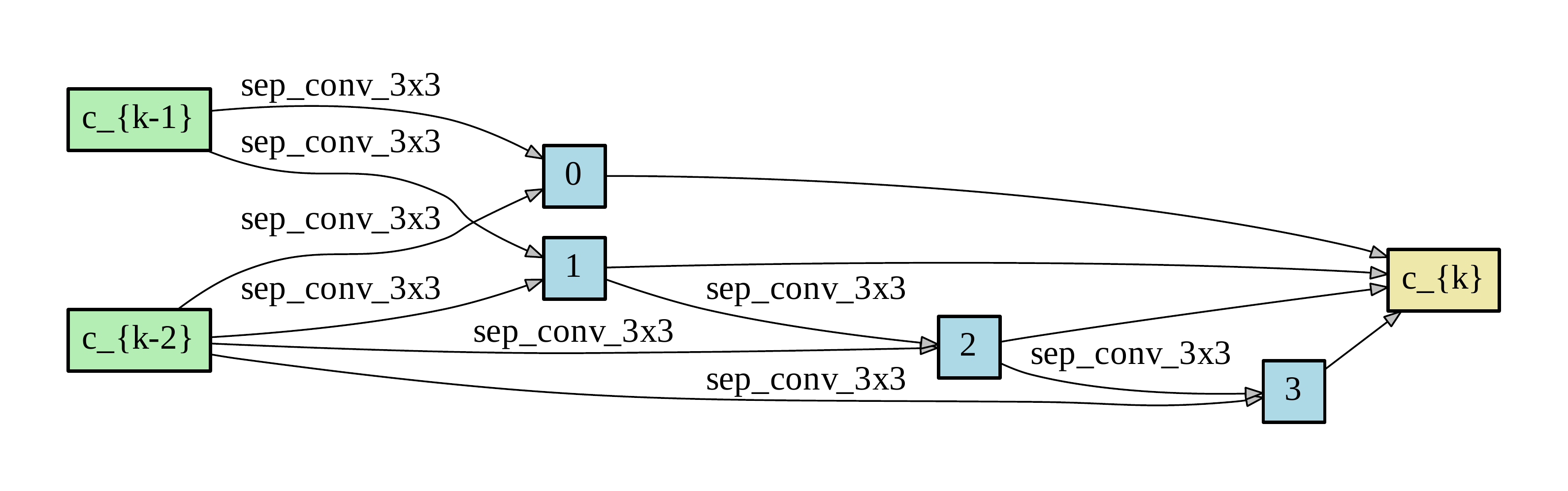}
 \caption{normal}
\end{subfigure}
\hfill
 \begin{subfigure}[b]{0.49\linewidth}
 \centering
\includegraphics[width=\textwidth]{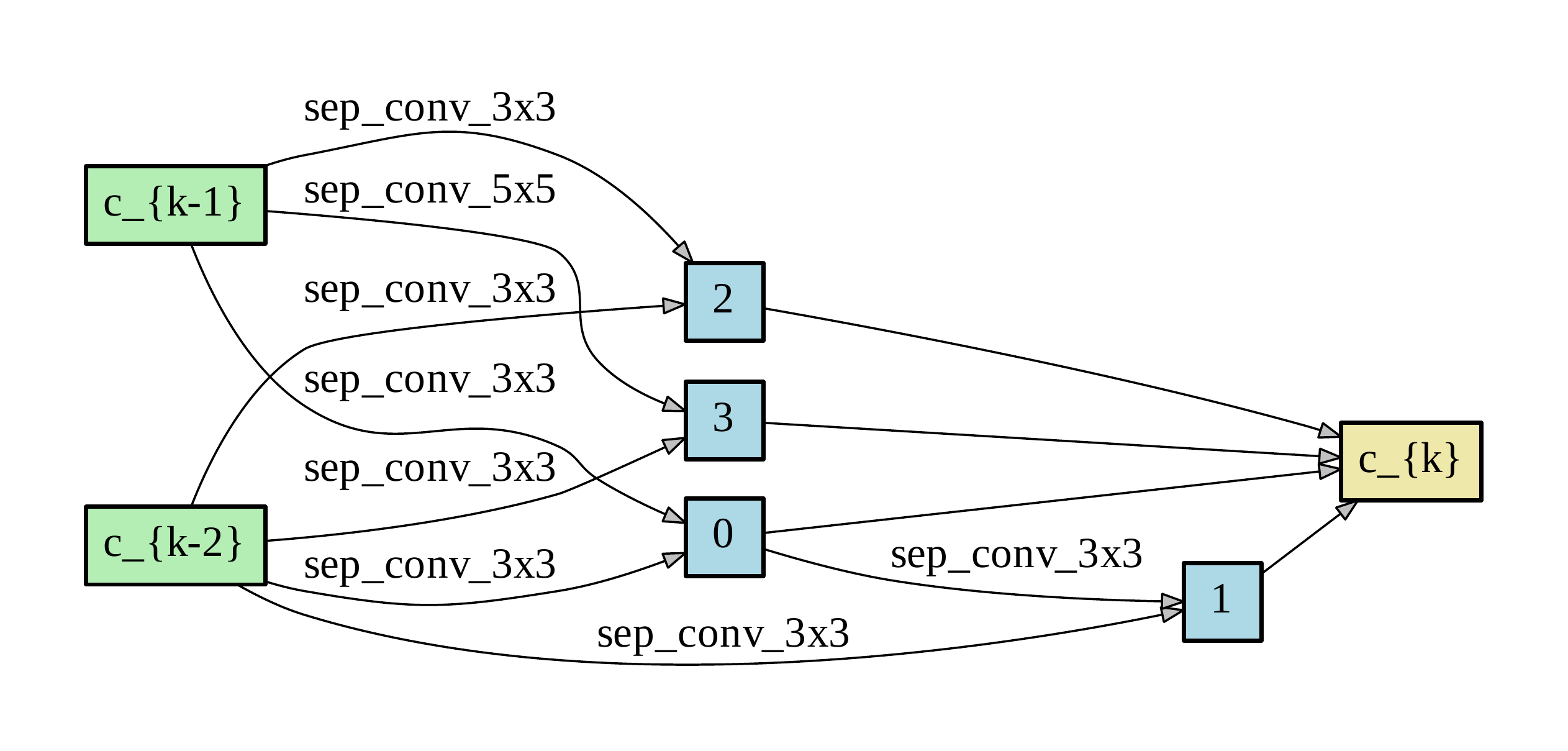}
 \caption{reduction}
\end{subfigure}
\hfill
\quad
\caption{activation=softmax, data=full, seed=1}
\end{figure}

\begin{figure}[h]
 \begin{subfigure}[b]{0.49\linewidth}
 \parbox[][2.0cm][c]{\linewidth}{
 \centering
\includegraphics[width=\textwidth]{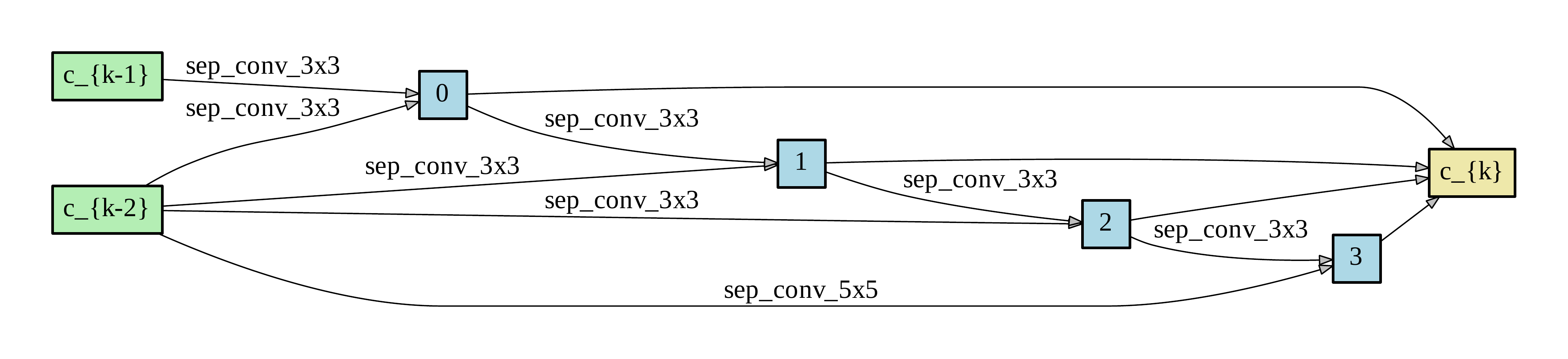}
 }
 \caption{normal}
\end{subfigure}
\hfill
 \begin{subfigure}[b]{0.49\linewidth}
 \centering
\includegraphics[width=\textwidth]{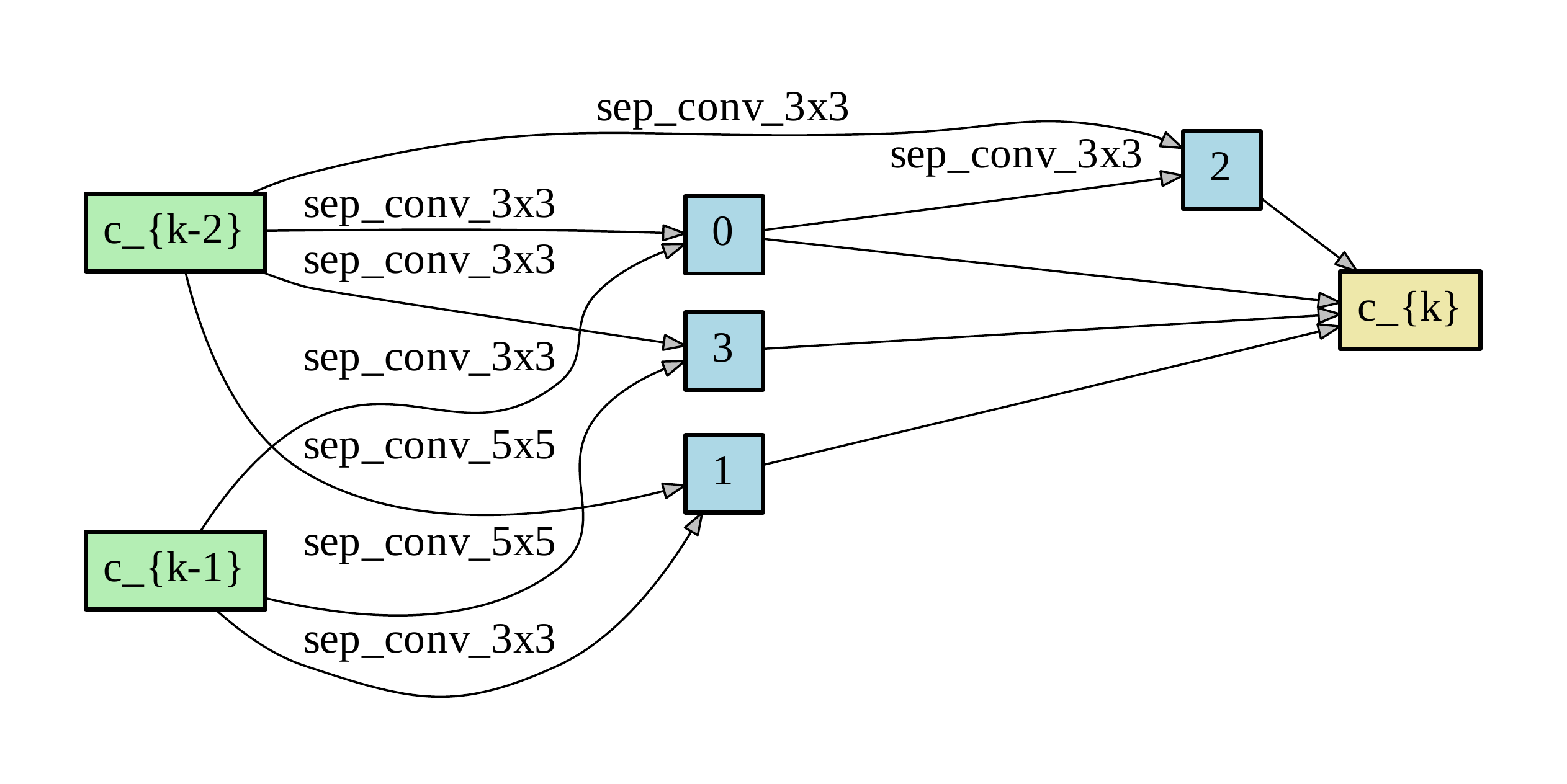}
 \caption{reduction}
\end{subfigure}
\hfill
\quad
\caption{activation=softmax, data=full, seed=2}
\end{figure}

\begin{figure}[h]
\centering
 \begin{subfigure}[b]{0.49\linewidth}
 \parbox[][3.5cm][c]{\linewidth}{
 \centering
\includegraphics[width=\textwidth]{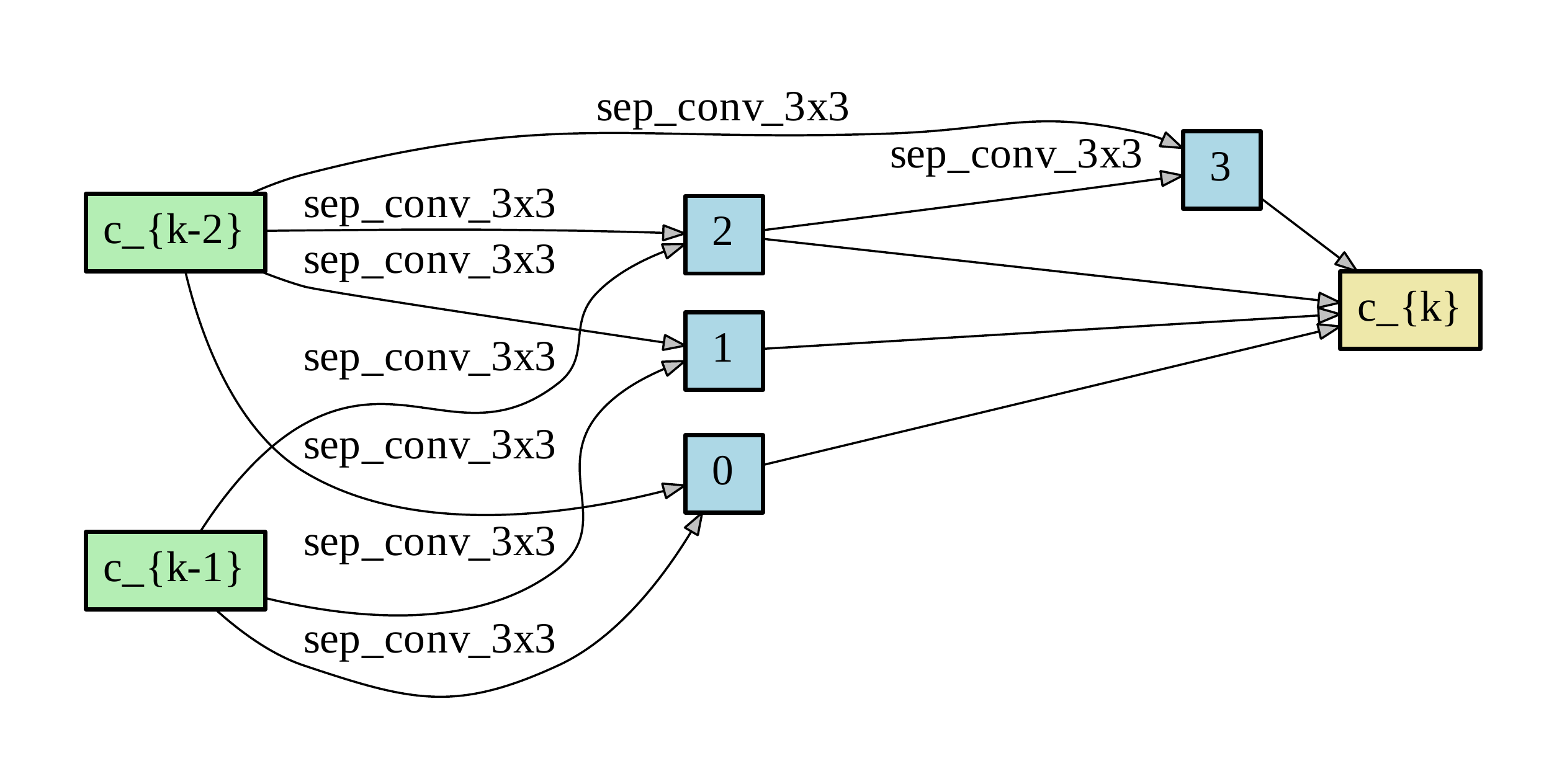}
}
 \caption{normal cell}
\end{subfigure}
\hfill
 \begin{subfigure}[b]{0.49\linewidth}
 \centering
\includegraphics[width=\textwidth]{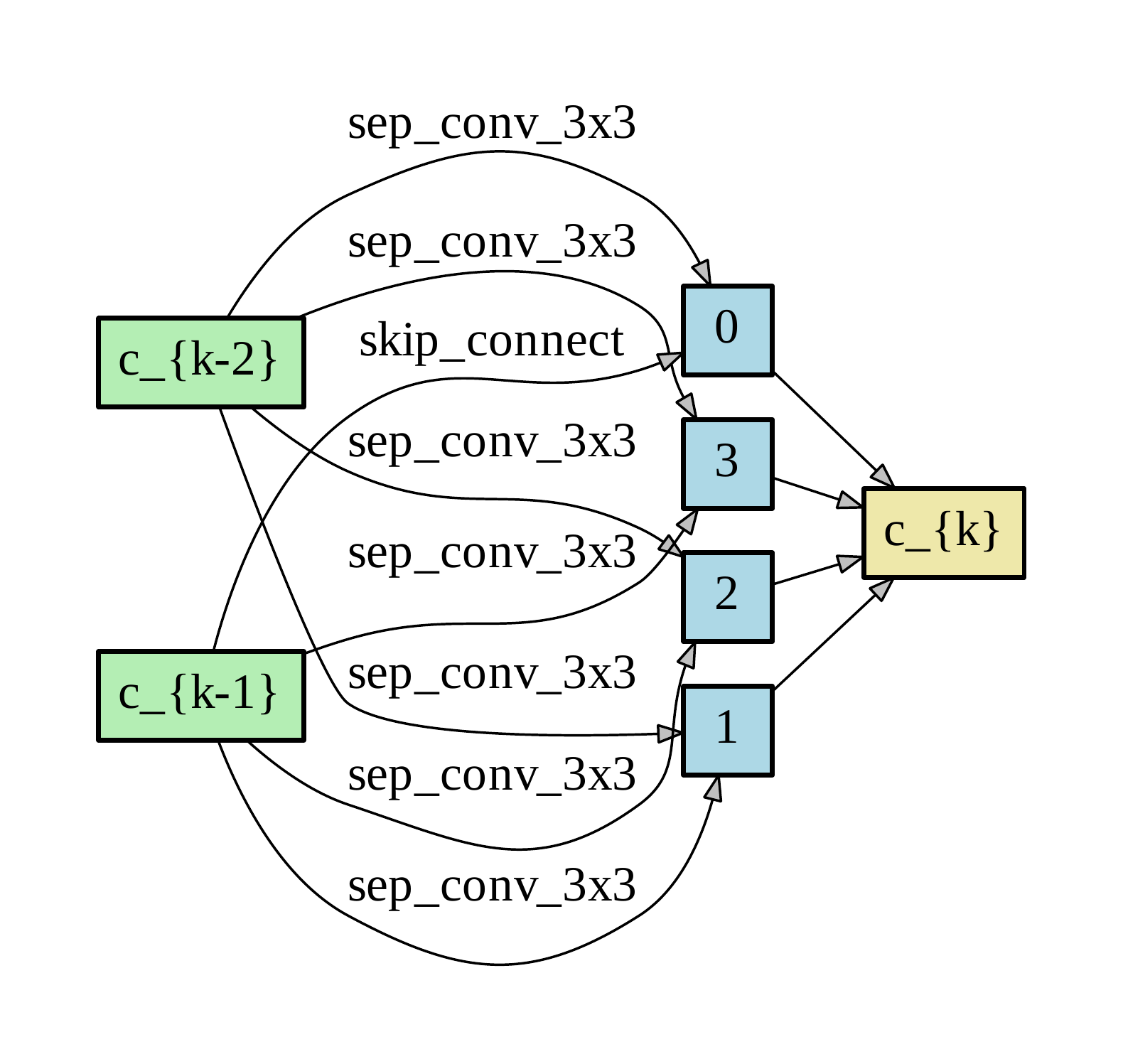}
 \caption{reduction cell}
\end{subfigure}
\hfill
\quad
\caption{activation=sigmoid*, data=full, seed=0}
\end{figure}

\begin{figure}[h]
 \begin{subfigure}[b]{0.49\linewidth}
 \parbox[][3.5cm][c]{\linewidth}{
 \centering
\includegraphics[width=\textwidth]{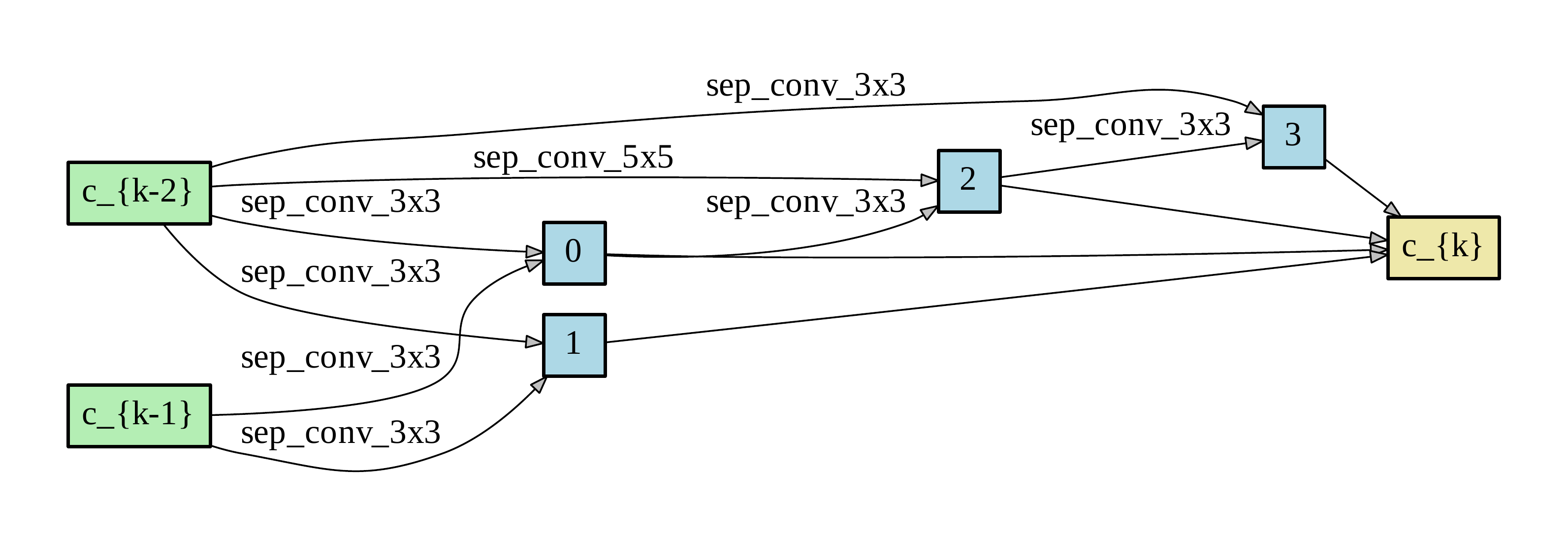}
 }
 \caption{normal}
\end{subfigure}
\hfill
 \begin{subfigure}[b]{0.49\linewidth}
 \centering
\includegraphics[width=\textwidth]{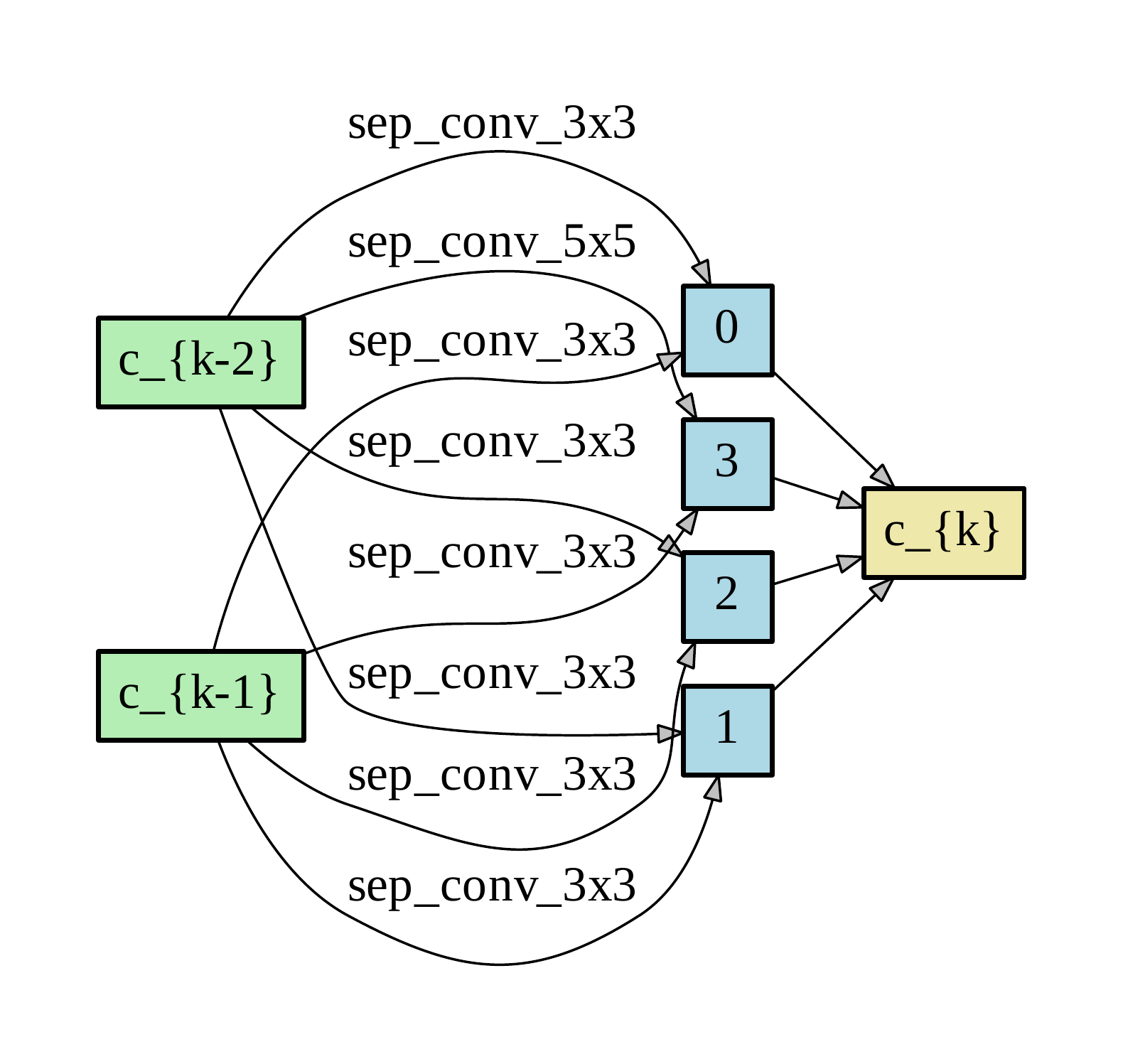}
 \caption{reduction}
\end{subfigure}
\hfill
\quad
\caption{activation=sigmoid*, data=full, seed=1}
\end{figure}

\begin{figure}[t]
 \begin{subfigure}[b]{0.49\linewidth}
 \parbox[][3.5cm][c]{\linewidth}{
 \centering
\includegraphics[width=\textwidth]{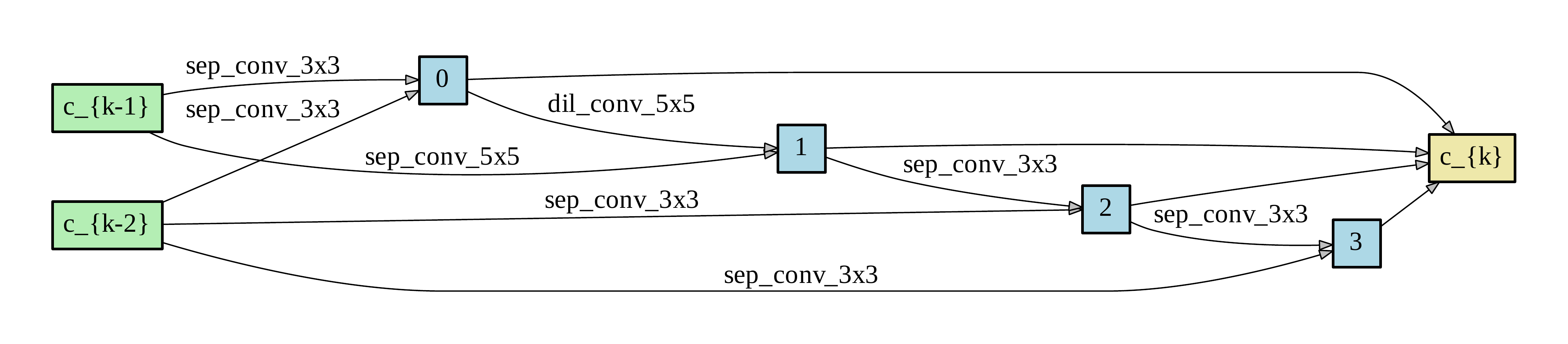}
 }
 \caption{normal}
\end{subfigure}
\hfill
 \begin{subfigure}[b]{0.49\linewidth}
 \centering
\includegraphics[width=\textwidth]{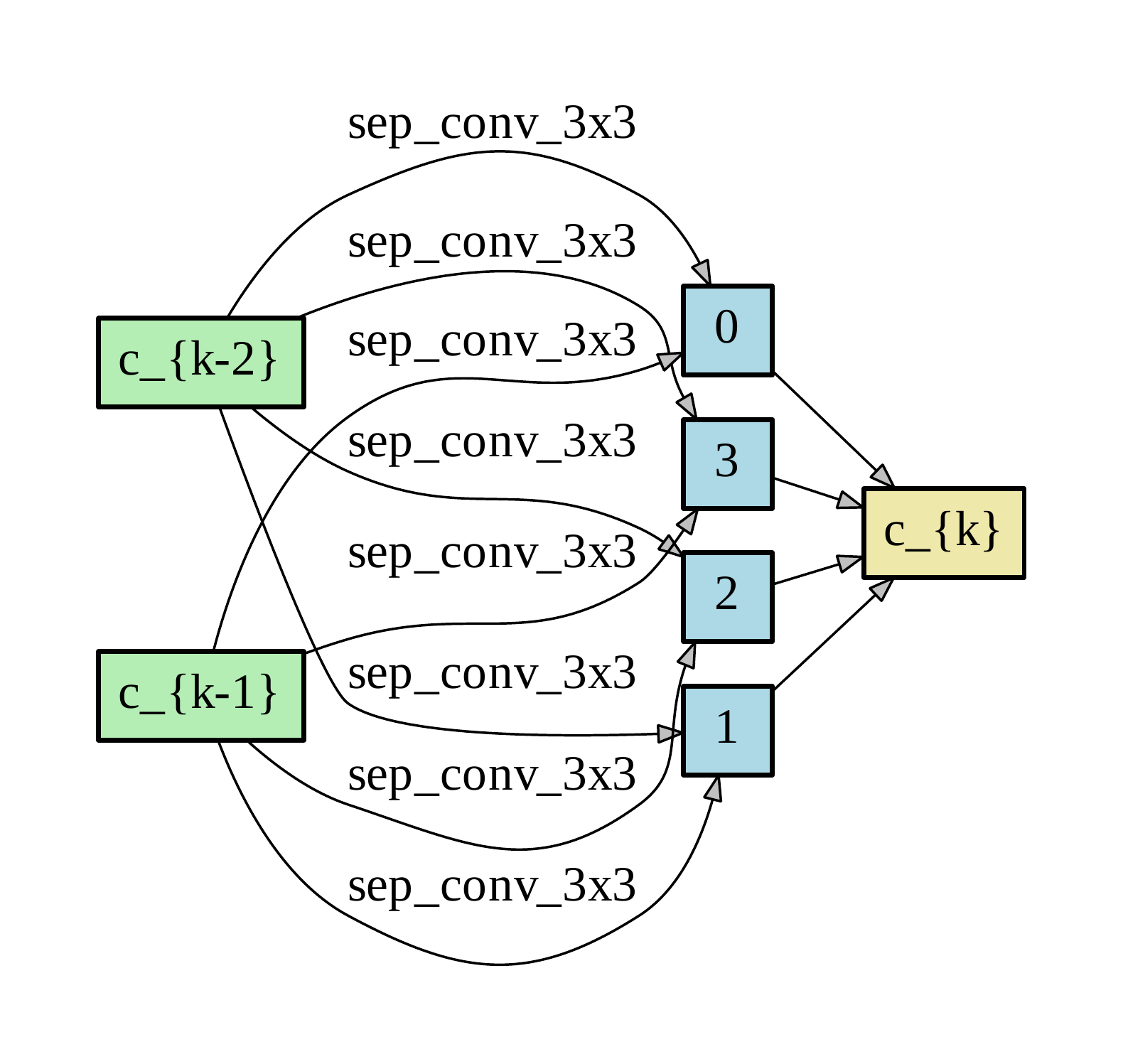}
 \caption{reduction}
\end{subfigure}
\hfill
\quad
\caption{activation=sigmoid*, data=full, seed=2}
\end{figure}

\end{document}